\documentclass[conference]{IEEEtran}

\usepackage{amsmath,amssymb,amsfonts}
\usepackage{algorithmic}
\usepackage[numbers,sort&compress]{natbib}

\usepackage{graphicx}
\usepackage{textcomp}

\usepackage{xcolor}
\usepackage{url}

\usepackage{array}
\usepackage{makecell}

\usepackage{bm}

\usepackage{fourier} 
\usepackage{float}

\usepackage{multicol}

\usepackage{moresize}

\usepackage{placeins}

\usepackage{enumerate}

\usepackage{multirow}

\usepackage{hyperref}

\usepackage{booktabs}   % nicer rules
\usepackage{tabularx}   % X column
\usepackage{multirow}   % multi‑row cells
\usepackage{arydshln}

\usepackage{pdflscape} 

\usepackage{pdfpages}

\def\BibTeX{{\rm B\kern-.05em{\sc i\kern-.025em b}\kern-.08em
    T\kern-.1667em\lower.7ex\hbox{E}\kern-.125emX}}

\newcommand{{\xx}}{{\mathbf x}}

\def\ppn{\vskip 6pt \noindent }
\def\R{{\mathbb{R}}}
\def\N{{\mathbb{N}}}
\def\P{{\mathbb{P}}}
\def\E{{\mathbb{E}}}
\newcommand{{\Xs}}{{\cal X}}
\newcommand{{\Ys}}{{\cal Y}}
\newcommand{{\Ls}}{{\cal L}}
\newcommand{{\Ss}}{{\cal S}}
\newcommand{{\Ms}}{{\cal M}}
\newcommand{{\Gs}}{{\cal G}}
\newcommand{{\Hs}}{{\cal H}}
\newcommand{{\Ns}}{{\cal N}}
\newcommand{{\Is}}{{\cal I}}
\newcommand{{\Vs}}{{\cal V}}
\newcommand{{\Ds}}{{\cal D}}
\newcommand{{\Bs}}{{\cal B}}
\newcommand{{\Zs}}{{\cal Z}}
\newcommand{{\Cs}}{{\cal C}}
\newcommand{{\Rs}}{{\cal R}}
\newcommand{{\Fs}}{{\cal F}}
\newcommand{{\Us}}{{\cal U}}
\newcommand{{\Ps}}{{\cal P}}
\newcommand{{\ttheta}}{{\bm{\theta}}}
\newcommand{{\Ttheta}}{{\bm{\Theta}}}
\newcommand{{\Oomega}}{{\bm{\Omega}}}
\newcommand{{\Sss}}{{\bm{\Ss}}}
\newcommand{{\pp}}{{\mathbf p}}
\newcommand{{\ww}}{{\mathbf w}}
\newcommand{{\mm}}{{\mathbf m}}
\newcommand{{\uu}}{{\mathbf u}}
\newcommand{{\ppi}}{{\bm{\pi}}}
\newcommand{{\phhi}}{{\bm{\phi}}}
\newcommand{{\pssi}}{{\bm{\psi}}}
\newcommand{{\XX}}{{\mathbf X}}
\newcommand{{\UU}}{{\mathbf U}}
\newcommand{{\BB}}{{\mathbf B}}
\newcommand{{\WW}}{{\mathbf W}}
\newcommand{{\KK}}{{\mathbf K}}
\newcommand{{\HH}}{{\mathbf H}}
\newcommand{{\II}}{{\mathbf I}}
\newcommand{{\PP}}{{\mathbf P}}
\newcommand{{\yy}}{{\mathbf y}}
\newcommand{{\ee}}{{\mathbf e}}
\newcommand{{\ab}}{{\mathbf a}}

\newcommand{{\dd}}{{\mathbf d}}
\newcommand{{\zero}}{{\mathbf 0}}
\newcommand{{\uno}}{{\mathbf 1}}

\newcommand{{\dep}}{{\mathfrak{D}}}
\newcommand*\im{\mathrm{i}} %% imaginary number
\newcommand{\pkg}[1]{\texttt{\small #1}}

\newcommand\indep{\protect\mathpalette{\protect\independenT}{\perp}}
\def\independenT#1#2{\mathrel{\rlap{$#1#2$}\mkern2mu{#1#2}}}

\newcommand{{\toL}}{{\overset{\mathcal{L}}{\longrightarrow}\ }}

\newcommand{{\MC}}{{\,  *_{\text{\scalebox{0.65}{$\Ms$}}}\,  }}

\newcommand{{\dou}}{$\leadsto$\ }

\DeclareMathOperator{\var}{\mathbb{V}ar}
\DeclareMathOperator{\cov}{\mathbb{C}ov}

\newcommand{\indic}[1]{
	\hbox{$\mathbb{I}_{\{ #1 \}}$}
}

\begin{document}
\title{Deep-testing: the case of dependence detection \\
%{\footnotesize \textsuperscript{*}Note: Sub-titles are not captured in Xplore and should not be used}
%\thanks{ }
}
\author{\IEEEauthorblockN {\textsc{Gery Geenens}}
\IEEEauthorblockA{\textit{School of Mathematics and statistics, } \\
\textit{UNSW Sydney, Australia}\\
\pkg{ggeenens@unsw.edu.au}}
\and
\IEEEauthorblockN{ \textsc{Pierre Lafaye de Micheaux}}
\IEEEauthorblockA{\textit{School of Mathematics and statistics, } \\
	\textit{UNSW Sydney, Australia}\\
	\pkg{lafaye@unsw.edu.au}}
\and
\IEEEauthorblockN{\textsc{Ivan Muyun Zou}}
\IEEEauthorblockA{\textit{School of Mathematics and statistics, } \\
	\textit{UNSW Sydney, Australia}\\
	\pkg{muyun.zou@unswalumni.com}}
}
\maketitle
\thispagestyle{plain}
\pagestyle{plain}
%%%%%
%%%%%
\begin{abstract}
Deep learning methods have proved highly effective for classification and image recognition problems. In this paper, we ask whether this success can be transferred to hypothesis testing: if a neural network can distinguish, for example, an image of a handwritten digit from another, can it also distinguish an `image of a sample' -- such as a scatter plot -- generated under a given statistical model from one generated outside that model? Motivated by this idea, we propose a novel procedure called \emph{deep-testing}, which approaches the classical inferential problem of hypothesis testing through deep learning. More specifically, the test statistic is a classification map learned by a deep neural network from simulated data satisfying the null and alternative hypotheses, leveraging its strong discriminating power to construct a highly powerful test. As a proof of concept, we apply deep-testing to the problem of independence testing, arguably one of the most important problems in statistics. In a large-scale simulation study, deep-testing achieves the highest overall power against nineteen competing methods across a broad range of complex dependence structures, confirming the viability of the proposed approach.
\end{abstract}

\section{Introduction} \label{sec:intro}

In recent years, {\it deep learning} \cite{Lecun15} has emerged as a tool of `unreasonable effectiveness' \cite{Sejnowski20}. Indeed deep neural networks have achieved remarkable practical performance on previously intractable tasks across many fields, including in computer vision, speech recognition, robotics, machine translation, natural language processing and game playing (e.g., chess or Go).

\ppn One of the settings in which deep learning has proved especially successful is {\it classification} ({\it supervised learning}); i.e., the task of assigning inputs to pre-determined categories. We are given a training dataset $\{(\xx_i,y_i) \}_{i=1}^N$, where for each unit $i$, $\xx_i \in \Xs \subseteq \R^d$ ($d \in \N$) is a vector of {\it input} features, and $y_i \in \{0,\ldots,\kappa\}$ ($\kappa \in \N$) is a label, i.e.\ a scalar {\it output} characterising a category which this unit belongs to. The classification problem is then to make use of this training set to design a map $\varphi: \Xs \to \{0,\ldots,\kappa\}$ maximising the prediction power: if a new unit described by features $\xx$ was to be observed, the predicted category $\varphi(\xx)$ would be optimal in a certain sense. Typically, we would seek to identify, within a suitable class of candidates, the map $\varphi$ which minimises a loss (contrast) function $\ell(y,\varphi(\xx))$ between the effective category $y$ and its prediction $\varphi(\xx)$.

\ppn A canonical instance of this classification problem is {\it image recognition}, where the input $\xx$ is a raw image represented as a matrix of pixel intensities (for example, 0 for black and 255 for white in a monochrome image), and the output $y$ is the corresponding class label.  A standard benchmark is the MNIST dataset \cite{Lecun98}, which consists of 60,000 labelled grayscale images of handwritten digits, each of size $28\times 28$. A wide range of machine learning and image-processing methods have been applied to this task, with deep neural networks -- in particular convolutional neural networks (CNNs) -- achieving test accuracies above 99\% \cite{Hasanpour16}.

\ppn In this paper, we bring this perspective back to a classical problem in statistics: {\it hypothesis testing}, one of the pillars of statistical inference \cite{Lehmann05}. Generally, the problem is the following: we have a sample $\Zs$ of observations ({\it data}) and we are interested in determining whether or not the distribution $F$ which generated these data satisfies some constraints -- typically, a statistical model. Formally, the said constraints characterise a strict subset $\Fs_0$ of the set $\Fs$ of all the distributions admissible for the situation at hand, and we wish to test the null hypothesis: 
\begin{equation} H_0: F \in \Fs_0 \label{eqn:H0} \end{equation}
against the alternative:
\begin{equation} H_1: F \in \Fs \backslash \Fs_0. \label{eqn:Ha} \end{equation}
The standard procedure is to compute on the observed sample $\Zs$ a relevant test statistic $\phi$, whose random behaviour (`sampling distribution') when computed on samples generated by distributions complying with $H_0$ is known (at least approximately). Comparing the observed value $\phi(\Zs)$ to what it is expected to be if $H_0$ is valid, allows us to make a decision (`reject $H_0$' or `do not reject $H_0$') while keeping control on the probability of Type I error (i.e., rejecting $H_0$ whereas it is true). A central aim of classical statistical theory and practice is to construct, for a given setting, a test statistic with the greatest possible power, that is, the highest probability of rejecting $H_0$ when it is false.

\ppn Less classically, we may view the hypothesis test above as a binary classification problem. Specifically, we can regard the sample $\Zs$ as a set of input features to which we wish to assign the label $y=0$ if the distribution $F$ that generated $\Zs$ belongs to $\Fs_0$, and $y=1$ otherwise. From this perspective, and assuming access to a large collection of training samples generated under both $H_0$ and $H_1$, we can use deep learning methods to construct the `best' classifier. In this setting, such a classifier yields a test statistic $\phi$ that should discriminate sharply between samples generated under $H_0$ and those generated under $H_1$. Our premise is that the `unreasonable effectiveness' of deep learning carries over to this testing framework, producing procedures with greater power than conventional alternatives. We refer to the resulting approach as \emph{deep-testing}. 

\ppn More concretely, the question we investigate in this paper is the following: if deep learning can distinguish highly structured objects such as handwritten digits with near-perfect accuracy, can it also distinguish suitable graphical and/or numerical representations of samples generated under $H_0$ from those generated under $H_1$? Our main contribution is to develop and assess this idea, and to offer a simple and flexible framework supporting it. This complements recent papers pursuing deep-learning methods for nonparametric hypothesis testing \cite{condepGAN,Gerber23,Paschali22,Pandeva24,Yang25,Xu26}, but which mainly emphasize learned representations or bounds for test statistics.

\ppn Our goal, at this stage, is not to develop a fully general theory of deep-testing. Indeed, even the statistical principles behind the success of deep learning more broadly remain only partially understood \cite{DL,Bahri20,Bartlett21,Survey}. Rather, our objective is to demonstrate here the viability of the approach through a detailed study of one of the most fundamental problems in statistics: {\it testing for independence between two random variables}. The paper should therefore be read as a \emph{proof of concept}: it introduces the general deep-testing framework and studies its empirical performance in depth in a specific setting -- independence testing -- where both the statistical problem and the competing methods are well understood. 

\ppn Relatedly, a second contribution of the paper is a substantial review of independence testing and dependence measures. This review is of independent interest, not only as background for the present work, but also as a broader overview of a rich and active area of statistics. This review is accompanied by the development of an R package that brings together, within a unified framework, practical implementations of most of the popular dependence measures and independence tests; see Table \ref{tab:indicators} below. This package forms the backbone of a large-scale simulation study comparing the power of deep-testing against independence tests based on nineteen other test statistics -- all results are fully reproducible,\footnote{see \href{https://biostatisticien.eu/DeepTesting}{\texttt{\scriptsize biostatisticien.eu/DeepTesting}}.} following the guidelines in \cite{PoweR}.

\ppn The remainder of the paper is organised as follows. Section \ref{sec:pierre} develops the general theoretical framework underlying deep-testing by viewing hypothesis testing through the lens of classification. Section \ref{sec:deptect} then specialises this perspective to independence testing. The dependence indicators used as inputs are reviewed in Section \ref{sec:litrev}. Section \ref{sec:meth} details the methodology, Section \ref{sec:res} presents the empirical results, and Section \ref{sec:ccl} concludes with directions for future work.

\section{Classification, hypothesis testing and deep learning} \label{sec:pierre}

\noindent We recall here the machinery of binary\footnote{$\kappa=1$ in the notation of Section \ref{sec:intro}.} classification and hypothesis testing, and we introduce briefly the elements of deep learning  that will guide our deep-testing procedure. 

\ppn Let $(\XX,Y)$ be a random vector, where $\XX = (X_1,\ldots,X_d)^\top\in\mathcal{X}\subset\mathbb{R}^d$ consists of quantitative explanatory variables (often called observable input features in machine learning) and $Y \in \{0,1\}$ serves as class label. The joint distribution of $(\XX,Y)$ is entirely characterised by the marginal distribution of $\XX$ and the conditional probability function $p:\Xs \to [0,1]$:
\begin{equation}  p(\cdot) = \P(Y=1 \mid \XX=\cdot). \label{eqn:pX} \end{equation}

\ppn Define a {\it classifier} as a map $\varphi: \mathcal{X} \to \{0,1\}$ that would predict by $\varphi(\xx)$ the unknown label $y$ of a unit described by the vector $\xx \in \mathcal{X}$. The classification error rate (or expected prediction error) of this classifier $\varphi$ is\footnote{$\indic{\bullet}$ is the indicator function, taking the value 1 if statement $\bullet$ is true and 0 if not.} 
\begin{equation}   R(\varphi)   = \E\left(\indic{\varphi(\XX)\ne Y} \right)  =  \P\left(\varphi(\XX)\ne Y\right). %=  
\label{eqn:Rphi}  \end{equation}
The optimal classifier -- that is, the one minimising (\ref{eqn:Rphi}) -- is known as the {\it Bayes classifier} and takes the simple form \cite[Chapter 2]{Devroye96}
\begin{equation} \varphi^\star(\xx) = \underset{\varphi}{\arg\min}~R(\varphi) = \indic{p(\xx) > \frac{1}{2}}. \label{eqn:bayesclass} \end{equation}
This simply prescribes to assign the unit $\xx$ to the class 1 ($Y=1$) if it is more likely than not that it belongs to that class; i.e., if $\P(Y = 1 \mid \XX = \xx) > \P(Y = 0 \mid \XX = \xx)$. Note that $\varphi^\star$ is an {\it oracle}, in the sense that it is based on the distribution of $(\XX,Y)$ which is typically unknown in practice. A feasible version of it would require to estimate $p(\cdot)$, a task for which there exist a variety of approaches \cite{Hastie13}. Note that (\ref{eqn:Rphi}) can be expanded further by the Law of Total Probability as 
\begin{multline} R(\varphi)  =  \P\left(\varphi(\XX)\ne Y\right) \\ = \pi_0\P\left(\varphi(\XX)\ne Y\mid Y=0\right)+\pi_1\P\left(\varphi(\XX)\ne Y\mid Y=1\right) \label{eqn:RvarphiTP} \end{multline} 
where $\pi_j=\P(Y=j)$, $j=0,1$, are the {\it a priori} class probabilities -- that is, the probabilities that a random unspecified unit belongs to each class.  

\ppn Now, as suggested in Section \ref{sec:intro}, from a hypothesis testing perspective $[$(\ref{eqn:H0}) versus (\ref{eqn:Ha})$]$, we may want the statement `$Y=0$' to mean `the null hypothesis $H_0$ is true'; that is, the distribution of $\XX$ is some $F$ in $\Fs_0$. Similarly, we might equate `$Y=1$' and `the alternative hypothesis $H_1$ is true'; i.e.,  the distribution of $\XX$ does not belong to $\Fs_0$. Within this framework, the `classifier' $\varphi$ is a test which will return 1 if $H_0$ is rejected, and 0 if not, and the probabilities $R_0(\varphi) \doteq \P\left(\varphi(\XX)\ne Y\mid Y=0\right)$ and $R_1(\varphi) \doteq \P\left(\varphi(\XX)\ne Y\mid Y=1\right)$ are the type-I error and type-II error rates, respectively. As opposed to bare classification as in (\ref{eqn:bayesclass})/(\ref{eqn:RvarphiTP}), (frequentist) hypothesis testing disregards `prior probabilities' of $H_0$ or $H_1$ to be true,\footnote{Such probabilities are considered under the Bayesian paradigm.} and defines the optimal classifier -- then referred to as the {\it Uniformly Most Powerful test} -- as the solution of the following  problem:
\begin{equation}
\varphi_\alpha^* = \underset{\varphi:R_0(\varphi)\leq\alpha}{\arg\min}~R_1(\varphi)
\label{NPcriterion}\end{equation}
for some pre-specified upper bound threshold value $\alpha \in (0,1)$ called the significance level (e.g., $\alpha=0.05$). If we let $f_0$ and $f_1$ be the densities (with respect to some appropriate reference measure) of the conditional distribution of $\XX$ given that $Y=0$ and $Y=1$, respectively, and $c_\alpha$ be a constant such that $$\P(f_1(\XX) > c_\alpha f_0(\XX) \mid Y=0) \leq \alpha,$$
then the Neyman-Pearson (NP) theory \cite{Neyman33,Shao03} establishes that 
\begin{equation}  \varphi_\alpha^*(\xx) = \indic{p(\xx) > \frac{\pi_1 c_\alpha}{\pi_0+\pi_1 c_\alpha}}. \label{eqn:NPtest} \end{equation} 
This line of thought has been developed from a machine learning perspective in the sequence of papers \cite{Rigollet11, Tong2013,Tong2016,Tong2020,Li20}. Of course, (\ref{eqn:NPtest}) is again an oracle, feasible versions of which requiring estimating $p$ and/or $f_0$ and $f_1$; see \cite{Tong2016}.

\ppn In the present work, we depart from these traditional approaches and avoid prescribing an optimal classifier or test `on paper' as in (\ref{eqn:bayesclass}) and (\ref{NPcriterion}). Instead, we let a deep learning algorithm learn a discriminative score from training data, which is then calibrated into a test procedure. Accordingly, the procedure unfolds in two steps. First, the algorithm identifies a map $\phi$ that strongly discriminates between the typical values of $\XX$ associated with $Y=0$ and those associated with $Y=1$; second, we use this map as a test statistic for $H_0$ $[Y=0]$ versus $H_1$ $[Y=1]$.

\ppn Specifically, we construct the test as \( \varphi_\alpha = \Delta_\alpha \circ \phi \), where \( \Delta_\alpha \) is a decision function that returns the testing decision (reject or fail to reject \( H_0 \), or equivalently the class label \( 1 \) or \( 0 \)) on the basis of the value of \( \phi(\xx) \). Ideally, \( \Delta_\alpha \) is calibrated so as to maintain the prescribed significance level, that is, to ensure that \( R_0(\varphi_\alpha)\leq \alpha \) for any sample size. In the statistical literature, such a procedure is often called an {\it exact test} \cite[p.\,52]{Kendall71}. We explain below why such an exact calibration is typically available in our setting -- at least `nearly' in a specific sense -- in contrast with most classical procedures, which rely on asymptotic approximations and are therefore reliable only when the sample size is sufficiently large. 

\ppn Structurally \cite{DL2,Liquet24}, deep learning algorithms are designed to identify the mapping ${\phi}$ in the class 
 \begin{multline} {\mathcal{\Phi}} \doteq \bigg\{{\phi}: \Xs \to [0,1] \text{ s.t. } \exists (b_1,\ldots,b_D; \WW_1,\ldots,\WW_D), \\ {\phi}(\xx)= \sigma(b_D+\WW_D \sigma_{D-1}(b_{D-1}+\WW_{D-1} \sigma_{D-2}(\ldots  \\  \sigma_2(b_2+\WW_2 \sigma_1(b_1+\WW_1 \xx))))) \bigg\}, \label{eqn:DLmap} \end{multline} 
where $b_1,\ldots,b_D$ and $\WW_1,\ldots,\WW_D$ are, respectively, the biases and weight matrices (of appropriate sizes) of the neural network. The (fixed) number of layers $D$ is called the `depth' of the network, while $\sigma_1,\ldots,\sigma_{D-1}$ are known non-linear element-wise activation functions (e.g., ReLU). The last output layer $\sigma$ is typically a sigmoid function guaranteeing that the output of any ${\phi} \in \Phi$ is in $[0,1]$. 

\ppn For defining the `best' $\phi$ in $\Phi$ in this context, we may use for loss function $\ell$ the {\it cross-entropy} -- it is a common choice in the classification framework, both for its probabilistic foundation and for numerical stability reasons \cite{DL2}. Explicitly, this means that the optimal classifier $\phi$ is defined here as

\[\phi^\star =  \arg\min_{\phi\in \Phi} \left\{ -\E\bigl[Y\log \phi(\XX)+(1-Y)\log \bigl(1-\phi(\XX)\bigr)\bigr] \right\}. \]
This is in fact a maximum likelihood criterion: in effect, $\phi^\star$ is the best approximation in the sense of Kullback-Leibler of the function $p$ (\ref{eqn:pX}) into the class (\ref{eqn:DLmap}). Of course, $\phi^\star$ is again an oracle, as it is a feature of the joint distribution of $(\XX,Y)$. However, if we have access to a training data set $\{(\xx_i,y_i) \}_{i=1}^N$ of realisations of $(\XX,Y)$, then the optimal classifier can be learned as
\begin{equation} \widehat{\phi} =  \arg\min_{\phi\in \Phi} \left\{ - \sum_{i=1}^N \bigl[y_i \log \phi(\xx_i)+(1-y_i)\log \bigl(1-\phi(\xx_i)\bigr)\bigr] \right\}, \label{eqn:hatphiDL} \end{equation} 
an optimisation performed via deep-learning algorithms \cite{DL2}; see Section \ref{subsec:deep_learning}.

\ppn Assume further that we can generate a large number \(N'\) of additional replications of \((\XX,Y)\), independent of the previous ones, for which \(Y=0\). Equivalently, we have access to a second training set \(\{(\xx^{(0)}_{i'},0)\}_{i'=1}^{N'}\), where the vectors \(\xx^{(0)}_{i'}\) are sampled from the conditional distribution of \(\XX\) given \(Y=0\). We will refer to $\{(\xx^{(0)}_{i'},0) \}_{i'=1}^{N'}$ as the {\it null-calibration set}. We can now compute \(\widehat{\phi}\) for each of these vectors, and the empirical distribution of \(\{\widehat{\phi}(\xx_{i'}^{(0)})\}_{i'=1}^{N'}\) then provides a reconstruction of the distribution of the test statistic \(\widehat{\phi}\) under the null hypothesis. This reconstruction is {\it exact up to Monte Carlo error}, which is typically negligible when \(N'\) is sufficiently large; for this reason, we will refer to the procedure as \emph{near-exact}. It is then numerical routine to identify a critical value $d_\alpha \in [0,1]$ such that
\begin{equation}  \frac{1}{N'} \sum_{i'=1}^{N'} \indic{\widehat{\phi}(\xx_{i'}^{(0)}) > d_\alpha} \leq \alpha, \label{eqn:dalpha} \end{equation}
this being a near-exact approximation of the quantile of the conditional distribution of $\widehat{\phi}(\XX)$ given that $Y = 0$.

\ppn Finally, we define the test as $\widehat{\varphi}_\alpha = \Delta_\alpha \circ \widehat{\phi}$, where $\widehat{\phi}$ is (\ref{eqn:hatphiDL}) and $\Delta_\alpha: [0,1] \to \{0,1\}$, $\Delta_\alpha(u) = \indic{u > d_\alpha}$, with $d_\alpha$  determined by (\ref{eqn:dalpha}). Specifically, our deep-test is
\begin{equation} \widehat{\varphi}_\alpha(\xx) = \indic{\widehat{\phi}(\xx) > d_\alpha}. \label{eqn:ourtest} \end{equation}
By construction, this test controls its level under $H_0$ (i.e., $R_0(\widehat{\varphi}_\alpha) =\P(\widehat{\varphi}_\alpha(\XX) \neq Y \mid Y = 0 ) \leq \alpha$) in an near-exact manner. Moreover, its power is expected to benefit from the strong discriminative ability of the underlying deep learning classifier $\widehat{\phi}$. Indeed (\ref{eqn:ourtest}) can be thought of as a feasible version of the oracle NP-optimal test (\ref{eqn:NPtest}) where $p$ is approximated directly in (\ref{eqn:DLmap}) and the whole quantity $\pi_1 c_\alpha/(\pi_1 c_\alpha+ \pi_0 )$ is estimated at once by $d_\alpha$ from the null-calibration set. This provides a principled theoretical justification for 
the procedure.

\section{The case of dependence detection} \label{sec:deptect}

\noindent As announced in Section \ref{sec:intro}, we evaluate the proposed deep-testing procedure on the problem of detecting dependence. This is a central question in statistics and in many scientific applications, since understanding whether variables are related is often a first step toward explaining, predicting, or modelling a system. In statistics, many modelling assumptions can be expressed in terms of independence, conditional independence, or specific forms of dependence \cite{Dawid79}. In machine learning, feature selection methods often rely on assessing the relevance of input variables for predicting the output, a task closely connected to measuring statistical dependence \cite{Guyon03}. Reliable procedures for detecting dependence are therefore of broad practical interest.

\ppn Here we study our proposed deep-testing approach to testing independence between two {\it continuous}, {\it univariate} random variables.\footnote{Other cases, such as those involving discrete variables or multivariate margins, could be addressed using similar deep-testing ideas, although the relevant dependence indicators would differ from those used here; see Table \ref{tab:indicators}.} Let $(Z_1,Z_2)$ be a continuous bivariate random vector with joint distribution $F_{Z_1 Z_2}$ and marginal distributions $F_{Z_1}$ and $F_{Z_2}$. We wish to test
\begin{equation}
H_0: F_{Z_1 Z_2} = F_{Z_1}F_{Z_2}
\label{eqn:H0indep}
\end{equation}
against the alternative
\begin{equation}
H_1: F_{Z_1 Z_2} \neq F_{Z_1}F_{Z_2}.
\label{eqn:Haindep}
\end{equation}
A decision regarding (\ref{eqn:H0indep}) is to be made on the basis of an observed random sample $\Zs \doteq \{(z_{1k},z_{2k})\}_{k=1}^n$ drawn from $F_{Z_1 Z_2}$. As suggested in Section \ref{sec:pierre}, we approach this test from a classification perspective, seeking to assign the label $y=0$ or $y=1$ to the sample according to whether the distribution that generated it factorises as in (\ref{eqn:H0indep}) or not, as in (\ref{eqn:Haindep}). We consider three scenarios, which differ in the choice of input features used to represent the sample $\Zs$ to the learning algorithm.

\ppn {\bf Scenario 1:} we formulate the test purely as an image-recognition problem. If deep learning can recognise images with near-perfect accuracy, can it also distinguish an `image of a bivariate sample' formed from two independent univariate samples, from one generated by a genuinely bivariate distribution with a non-trivial dependence structure? Such an image may, for instance, be a scatter plot. Accordingly, the input \(\xx\) may be encoded either as a matrix of black-and-white pixels marking the locations of the observations in \(\mathbb{R}^2\), or as a simple grayscale density image produced by binning, analogous to a bivariate histogram.

\ppn {\bf Scenario 2:} the statistical and machine learning literature on testing, quantifying, and modelling dependence is vast. A broad range of dependence measures and independence test statistics -- hereafter referred to as {\it dependence indicators} -- has been proposed, from classical coefficients such as Pearson's, Kendall's, and Spearman's correlations \cite[Section 6.2]{DrouetMari01} to more recent machine learning tools such as the Hilbert--Schmidt independence criterion \cite{HSIC}. Because different indicators capture different aspects of dependence, they may jointly provide a rich description of the dependence structure present in a data set. Accordingly, rather than using the entire scatter plot as input as in Scenario 1, we may work directly with these indicators and thus focus on features expected to be especially informative for dependence detection. Here the input -- denoted \(\tilde{\xx}\) to distinguish it from \(\xx\) in Scenario 1 -- is the vector of observed values of these dependence indicators, each computed from the sample \(\Zs\). A detailed description of the nineteen indicators we chose to include in this representation is provided in Section \ref{subsec:depind}. 

\ppn {\bf Scenario 3:} a third natural option is to combine the previous two scenarios, so that the input features for the deep learning algorithm consist of {\it both} an image-based representation of the scatter plot {\it and} a vector of dependence indicators, as above. Thus, the input is \((\xx,\tilde{\xx})\), comprising the scatter-plot representation \(\xx\) from Scenario 1 and the vector of dependence-indicator values \(\tilde{\xx}\) from Scenario 2.

\ppn Details on the construction of the scatter-plot representation \(\xx\) are given in Section~\ref{subsec:scatinp}, while the components of the dependence-indicator vector \(\tilde{\xx}\) are described in Section~\ref{subsec:depind}. To simplify notation, we will often write \(\check{\xx}\) generically for the input representation under consideration; thus, \(\check{\xx}=\xx\), \(\check{\xx}=\tilde{\xx}\), or \(\check{\xx}=(\xx,\tilde{\xx})\) depending on the current scenario. 

\ppn In Section \ref{subsec:deep_learning}, each scenario will be translated into a different network architecture. The corresponding discriminative map $\hat{\phi}$ used in the test will then be learned as in (\ref{eqn:hatphiDL}), based on a large collection of $N$ simulated samples exhibiting a range of dependence structures and strengths, including independence -- the data-generating mechanism for these training samples is described in Section~\ref{subsec:train}. In essence, the goal is to assess the presence or absence of dependence in the observed sample $\Zs$ by comparing its representation $\check{\xx}$ with those of many simulated samples for which the dependence status is known. 

\ppn The role of the map $\hat{\phi}$ is particularly transparent in Scenario~2, where the input consists of the vector $\tilde{\xx}$ of dependence indicators. Indeed, these indicators are not equally informative in all settings. For example, it is known that Pearson's correlation coefficient $\rho$ is not a reliable general indicator of dependence, since $\rho=0$ does not imply independence -- but it sometimes is in specific cases (e.g., a bivariate Gaussian distribution). We therefore expect the deep learning algorithm to determine which indicators, or combinations thereof, are most informative for detecting dependence, and in which ranges of their values. The resulting {\it dependence score} $\hat{\phi}(\tilde{\xx})$ will thus be a highly non-linear combination of the available dependence indicators, weighted so as to optimise discriminative performance. The same interpretation applies to Scenario 1 (and therefore to Scenario 3 as well), albeit in a slightly less intuitive manner.

\ppn The proposed procedure may then be summarised as follows:

\medskip 

\begin{enumerate}
    \item \label{it:step1} Fix a large even integer $N$, and for each pre-determined sample size $n$, generate $N$ bivariate samples
    $\Zs^{(i,n)} \doteq \{(z^{(i,n)}_{1k},z^{(i,n)}_{2k})\}_{k=1}^n$, $i=1,\ldots,N$:
    $N/2$ under the null hypothesis of independence between $Z_1$ and $Z_2$ (so that $y_i=0$ for $i=1,\ldots,N/2$), and $N/2$ under alternative models exhibiting dependence (so that $y_i=1$ for $i=N/2+1,\ldots,N$) -- these are the {\it training samples}.

    \medskip

    \item For each training sample $\Zs^{(i,n)}$, construct the corresponding input feature vector $\check{\xx}_i$. The collection $\{(\check{\xx}_i,y_i)\}_{i=1}^N$ forms the {\it training set}.   

    \medskip

    \item Fit the network's probability map to this training set by solving (\ref{eqn:hatphiDL}), where $\Phi$ is the class (\ref{eqn:DLmap}) of deep-net-parameterised probability maps (that is, neural networks with a given architecture). This yields $\widehat{\phi}$.
\end{enumerate}

\ppn Now suppose that we observe a new bivariate sample $\Zs^{\mathrm{new}} \doteq \{(z^{\mathrm{new}}_{1k},z^{\mathrm{new}}_{2k})\}_{k=1}^{n^{\mathrm{new}}}$ of size $n^{\mathrm{new}}$, for which we wish to assess independence between the components. We then proceed as follows:

\medskip

\begin{enumerate}
\setcounter{enumi}{3}

    \item \label{it:nullcalib} For the sample size $n^{\mathrm{new}}$ of interest, generate $N'\in\mathbb{N}$ samples under the null hypothesis of independence ($Z_1 \indep Z_2$), and compute for each of them the corresponding input feature vector $\check{\xx}_{i'}^{(0)}$, $i'=1,\ldots,N'$. The collection $\{(\check{\xx}_{i'}^{(0)},0)\}_{i'=1}^{N'}$ (or simply $\{\check{\xx}_{i'}^{(0)}\}_{i'=1}^{N'}$) is the {\it null-calibration set}.

    \medskip

    \item \label{it:crit} For the significance level $\alpha\in (0,1)$, compute $d_\alpha\in[0,1]$, the empirical $(1-\alpha)$-quantile of $\{\widehat{\phi}(\check{\xx}_{i'}^{(0)})\}_{i'=1}^{N'}$, as in (\ref{eqn:dalpha}).

    \medskip

    \item  Compute $\check{\xx}^{\mathrm{new}}$, the input feature vector describing $\Zs^{\mathrm{new}}$. Set
    \[
    \widehat{\varphi}_\alpha(\check{\xx}^{\mathrm{new}})
    =
    \indic{\widehat{\phi}(\check{\xx}^{\mathrm{new}})>d_\alpha}.
    \]
    Reject $H_0$ if $\widehat{\varphi}_\alpha(\check{\xx}^{\mathrm{new}})=1$; otherwise, do not reject $H_0$.
\end{enumerate}

\section{Dependence and Dependence Indicators} \label{sec:litrev}

\subsection{Selection of indicators and margin-freeness} \label{subsec:prelim}

One of the most intriguing features of deep learning algorithms, and one that appears to play an important role in their empirical success, is their ability to exploit highly overparameterized and potentially redundant representations without necessarily suffering substantial losses in predictive performance \cite{DL,Belkin19,Bartlett20,Bartlett21,AllenZhu19}. In the present context, this suggests that the input representation may contain a large amount of information, and that even highly detailed feature constructions may still support good performance on previously unseen samples. In particular, using the entire scatter plot as input representation $\xx$ (Scenario 1) need not be detrimental to the procedure. Similarly, the vector $\tilde{\xx}$ of dependence indicators considered in Scenario 2 should be chosen so as to be as rich as possible.

\ppn At the same time, the statistical and machine learning literature on describing and quantifying dependence within a random vector is extraordinarily broad. It is therefore hardly feasible to make $\tilde{\xx}$ an exhaustive collection of all dependence measures and independence tests proposed to date. We therefore made principled choices guided by two criteria: 
\begin{enumerate}[(a)]
\item a theoretical criterion, namely that the selected indicators should satisfy desirable properties such as those discussed in \cite{HellCor,Renyi}; 
\item a practical criterion, namely that they should be straightforward to implement and compute, or readily available in standard software packages.
\end{enumerate}

\ppn For continuous random variables $Z_1$ and $Z_2$, a fundamental requirement for a meaningful dependence measure is that it should depend only on the copula $C_{12}$ of $(Z_1,Z_2)$. The copula $C_{12}$ is the joint distribution of the vector $(F_{Z_1}(Z_1),F_{Z_2}(Z_2))$; by the Probability Integral Transform, both $F_{Z_1}(Z_1)$ and $F_{Z_2}(Z_2)$ are uniformly distributed on $[0,1]$. The copula is meant to capture all features of $(Z_1,Z_2)$ that are free of the marginal distributions, that is, invariant under strictly increasing transformations of $Z_1$ and $Z_2$ \cite{SchW}. In particular, any dependence measure $\Delta$ satisfying
\begin{equation} \Delta(Z_1,Z_2)=\Delta(\psi_1(Z_1),\psi_2(Z_2)) \label{eqn:marginfree} \end{equation}
for all strictly increasing functions $\psi_1$ and $\psi_2$ can be represented as a functional of $C_{12}$; conversely, any dependence measure defined as a functional of the copula is margin-free in the sense of (\ref{eqn:marginfree}).

\ppn Intuitively, dependence is a margin-free concept, indeed. Each marginal distribution, \(F_{Z_1}\) or \(F_{Z_2}\), describes the behaviour of one variable in isolation and is defined without reference to any joint structure. In fact, the marginal distribution of $Z_1$ is the same object whether $Z_1$ is studied as a single variable on its own or as one component of a vector $(Z_1,Z_2)$. Therefore, $F_{Z_1}$ cannot by itself contain information about the existence of $Z_2$, let alone about the dependence structure between the two. Hence, amending the marginals individually through \(\psi_1\) and \(\psi_2\) should not affect \(\Delta\). This intuition was made precise in \cite{HellCor}, who showed that dependence measures that are not margin-free in the sense of (\ref{eqn:marginfree}) necessarily violate data-processing-type inequalities \cite[Section 2.8]{Cover06}. In this setting, such inequalities express the indisputable principle that the dependence cannot be made stronger by diluting the variables in white noise.

\ppn Some popular dependence indicators are not margin-free and are therefore affected by the above incongruity. To mitigate this problem, we consider only copula-based versions of such indicators. For example, Pearson's correlation coefficient $\rho(Z_1,Z_2)$ is not copula-based and hence is not margin-free; accordingly, we do not include it in that form among our indicators. Instead, we use the correlation coefficient between $F_{Z_1}(Z_1)$ and $F_{Z_2}(Z_2)$, that is, the correlation coefficient of the copula $C_{12}$, which coincides with Spearman's $\rho_S$. Similarly, for distance correlation \cite{dcor} and HSIC \cite{hsic}, we consider only their copula-based versions; see below. By contrast, indicators that are inherently copula-based, such as Kendall's $\tau$, mutual information, and the Hellinger correlation, are used in their standard form. 

\ppn For the same reason, the scatter plot used in Scenarios~1 and 3 is the scatter plot at the copula level, that is, a representation of the distribution of \((F_{Z_1}(Z_1),F_{Z_2}(Z_2))\) rather than that of \((Z_1,Z_2)\) itself. In other words, the image-based input $\xx$ considered in Scenarios 1 and 3 is also margin-free and captures only the dependence structure.

\ppn Now, the marginal distributions $F_{Z_1}$ and $F_{Z_2}$ are typically unknown in practice, so we will also treat them as unknown here. The copula transformation
\begin{equation} (Z_1,Z_2)\mapsto \bigl(F_{Z_1}(Z_1),F_{Z_2}(Z_2)\bigr) \label{eqn:coptrans} \end{equation} 
is therefore infeasible and must be replaced by an empirical counterpart based on the observed bivariate sample $\{(z_{1k},z_{2k})\}_{k=1}^n$. It is customary \cite{Deheuvels79} to do so using the empirical distribution functions
\begin{equation}
\widehat{F}_{Z_j;n}(z)=\frac{1}{n+1}\sum_{k=1}^n \indic{z_{jk}\le z},
\qquad (j=1,2).
\label{eqn:empcdf}
\end{equation}
The quantity
\[
r(z_{j\ell}) \doteq \sum_{k=1}^n \indic{z_{jk}\le z_{j\ell}}
\]
is the {\it rank} of $z_{j\ell}$, for $j=1,2$ and $\ell=1,\ldots,n$, that is, its position in the ordered sample $\{z_{jk}\}_{k=1}^n$. Hence the corresponding {\it pseudo-sample} associated with the copula $C_{12}$ is
\begin{equation}
\left\{\left(\widehat{F}_{Z_1;n}(z_{1k}),\widehat{F}_{Z_2;n}(z_{2k})\right)\right\}_{k=1}^n
=
\left\{\left(\frac{r(z_{1k})}{n+1},\frac{r(z_{2k})}{n+1}\right)\right\}_{k=1}^n.
\label{eqn:samprank}
\end{equation}
This explains why copula-based indicators are empirically rank-based. The normalization by $n+1$ rather than $n$ in (\ref{eqn:empcdf}) simply ensures that the pseudo-observations lie strictly between 0 and 1, thereby avoiding unnecessary technical complications for some indicators.

\subsection{The scatter-plot input $\xx$} \label{subsec:scatinp}

To construct the input vector $\xx$ used in Scenario~1 (and also in Scenario~3), each sample is first transformed into rank-based pseudo-observations using~\eqref{eqn:empcdf}, and is then converted into a greyscale image through a bivariate histogram representation. More specifically, for each sample of size $n$, $\mathcal{Z}=\{(z_{1,k},z_{2,k})\}_{k=1}^n$, we first compute the rank-based pseudo-observations (\ref{eqn:samprank}). We then partition the unit square $[0,1]^2$ into a $25\times25$ regular grid of rectangular bins and count, for each bin, the number of pseudo-observations $\left(\widehat{F}_{Z_1;n}(z_{1k}),\widehat{F}_{Z_2;n}(z_{2k})\right)$ falling into it. This produces a matrix $A\in\mathbb{R}^{25\times25}$, whose entries record the bin counts and may therefore be interpreted as grey-intensity values in an image representation of the sample. We next flatten $A$ in column-major order into
\[
\xx=(A_{1,1},A_{2,1},\dots,A_{25,25})^\top\in\mathbb{R}^{625},
\]
and normalize by its largest entry $\xx\leftarrow \xx/\max(\xx)$, thus obtaining the final image descriptor in $[0,1]^{625}$, which is fed to the learning algorithm.

\begin{figure}
	\begin{center}
		\includegraphics[width=0.45\textwidth]{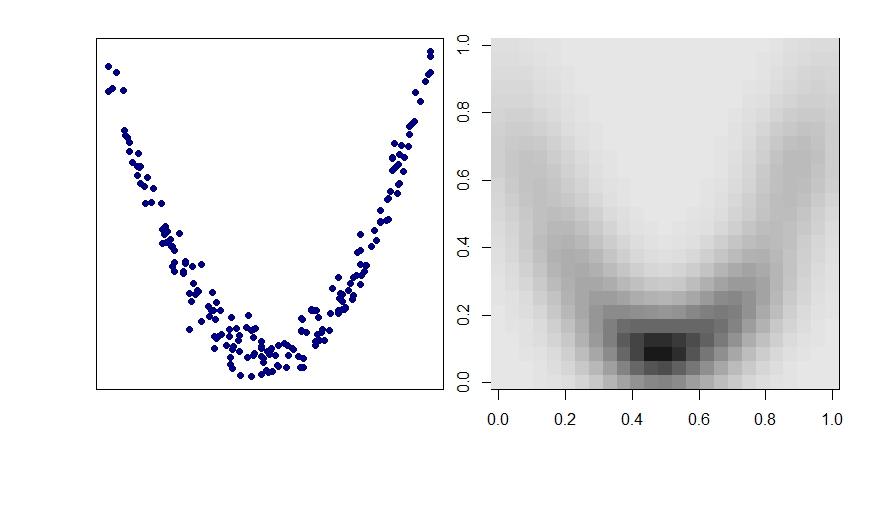}
	\end{center}
	\caption{Bivariate histogram representation (right) of a typical `parabola' sample of size $n=400$ (left), obtained on an equispaced $25\times25$ grid.} \label{fig:image}
\end{figure}

\ppn As an illustration, Figure~\ref{fig:image} shows the resulting image representation for a typical `parabola' sample of size $n=400$. We use a $25\times25$ equispaced grid as a compromise between visual resolution and computational cost. This resolution is comparable to that of the MNIST handwritten-digit images, which are stored as $28\times28$ grayscale images \cite{Lecun98}. Coarser grids may blur dependence patterns, whereas finer grids increase the input dimension and may lead to sparser representations. We leave the question of tuning the image resolution for future work.

\subsection{The dependence indicators input $\tilde{\xx}$} \label{subsec:depind}

Here we list the nineteen indicators which we include in the feature vector $\tilde{\xx}$ for Scenario $2$ (and also in Scenario~3) based on the two criteria (a) and (b) described in Section \ref{subsec:prelim}; see Table~\ref{tab:indicators} for an overview. We give only a concise description of the indicator and how to compute it on an observed sample. Details are omitted here as we refer to the existing literature.  %for a list of the associated R functions and our R package \pkg{depstats} %\url{https://github.com/ivanmyzou/depstats} for the computing codes implementing these measures.

\ppn First, we consider the `historical' copula-based correlations: Spearman's $\rho_S$, Kendall's $\tau$ and Blomqvist's $\beta$. We note that these three indices are not {\it dependence} measures {\it per se} -- in particular, they may take a value of 0 even in the presence of dependence. Rather, they measure {\it concordance} \cite[Section 5]{Geenens23}, that is, the extent up to which large (small) values of $Z_1$ go together with large (small) values of $Z_2$ (or vice-versa). Yet, a large (absolute) value of any of those indices is a clear indication of the presence of dependence, and including them in our list of features is legitimate. More details about these parameters may be found in \cite[Section 5.1]{Nelsen06} and \cite[Section 2.12]{Joe14}.

\medskip 

\subsubsection{Spearman's $\rho_S$} this is Pearson's correlation between $F_{Z_1}(Z_1)$ and $F_{Z_2}(Z_2)$, which can be seen to be 
\[\rho_S = 12\iint_{[0,1]^2} uvdC_{12}(u,v) -3. \]
Also called Spearman's rank correlation, its empirical estimate can be shown to reduce to 
\[\widehat{\rho}_S =  1- \frac{6 \sum_{k=1}^n \left(r(z_{1k})-r(z_{2k}) \right)^2}{n(n^2-1)}.\]

\medskip

\subsubsection{Kendall's $\tau$} independent pairs $(Z_{1k},Z_{2k}),(Z_{1k'},Z_{2k'}) \sim F_{Z_1 Z_2}$ are in concordance if $(Z_{1k}-Z_{1k'})(Z_{2k}-Z_{2k'}) > 0 $ and in discordance if $(Z_{1k}-Z_{1k'})(Z_{2k}-Z_{2k'}) < 0 $ (the case $(Z_{1k}-Z_{1k'})(Z_{2k}-Z_{2k'}) = 0 $ is ruled out by the assumed continuity of $(Z_1,Z_2)$). Kendall's $\tau$ is
\begin{multline*} \tau = \mathbb{P}\left((Z_{1k}-Z_{1k'})(Z_{2k}-Z_{2k'}) > 0\right) \\ - \mathbb{P}\left((Z_{1k}-Z_{1k'})(Z_{2k}-Z_{2k'}) < 0\right), \end{multline*}
the difference between probabilities of concordance and discordance. In terms of the copula of $(Z_1,Z_2)$, this amounts to
\[\tau = 4\iint_{[0,1]^2} C_{12}(u,v)dC_{12}(u,v) -1. \]
Empirically, $\tau$ is estimated by 
\[ \widehat{\tau} = \widehat{p}_C - \widehat{p}_D,\]
where $\widehat{p}_C = {{n\choose 2}}^{-1}\sum_{k=1}^n \sum_{k' \neq k} \indic{(z_{1k}-z_{1k'})(z_{2k}-z_{2k'}) > 0}$ and $\widehat{p}_D = {{n\choose 2}}^{-1}\sum_{k=1}^n \sum_{k' \neq k} \indic{(z_{1k}-z_{1k'})(z_{2k}-z_{2k'}) < 0}$ are the fraction of pairs of observations in concordance and discordance, respectively, in the sample. Note that this, indeed, only depends on the ranks defined earlier (e.g., $z_{1k}-z_{1k'} > 0 \iff r(z_{1k})-r(z_{1k'}) > 0$).

\medskip

\subsubsection{Blomqvist's $\beta$} Let $m_1$ and $m_2$ be the medians of the marginal distributions $F_{Z_1}$ and $F_{Z_2}$, respectively. Then, Blomqvist $\beta$, also called {\it medial correlation coefficient}, is defined as
\[\beta = \P\left\{(Z_1-m_1)(Z_2-m_2) >0\right\} - \P\left\{(Z_1-m_1)(Z_2-m_2) <0\right\}, \]
which can be shown to be
\[\beta  = 4C_{12}(1/2,1/2)-1.\]
%(Again, this only depends on the ranks -- if $C_{12}$ is the copula of $(Z_1,Z_2)$, then it can be shown that $\beta = 4C_{12}(1/2,1/2)-1$.)
Now, denote $\widehat{m}_j$ ($j=1,2$) the empirical medians of the observed samples $\{z_{jk}\}_{k=1}^n$. Then $\beta$ can be estimated empirically by 
\[\widehat{\beta} = \frac{1}{n} \sum_{k=1}^n \indic{(z_{1k}-\widehat{m}_1)(z_{2k}-\widehat{m}_2) >0} - \frac{1}{n} \sum_{k=1}^n \indic{(z_{1k}-\widehat{m}_1)(z_{2k}-\widehat{m}_2) <0}. \]
Note that this, again, only depends on the ranks (e.g., $z_{1k}-\widehat{m}_1 > 0 \iff r(z_{1k}) > n/2$).
%If we think of $(\widehat{m}_1,\widehat{m}_2)$ dividing the plane $(Z_1,Z_2)$ into 4 quadrants, then $\widehat{\beta}$ is the difference between the fraction of observations falling in the first and third quadrants ($Z_1$ and $Z_2$ both positive or negative) and the fraction of observations falling in the second and fourth quadrant ($Z_1$ and $Z_2$ 

\normalsize

\ppn A less standard alternative is:

\subsubsection{Area Under Kendall Plot} The Kendall distribution \cite{Nelsen03} is $K(w) = \P(F_{Z_1Z_2}(Z_1,Z_2) \leq w)$, the distribution of the univariate variable $F_{Z_1Z_2}(Z_1,Z_2)$ (akin to a bivariate Probability Integral Transform). When $Z_1$ and $Z_2$ are independent, then it reduces down to $K_0(w) = \P(UV \leq w)$, for $U,V$ two independent standard Uniform random variables; which can be worked out as $K_0(w) = w -w \log w$, $w \in (0,1]$. Inspired by the classical $PP$-plot (visual tool for checking normality of an observed univariate sample), \cite{Kplot} proposed to plot the quantiles of an empirical estimate of $K$ against the theoretical quantiles of $K_0$ (`$K$-plot') -- any lack of linearity would indicate non-independence. Departure from such linearity (hence from independence) can be quantified by the `Area under $K$-plot' (AUK), similarly to what is commonly done for analysing Receiver Operating Characteristic (ROC) curve in biostatistics. It was shown in \cite{AUK} that 
\begin{multline*} \text{AUK} = \P\left(F_{Z_1}(Z_1)F_{Z_2}(Z_2) < F_{Z_1Z_2}(Z_1,Z_2) \right) \\ = \P\left(UV < C_{12}(U,V) \right),\end{multline*}
an empirical estimate of which being easily obtained:
\[\widehat{\text{AUK}} = \frac{1}{n} \sum_{k=1}^n \indic{r(z_{1k}) r(z_{2k}) < (n+1)\sum_{\ell=1}^n \indic{r(z_{1\ell}) \leq r(z_{1k})}\indic{r(z_{2\ell}) \leq r(z_{2k})}}. \]
% Alternatively, one may compute the statistic
% \[
% \left(\widehat{\text{AUK}}^{(F)} - \tfrac{1}{2}\right)^2,
% \]
% where \(\widehat{\text{AUK}}^{(F)}\) is the empirical area under the Kendall curve, as implemented in the \texttt{testforDEP::AUK()} function. This formulation slightly differs from the rank-based estimator \(\widehat{\text{AUK}}\) above, though both aim to measure deviations from independence, for which the value is \(\tfrac{1}{2}\).

\ppn Other natural measures of dependence are distances between (empirical estimates of) $F_{Z_1Z_2}$ and the independence base case $F_{Z_1}F_{Z_2}$. Hoeffding's $D$, the Hellinger correlation and the Mutual Information belong to this class.

\normalsize

\medskip

\subsubsection{Hoeffding's $D$} this is the (weighted) $L_2$-distance between $F_{Z_1Z_2}$ and $F_{Z_1}F_{Z_2}$, viz.
\begin{align*} D & = \iint_{\R^2} \left(F_{Z_1Z_2}(z_1,z_2) -F_{Z_1}(z_1)F_{Z_2}(z_2)\right)^2\,dF_{Z_1Z_2}(z_1,z_2) \\ 
& = \iint_{[0,1]^2} \left(C_{12}(u,v)- uv \right)^2 dC_{12}(u,v).\end{align*}
Substituting the empirical versions of the distribution functions in $D$, we may show \cite{Hoef} that 
\[\widehat{D} = 30\frac{(n-2)(n-3)\widehat{D}_1 + \widehat{D}_2 - 2(n-2)\widehat{D}_3}{n(n-1)(n-2)(n-3)(n-4)} \]
is a consistent estimator of $D$, with $Q_k = \sum^n_{\ell=1} \indic{z_{1\ell} < z_{1k}} \indic{z_{2\ell} < z_{2k}}$ and
\begin{align*}
\widehat{D}_1 &= \sum^n_{k=1} Q_k (Q_k - 1) ,
\\
\widehat{D}_2 &= \sum^n_{k=1} \big[ r(z_{1k}) - 1 \big] \big[ r(z_{1k}) - 2 \big] \big[ r(z_{2k}) - 1 \big] \big[r(z_{2k}) - 2 \big] ,
\\
\widehat{D}_3 &= \sum^n_{k=1} \big[ r(z_{1k}) - 2 \big] \big[ r(z_{2k}) - 2 \big] Q_k.
\end{align*}

\normalsize

\subsubsection{Hellinger Correlation} this is based on the (squared) Hellinger distance between $F_{Z_1Z_2}$ and $F_{Z_1}F_{Z_2}$, viz.
\begin{align*} \Hs^2 & = \frac{1}{2} \iint_{\R^2} \left(\sqrt{f_{Z_1Z_2}}(z_1,z_2) -\sqrt{f_{Z_1}}(z_1)\sqrt{f_{Z_2}}(z_2)\right)^2 \,dz_1 \,dz_2\\ 
& = 1- \iint_{[0,1]^2} \sqrt{c_{12}}(u,v) \,du\,dv,\end{align*}
where $f_{Z_1Z_2}$, $f_{Z_1}$, $f_{Z_2}$ and $c_{12}$ are the densities of $F_{Z_1Z_2}$, $F_{Z_1}$, $F_{Z_2}$ and $C_{12}$, respectively (assuming existence thereof). Note that $\Bs \doteq \iint_{[0,1]^2} \sqrt{c_{12}}(u,v) \,du\,dv$ is the {\it Bhattacharyya affinity} between $c_{12}$ and the independence copula density (identically equal to 1 on $[0,1]^2$). The Hellinger correlation~$\eta$ is a one-to-one function of $\Bs$ which facilitates interpretation. Empirically, \cite{HellCor} showed that a basic estimator of $\Bs$ is
\begin{equation} \widehat{\Bs} = \frac{2\sqrt{n-1}}{n(n+1)} \sum_{k=1}^n \min_{\ell \neq k} \sqrt{\left(r(z_{1k})-r(z_{1\ell})\right)^2 + \left(r(z_{2k})-r(z_{2\ell})\right)^2}, \label{eqn:Bhat} \end{equation}
of which an empirical estimator of $\eta$ follows. (Note that \cite{HellCor} actually recommended a refined version of $\widehat{\Bs}$, involving shrinking, for improved performance. In our computations we use that refined version.).

\medskip

\subsubsection{Mutual Information} this is the Kullback-Leibler divergence (a.k.a.\ {\it relative entropy} in information theory \cite[Section 2.3]{Cover06}) between $F_{Z_1Z_2}$ and $F_{Z_1}F_{Z_2}$, viz. 
\begin{align*} \Is & = \iint_{\R^2} f_{Z_1Z_2}(z_1,z_2) \log \left(\frac{f_{Z_1Z_2}(z_1,z_2)}{f_{Z_1}(z_1) f_{Z_2}(z_2)} \right)  \,dz_1 \,dz_2\\ 
& = \iint_{[0,1]^2} c_{12}(u,v) \log c_{12}(u,v)  \,du \,dv.\end{align*}
To avoid infinite values (which is possible for $\Is$), we have adopted here a normalised version known as the {\it Linfoot correlation} \cite{infocor}, specifically
\[\Is^* = \sqrt{1-\exp(-2\Is)} \in [0,1]. \]
Empirical estimation of $\Is$ (and therefore $\Is^*$) for continuous variables has proved challenging. Recently, \cite{Berrett19} extended initial ideas from \cite{Kozachenko87} and proposed a nearest-neighbour-based estimator, similar in spirit to (\ref{eqn:Bhat}). This is the estimator we have used here.
% which maps mutual information to the interval $[0,1]$. Empirical estimation of $\Is$ (and hence $\Is^*$) is carried out using a refined nearest‐neighbour approach. In our implementation, the data are first transformed into pseudo‐observations on $[0,1]^2$ via a rank transform (using, e.g., \texttt{srank()}). A mutual information estimate is then obtained using the estimator proposed in \cite{Berrett19}---an extension of the ideas introduced by \cite{Kozachenko87}---. Since mutual information is theoretically nonnegative, any negative estimates due to finite-sample or numerical issues are set to zero. Finally, the raw estimate is transformed into the Linfoot correlation as defined above. Although closed-form expressions exist for simpler nearest‐neighbour estimators, the refined estimator used here is best described by its algorithmic procedure rather than by a succinct mathematical formula.

\medskip

\subsubsection{Maximal Information Criterion} In an attempt to get around the above-mentioned difficulty for estimating $\Is$ in a continuous setting, one can bin the data; i.e., overlay a rectangular grid on the scatter-plot of $\{(z_{1k},z_{2k})\}_{k=1}^n$ and assign each observed value to the column and row bin $\tilde{z}_1/\tilde{z}_2$ in which it falls. Then $\Is$ can be approximated by
\[\widetilde{\Is}_{\tilde{\bm{z}}} = \sum_{\tilde{z}_1,\tilde{z}_2} \hat{p}_{\tilde{z}_1,\tilde{z}_2} \log \frac{\hat{p}_{\tilde{z}_1,\tilde{z}_2}}{\hat{p}_{\tilde{z}_1} \hat{p}_{\tilde{z}_2}}, \]
where $\tilde{\bm{z}}$ describes the grid and $\hat{p}_{\tilde{z}_1,\tilde{z}_2}$ is the fraction of data points falling into bin $(\tilde{z}_1,\tilde{z}_2)$ (and same for the marginals).
The Maximal Information Criterion \cite{MIC} is defined as 
\begin{align*}
\text{mic} = \max_{\tilde{\bm{z}}} \Big\{ \frac{\widetilde{\Is}_{\tilde{\bm{z}}}}{\log_2 \min\{ \tilde{n}_1,\tilde{n}_2 \} } \Big\} ,
\end{align*}
where the maximum is taken over all possible partitions (in principle), and $\tilde{n}_1$, $\tilde{n}_2$ are the number of column and row bins in $\tilde{\bm{z}}$. Note that mic is mostly an explanatory tool (it does not correspond to any population parameter), and has been criticised for that reason \cite{Kinney14}.\footnote{See also the response \cite{Reshef16}.}

\medskip 

\ppn We add to this list the `Maximum Correlation Coefficient' and some of its variants.

\medskip

\subsubsection{Maximum Correlation Coefficient} This is 
\[R = \sup_{\psi_1,\psi_2} |\rho(\psi_1(Z_1),\psi_2(Z_2))|\]
where $\rho$ is Pearson's correlation and the supremum is taken over all measurable functions $\psi_j \in \Ls^2(F_j)$, $j=1,2$ \cite{Gebelein41,Renyi}. This makes the empirical estimation of $R$ a difficult problem \cite{Maxcor}. Here we approximate it by the {\it Alternating Conditional Expectations} algorithm, as suggested in \cite{ACE}.

\medskip

\subsubsection{Randomised Dependence Coefficient} This was proposed in \cite{randcor} as another attempt to approximate numerically the above Maximum Correlation Coefficient, based on random non-linear projections.\footnote{Note that it is {\it not} a consistent estimator of $R$, though.} Fix $P \in \N$ and for $p =1 ,\ldots,P$  generate random parameters $W_p \in \mathbb{R}$ and $B_p \in \mathbb{R}$. Define the $(n \times (d\cdot P))$-matrix
\begin{align*}
\bm{\Phi}_1= 
\begin{pmatrix}
f(W_1 r(z_{11}) + B_1) & \hdots & f(W_P r(z_{11}) + B_P)
\\
\vdots & \ddots & \vdots
\\
f(W_1 r(z_{1n}) + B_1) & \hdots & f(W_P r(z_{1n}) + B_P)
\end{pmatrix},
\end{align*} 
(and similarly for $\bm{\Phi}_2$ from the ranks $\{r(z_{2k})\}_{k=1}^n$) with $f$ a ($d$-dimensional vector-valued) projection function. One may consider such matrix as $n$ observations of a $(d\cdot P)$--dimensional random vector. The randomised dependence coefficient is then defined as
\begin{align*}
\widehat{r} = \sup_{(\bm{\alpha},\bm{\beta}) \in \R^{(d\cdot P)} \times \R^{(d\cdot P)}} \widehat{\rho}( \bm{\Phi}_1 \bm{\alpha} , \bm{\Phi}_2 \bm{\beta}),
\end{align*} 
where $\widehat{\rho}$ is empirical Pearson's correlation. This is equivalent to computing the canonical correlation \cite{Hardle07} between two samples of random vectors. As suggested in \cite{randcor}, we set $d=2$ and $f(\cdot) = (\sin(\cdot), \cos(\cdot))$, $P=10$ and draw $W_1,...,W_{10}$ independently from $N(0,s)$ (with standard deviation $s=0.5$) and $B_1,...,B_{10}$ independently from $\textrm{Unif}(-\pi,\pi)$.

\ppn Another approach is to assess departure from independence through the characteristic functions. This is what the following `Distance correlation' and `Martingale difference correlation' are based on.

\medskip

\subsubsection{(Copula) Distance Correlation} Denote $\phi_1(t) = \E(e^{\im tZ_1})$, $\phi_2(s) = \E(e^{\im sZ_2})$ and $\phi_{12}(t,s)= \E(e^{\im (tZ_1 + sZ_2)})$ the marginal and joint characteristic functions of $Z_1$, $Z_2$ and $(Z_1,Z_2)$, respectively. Then distance covariance \cite{dcor} is defined as
\begin{align*}
\text{dCov}(Z_1,Z_2) = \bigg( \frac{1}{\pi^2} \int \frac{ \big| \phi_{12} (t,s) - \phi_{1} (t) \phi_{2} (s) \big|^2 }{t^2 s^2} \, ds \, dt \bigg)^{\frac{1}{2}},
\end{align*}
and its normalised version, distance correlation, as
\[\text{dCor}(Z_1,Z_2) =  \frac{\text{dCov}(Z_1,Z_2)}{\sqrt{\text{dCov}(Z_1,Z_1)}\sqrt{\text{dCov}(Z_2,Z_2)}}.\]
Essentially a weighted $L_2$-distance between $\phi_{12}$ and the independence base case $\phi_1 \phi_2$, the name `distance correlation' originates from its empirical version, which can be seen to be analogous to Pearson's correlation computed on pairwise distances between the observations. Specifically, if for $j=1,2$ define 
\begin{align*}
a^{(j)}_{k\ell} &= \big|z_{jk} - z_{j\ell} \big|  ;
\\
\bar{a}^{(j)}_{k \cdot} = \frac{1}{n} \sum^n_{\ell=1} a^{(j)}_{k\ell} \, ;
\, \, \, \,
\bar{a}^{(j)}_{\cdot \ell} &= \frac{1}{n} \sum^n_{k=1} a^{(j)}_{k\ell} \, ;
\, \, \, \, 
\bar{a}^{(j)}_{\cdot \cdot} = \frac{1}{n^2} \sum^n_{k=1} \sum^n_{\ell=1} a^{(j)}_{k\ell} \, ;
\end{align*}
and
\begin{equation} A^{(j)}_{k\ell} = a^{(j)}_{k\ell} - \bar{a}^{(j)}_{k \cdot} - \bar{a}^{(j)}_{\cdot \ell} + \bar{a}^{(j)}_{\cdot \cdot}, \label{eqn:Akl} \end{equation}
then the empirical distance covariance is
\begin{equation*} \widehat{\text{dCov}}_n(Z_1,Z_2) = \left(\frac{1}{n^2} \sum^n_{k=1} \sum^n_{\ell=1} A^{(1)}_{k\ell} A^{(2)}_{k\ell}\right)^{1/2} \label{eqn:dcov} \end{equation*}
and the empirical distance correlation
\begin{equation} \widehat{\text{dCor}}_n(Z_1,Z_2) = \frac{\widehat{\text{dCov}}_n(Z_1,Z_2)}{\sqrt{\widehat{\text{dCov}}_n(Z_1,Z_1)}\sqrt{\widehat{\text{dCov}}_n(Z_2,Z_2)}}. \label{eqn:dcor} \end{equation}
As indicated in Section \ref{subsec:prelim}, though, dcor is {\it not} margin-free. Hence here we consider only the copula version of it, viz.\ dCor$(F_{Z_1}(Z_1),F_{Z_2}(Z_2))$, and its empirical counterpart (\ref{eqn:dcor}) computed on the sample of ranks (\ref{eqn:samprank}).

%Implementation of such is available with function \texttt{dcor()} in the package \texttt{energy} \cite{Renergy}. Distance correlation is not marginal--free so instead of applying the estimate to the samples, we apply such to the sample ranks $\big( r(\bm{x}),r(\bm{y}) \big)$ and $\big( r(\bm{y}),r(\bm{x}) \big)$.

\medskip

\subsubsection{(Copula) Martingale Difference Correlation} This is an extension of the above distance covariance/correlation \cite{Martdiff}. The Martingale Difference Covariance is defined as
\[\text{MDD}(Z_2|Z_1) = \bigg( \frac{1}{\pi} \int \frac{ \big| g_{2,1} (s) - g_2 \phi_1(s) \big|^2 }{s^2} \, ds \bigg)^{\frac{1}{2}} \]
where $g_{2,1} (s) = \E(Z_2 e^{\im Z_1 s})$, $g_2 = \E(Z_2)$ and $\phi_1(s) = \E(e^{\im Z_1 s})$. The normalised Martingale Difference Correlation is
\[ \text{MDC}(Z_2|Z_1) = \frac{\text{MDD}(Z_2|Z_1)}{\sqrt{\var(Z_2)} \sqrt{\text{dCov}(Z_1,Z_1)}}. \]
An empirical estimator of this is given by 
\[\widehat{\text{MDD}}_n(Z_2|Z_1) = - \frac{1}{n^2}   \sum^n_{k=1} A^{(1)}_{k \ell} B^{(2)}_{k\ell}, \]
where $B^{(2)}_{k\ell} = b^{(2)}_{k\ell} - \bar{b}^{(2)}_{k \cdot} - \bar{b}^{(2)}_{\cdot \ell} + \bar{b}^{(2)}_{\cdot \cdot}$, and 
\begin{align*}
b^{(2)}_{k\ell} &= z_{2k}z_{2\ell}  ;
\\
\bar{b}^{(2)}_{k \cdot} = \frac{1}{n} \sum^n_{\ell=1} b^{(2)}_{k\ell} \, ;
\, \, \, \,
\bar{b}^{(2)}_{\cdot \ell} &= \frac{1}{n} \sum^n_{k=1} b^{(2)}_{k\ell} \, ;
\, \, \, \, 
\bar{b}^{(2)}_{\cdot \cdot} = \frac{1}{n^2} \sum^n_{k=1} \sum^n_{\ell=1} b^{(2)}_{k\ell} \, ;
\end{align*}
and $A^{(1)}_{k \ell}$ is defined as in (\ref{eqn:Akl}). The empirical Martingale Difference Correlation follows as
\[ \widehat{\text{MDC}}_n(Z_2|Z_1) = \frac{\widehat{\text{MDD}}_n(Z_2|Z_1)}{\widehat{S}_{2n} \sqrt{\widehat{\text{dCov}}_n(Z_1,Z_1)}},  \]
where $\widehat{S}_{2n}$ is the empirical standard deviation of the sample $\{z_{2k} \}_{k=1}^n$. Like dCor, MDC is not margin-free. Besides, it is {\it not} symmetric in $(Z_1,Z_2)$ ($\text{MDD}(Z_2|Z_1) \neq \text{MDD}(Z_1|Z_2)$, in general). So here we use the symmetrised copula version of it, specifically: 
\[MDC = \frac{1}{2} \left(\text{MDC}(F_{Z_2}(Z_2)|F_{Z_1}(Z_1)) + \text{MDC}(F_{Z_1}(Z_1)|F_{Z_2}(Z_2))\right), \]
and its empirical counterpart computed on the ranks.

\ppn In fact, dCor is a particular case of the following indicator which lies within the framework of reproducing kernel Hilbert spaces (RKHS).

\medskip

\subsubsection{(Copula) Hilbert-Schmidt Independence Criterion} For $j=1,2$, associate to $Z_j$ the RKHS $\Fs_j$ consisting of mappings $f_j$ from $\R$ to $\R$ equipped with its scalar product $\langle \cdot , \cdot \rangle_{\Fs_j}$. Define the cross-covariance operator $\Cs_{12}: \Fs_2 \to \Fs_1$ as the one verifying
\[\langle f_1, \Cs_{12}(f_2) \rangle_{\Fs_1} = \cov(f_1(Z_1),f_2(Z_2))\] 
for all $(f_1,f_2) \in \Fs_1 \times \Fs_2$. The HSIC criterion \cite{depHS,hsic,HSIC} is the squared Hilbert--Schmidt norm of $\Cs_{12}$:
\begin{align*}
\text{HSIC}(Z_1,Z_2) = \| \Cs_{12}  \|^2_{\text{HS}}. 
\end{align*} 
If $\Fs_1$ and $\Fs_2$ are the RKHS induced by {\it characteristic} kernels~$k_1$ and $k_2$, respectively, then 
\begin{multline*} \text{HSIC}(Z_1,Z_2) =  \E\left(k_1(Z_1,Z_1')k_2(Z_2,Z_2') \right) \\ + \E\left(k_1(Z_1,Z_1')\right) \E\left(k_2(Z_2,Z_2') \right)  \\
-2\E\left(\E\left(k_1(Z_1,Z_1')|Z_1 \right)\E\left(k_2(Z_2,Z_2')|Z_2 \right)\right), \end{multline*} 
where $(Z_1',Z_2')$ is an independent copy of $(Z_1,Z_2)$. This allows easy empirical estimation of HSIC; viz.
\[\widehat{\text{HSIC}}_n(Z_1,Z_2) = \frac{1}{(n-1)^2} \text{trace} \big( \bm{K}_1 \bm{H} \bm{K}_2 \bm{H} \big) \]
where $\bm{H},\bm{K}_1,\bm{K}_2$ are $(n \times n)$-matrices defined as
\begin{align*}
\bm{K}_1 = \big[ k_1(Z_{1k},Z_{1\ell}) \big]_{k \ell},& \qquad \bm{K}_2 = \big[ k_2(Z_{2k},Z_{2\ell}) \big]_{k \ell}, \\
\bm{H} &= \Big[ \delta_{k \ell} - \frac{1}{n}  \Big]_{k \ell}.
\end{align*} 
($\delta_{k \ell}$ denotes the Kronecker delta.) Here we use 
\begin{align*}
k_1(z_k,z_\ell) = k_2(z_k,z_\ell) = \sqrt{(2\pi)^{-1}}e^{ -\frac{ (z_k - z_\ell)^2}{2 \sigma^2} }
\end{align*} 
the Gaussian kernel, with bandwidth $\sigma > 0$ chosen according to the \emph{median heuristic} as suggested in \cite{Pfister19}. Specifically, $\sigma$ is set to the square root of half the median of all pairwise squared distances among the ranks, i.e.,
\[
\sigma = \sqrt{ \frac{1}{2} \ \mathrm{median}_{k < \ell} \left( z_k - z_\ell \right)^2}.
\]
As for the previous two indicators, HSIC is {\it not} margin-free; therefore, we consider here only the copula version of it, viz.\ HSIC$(F_{Z_1}(Z_1),F_{Z_2}(Z_2))$ and its empirical rank-based counterpart.
%\vspace{1 mm}
%
\medskip

\noindent Finally, we consider two independence tests which are not based on any interpretable dependence measures, but have proved very powerful in previous empirical studies. The indicator taken into account in $\tilde{\xx}$ is then the $p$-value returned by the test when run on the given sample.  %and \cite{ddr}

\medskip

\subsubsection{Heller-Heller-Gorfine `HHG' test} %\cite{hhgtest} built their test statistic by noting that, for continuous random variables, dependence is equivalent to the existence of a point $(\zeta_1^\circ,\zeta_2^\circ)$ in the support of $F_{Z_1 Z_2}$ and some radii $r_1,r_2 >0$, such that $F_{Z_1 Z_2}(z_1,z_2) \neq F_{Z_1}(z_1)F_{Z_2}(z_2)$ for $(z_1,z_2)$ in $B(\zeta_1^\circ,r_1) \times B(\zeta_2^\circ,r_2)$. 
For $k,\ell \in \{1,\ldots,n\}$, let 
\begin{align*}
A_{00} (k,\ell) &= \sum_{m \notin \{k,\ell\}} \indic{|z_{1k} - z_{1m}| \leq |z_{1k} - z_{1\ell}|} \indic{|z_{2k} - z_{2m}| \leq |z_{2k} - z_{2\ell}|} \\
A_{10} (k,\ell) &= \sum_{m \notin \{k,\ell\}} \indic{|z_{1k} - z_{1m}| > |z_{1k} - z_{1\ell}|} \indic{|z_{2k} - z_{2m}| \leq |z_{2k} - z_{2\ell}|} \\
A_{01} (k,\ell) &= \sum_{m \notin \{k,\ell\}} \indic{|z_{1k} - z_{1m}| \leq |z_{1k} - z_{1\ell}|} \indic{|z_{2k} - z_{2m}| > |z_{2k} - z_{2\ell}|} \\
A_{11} (k,\ell) &= \sum_{m \notin \{k,\ell\}} \indic{|z_{1k} - z_{1m}| > |z_{1k} - z_{1\ell}|} \indic{|z_{2k} - z_{2m}| > |z_{2k} - z_{2\ell}|}. 
\end{align*} 
These count the number of observations in the four cells of a $(2 \times 2)$-contingency table induced by the double-dichotomy $\indic{|z_{1k} - Z_1| \leq |z_{1k} - z_{1\ell}|}$ and $\indic{|z_{2k} - Z_2| \leq |z_{2k} - z_{2\ell}|}$. For all $k,\ell$ and $q \in \{0,1\}$, define $A_{q \bullet }(k,\ell) = A_{q 0}(k,\ell) + A_{q 1}(k,\ell)$ and $A_{\bullet  q}(k,\ell) = A_{0 q}(k,\ell) + A_{1 q}(k,\ell)$; then 
\begin{align*}
S(k,\ell) = \frac{(n-2) \Big[ A_{00}(k,\ell) A_{11}(k,\ell) - A_{01}(k,\ell) A_{10}(k,\ell) \Big]^2 }{ A_{0\bullet }(k,\ell) A_{1\bullet }(k,\ell) A_{\bullet  0}(k,\ell) A_{\bullet 1}(k,\ell) },
\end{align*} 
the $\chi^2$-independence test statistic for that particular table. The test proposed in \cite{hhgtest} uses the amalgamated 
\[T = \sum^n_{k=1} \sum_{\ell\neq k} S(k,\ell) \]
as basic test statistic for (\ref{eqn:H0indep}). Note that, we could also just use $T^* = \max_{k,\ell} S(k,\ell)$ as test statistic. Likewise, given that the $\chi^2$-statistic is asymptotically equivalent to the likelihood ratio test (`$G$-test'), we could built an amalgamated test statistic as the sum of the likelihood-ratio test statistic, or as their maximum value over $(k,\ell)$. We keep these 4 versions of the `HHG' tests in our study. The (one minus) $p$-values were computed on 100 permutations. Note that the above HHG tests are not rank-based; hence for our computations we use sample ranks instead as the actual inputs, for sake of consistency with the above.

\medskip 

\subsubsection{Data--Driven Rank `DDR' Test} For $q \in \N$, denote $\Ls_q(x)$ the $q$th (orthonormal) Legendre polynomial shifted to $[0,1]$ \cite[Chapter 22]{Abramowitz83}. Then, noting that an equivalent statement to independence  between $Z_1$ and $Z_2$ is that $\rho \Big( \Ls_q \big( F_{Z_1}(Z_1) \big) , \Ls_{q'} \big( F_{Z_2}(Z_2) \big) \Big) = 0$ for all $q,q'\geq 1$, \cite{ddr} proposed the test statistic
\[T_Q = \sum^Q_{q=1} \bigg\{ \frac{1}{\sqrt{n}} \sum^n_{k=1} \Ls_q \Big( \frac{ r(z_{1k}) - \frac{1}{2} }{n} \Big) \Ls_q \Big( \frac{ r(z_{2k}) - \frac{1}{2} }{n} \Big) \bigg\}^2, \]
where $Q$ has to be appropriately selected (e.g., $Q=4$). Two criteria are suggested in \cite{ddr}. The first one uses modified Schwarz's rule
\begin{multline*}
S2 = \min_Q \big\{  1\leq Q \leq d(n), \, T_Q - Q \log n \geq T_j - j \log n, \, \\  j = 1,...,d(n) \big\}
\end{multline*} with $d(n)$ being a sequence of numbers tending to infinity as $n \rightarrow \infty$. The second criterion identifies the set $\Lambda = \{ (i,j) : i,j \in \{ 1,...,d(n) \} \} \cup \{ (1,1) \}$ such that $T_{\Lambda} - \big| \Lambda \big| \log n$ is maximised ($\big| \Lambda \big|$ is the cardinality of $\Lambda$).\footnote{The actual implementation in the package {\texttt{\scriptsize testforDEP::testforDEP()}} , which we used in our computations, involves slight modifications to the test statistics as described here, and full details can be found in the package source code.} We keep both versions in our study. The (one minus) $p$-values were computed on 100 permutations.

\ppn Altogether, this yields 13 dependence measures ({\it 1--13} above), 4 variants of the HHG test ({\it 14}), and 2 variants of the DDR test ({\it 15}), for a total of 19 dependence indicators. For ease of reference, Table~\ref{tab:indicators} lists them systematically. We stress once more that the purpose of this study is solely to illustrate the proposed idea, not to claim that the collection of indicators considered here is exhaustive or universally the most relevant. Accordingly, the reader may enrich this collection with any preferred indicator not included here, without altering the rationale of the approach.

\begin{table}[H]
  \centering
  \caption{Dependence indicators}\label{tab:indicators}
  \renewcommand{\arraystretch}{1.1}          % slightly taller rows
  \setlength{\tabcolsep}{5pt}                % a bit tighter columns
  \begin{tabularx}{\linewidth}{@{}r  % number
    >{\raggedright\arraybackslash}X  % description (automatic width)
    >{\centering\arraybackslash}m{1.6cm}  % abbreviation
    >{\raggedright\arraybackslash}m{2.3cm} % R function
    @{}}
    \toprule
    & \textbf{Indicator} & \textbf{Abb.} & \textbf{R function} \\ 
    \midrule
    1 & Spearman’s $\rho_S$ \cite{Spearman1904}                        & Spear & \texttt{mspear()} \\ 
    2 & Kendall’s $\tau$ \cite{Kendall1938}                            & Ken   & \texttt{mken()}   \\ 
    3 & Blomqvist’s $\beta$ \cite{Blomqvist1950}                       & Blom  & \texttt{mblom()}  \\ 
    4 & Area under Kendall plot \cite{Kplot}                           & AUK   & \texttt{mauk()}   \\ 
    5 & Hoeffding’s $D$ \cite{Hoef}                                    & Hoeff & \texttt{mhoeff()} \\ 
    6 & Hellinger correlation \cite{HellCor}                           & Hell  & \texttt{mhell()}  \\ 
    7 & Mutual information \cite{Rentropy}                             & Info  & \texttt{minf()}   \\ 
    8 & Maximal information \cite{MIC} criterion                      & MIC   & \texttt{mmic()}   \\ 
    9 & Maximum correlation \cite{ACE} coefficient via alternating conditional  expectations                                         & ACE   & \texttt{mace()}   \\[2pt]
   10 & Randomised dependence \cite{randcor} coefficient                & Rand  & \texttt{mrand()}  \\ 
   11 & Distance correlation \cite{dcor}                               & dCor  & \texttt{mdist()}  \\ 
   12 & Martingale difference \cite{Martdiff} correlation               & Martdiff   & \texttt{mmdif()}  \\ 
   13 & Hilbert–Schmidt \cite{hsic} independence criterion              & HSIC  & \texttt{mhsic()}  \\ 
    \midrule
    14 & Heller–Heller–Gorfine test \cite{hhgtest} ($\chi^2$‑sum)        & HHGPsum & \texttt{phhg()} \\
                        & Heller–Heller–Gorfine test (likelihood‑sum)    & HHGsum  &                         \\
                        & Heller–Heller–Gorfine test ($\chi^2$‑max)      & HHGPmax &                         \\
                        & Heller–Heller–Gorfine test (likelihood‑max)    & HHGmax  &                         \\ 
    \midrule
    15 & Data‑driven rank test \cite{ddr} & ddrV & \texttt{pddr()}\\ 
                        & ($\Lambda$ criterion)             &     &  \\ 
                        & Data‑driven rank test             & ddrTS2   &                         \\ 
                        &  (S2 criterion) & & \\
    \bottomrule
  \end{tabularx}
\end{table}

\ppn In addition to the above indicators, we include the sample size \( n \) explicitly in the input vector \(\tilde{\xx}\) associated with each sample in Scenarios~2 and~3. This is necessary because the significance of an observed dependence indicator cannot be evaluated independently of the sample size: the same indicator value may provide much stronger evidence of dependence in a large sample than in a small one. Since the deep network takes as input only the indicators, rather than the sample itself, omitting \( n \) would eliminate all information about the sample size. For this reason, \( n \) must be retained as an explicit input, and the resulting input vector \( \tilde{\xx} \) has length 20.

\subsection{Package} \label{subsec:compdet}

An R package\footnote{Available at \url{https://biostatisticien.eu/DeepTesting/}.} named \pkg{depstats} was developed to support the full infrastructure of our empirical study. Specifically, it contains functions to handle sample generation, computation of dependence indicators, model training, prediction and classification, output summarisation and visualisation of power results. 

\ppn In particular, the package implements all 19 dependence indicators listed in Table~\ref{tab:indicators} -- which also provides the name of the corresponding R function in the package. It also provides tools for generating synthetic samples under various dependence structures, including the 20 data generation processes (DGP) detailed in Appendix \ref{app:training} and which form the basis of the training set described in Section Section \ref{subsec:train}. It supports the training and application of deep-testing procedures based on greyscale image features ({Scenario 1}), dependence indicators ({Scenario 2}),  or their combination ({Scenario 3}), and facilitates the comparison of their power. 

\section{Methodology} \label{sec:meth}
 
\subsection{Generation of training samples} \label{subsec:train}

Supervised learning methods require a large number \(N\) of training units for which the label \(y\) is known (Step~\ref{it:step1} of the procedure; end of Section~\ref{sec:deptect}). Here, each unit \(i \in \{1,\ldots,N\}\) consists of a bivariate sample
\[
\Zs^{(i,n)} \doteq \{(z^{(i)}_{1k},z^{(i)}_{2k})\}_{k=1}^n,
\]
generated either from a factorisable distribution \(F_{Z_1Z_2}=F_{Z_1}F_{Z_2}\) (independence, in which case \(y_i=0\)) or from a non-factorisable distribution \(F_{Z_1Z_2}\) (dependence, in which case \(y_i=1\)). We included in the training set samples of sizes $n=50$, $100$, $200$ and \(400\). (Smaller or larger sample sizes could equally well be included, depending on the real data sets to which the deep-testing procedure is intended to be applied.) For each of these sample sizes, \(N=40{,}000\) training samples were generated, of which \(20{,}000\) are independent samples (labelled \(y_i=0\)) and \(20{,}000\) are dependent samples (labelled \(y_i=1\)), the latter consisting of \(1{,}000\) samples from each of the 20 dependence models described in Appendix \ref{app:training}. On each of these samples, we computed the vectors $\xx$ containing the greyscale density image, and the vectors $\tilde{\xx}$ containing both the values of the 19 dependence indicators (Table \ref{tab:indicators}) and the sample size.

\ppn By construction, both the input features $\xx$ and $\tilde{\xx}$ are margin-free; i.e., they take exactly the same values when computed on the transformed sample \(\{(\psi_1(z^{(i)}_{1k}),\psi_2(z^{(i)}_{2k}))\}_{k=1}^n\), for arbitrary monotone functions \(\psi_1\) and \(\psi_2\), as when computed on the original sample \(\{(z^{(i)}_{1k},z^{(i)}_{2k})\}_{k=1}^n\). Beyond its theoretical relevance (see Section~\ref{subsec:prelim}), this property has an important practical consequence in our setting: training samples may be generated without reference to specific marginal distributions. In particular, it is sufficient to simulate training samples -- under either \(H_0\) or \(H_1\) -- with uniform marginals, without loss of generality, since the marginals play no role in the subsequent analysis.

\ppn In particular, to represent the collection of samples \(\Zs\) that may arise under \(H_0\) (independence), it is enough to draw \(n\) pairs of independent \(\Us_{[0,1]}\) random variables; that is, \(\{(z_{1k},z_{2k})\}_{k=1}^n\) is generated by taking \((z_{11},\ldots,z_{1n}) \overset{\mathrm{i.i.d.}}{\sim} \Us_{[0,1]}\) and, independently, \((z_{21},\ldots,z_{2n}) \overset{\mathrm{i.i.d.}}{\sim} \Us_{[0,1]}\). This reflects the fact that dependence need only be studied at the copula level: any sample \((Z_1,Z_2)\) generated from a distribution satisfying \(F_{Z_1Z_2}=F_{Z_1}F_{Z_2}\) becomes uniformly distributed on the unit square after copula transformation (\ref{eqn:coptrans}). For each sample size, the \(20{,}000\) training samples generated under independence are therefore such bivariate uniform samples.

\ppn A similar observation applies to the generation of training sets under \(H_1\): while we seek to produce samples exhibiting {\it dependence}, the marginal distributions remain entirely irrelevant. For instance, it is natural to generate samples from a bivariate Gaussian distribution with \(\rho \neq 0\) to represent the classical form of Gaussian dependence (`linear dependence' if the marginals are themselves Gaussian). There is then no need to generate further samples having the same Gaussian copula but different marginals, since they are equivalent from the point of view of the underlying dependence structure and, therefore, of the dependence indicators.

\begin{figure}
	\begin{center}
		\includegraphics[width=0.5\textwidth]{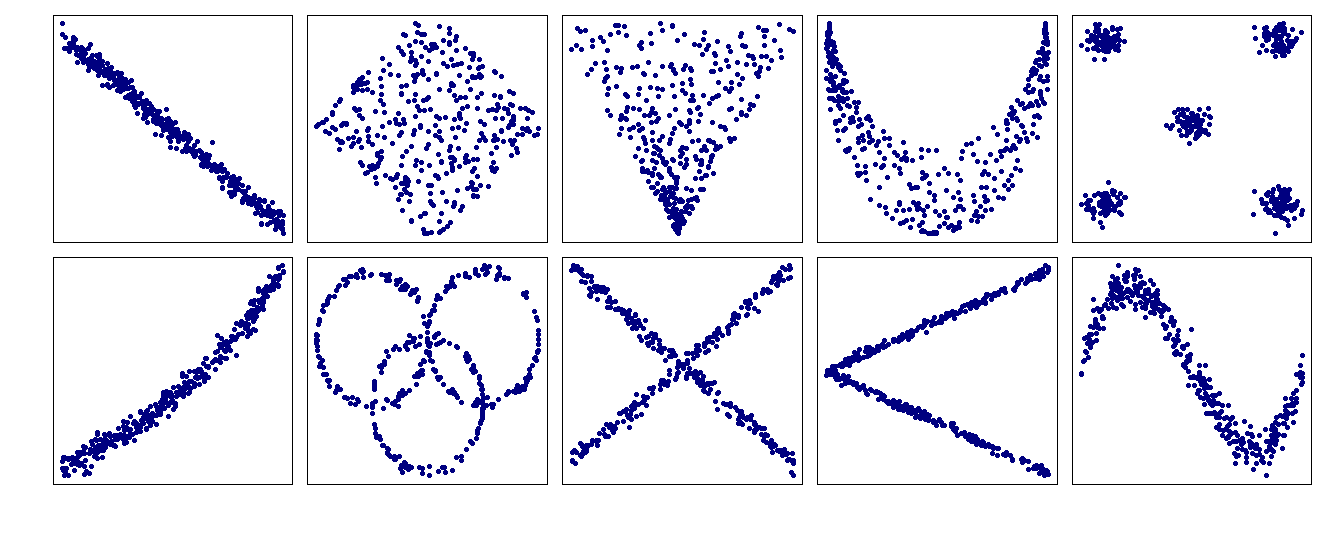}
	\end{center}
	\caption{Typical samples of size $n=400$ generated from the training models 1--10; from top left to bottom right: Linear, Diamond, Triangle, Crescent, Points, Exponential, Circles, Cross, Wedge, Cubic.} \label{fig:train1}
\end{figure}

\begin{figure}
	\begin{center}		\includegraphics[width=0.5\textwidth]{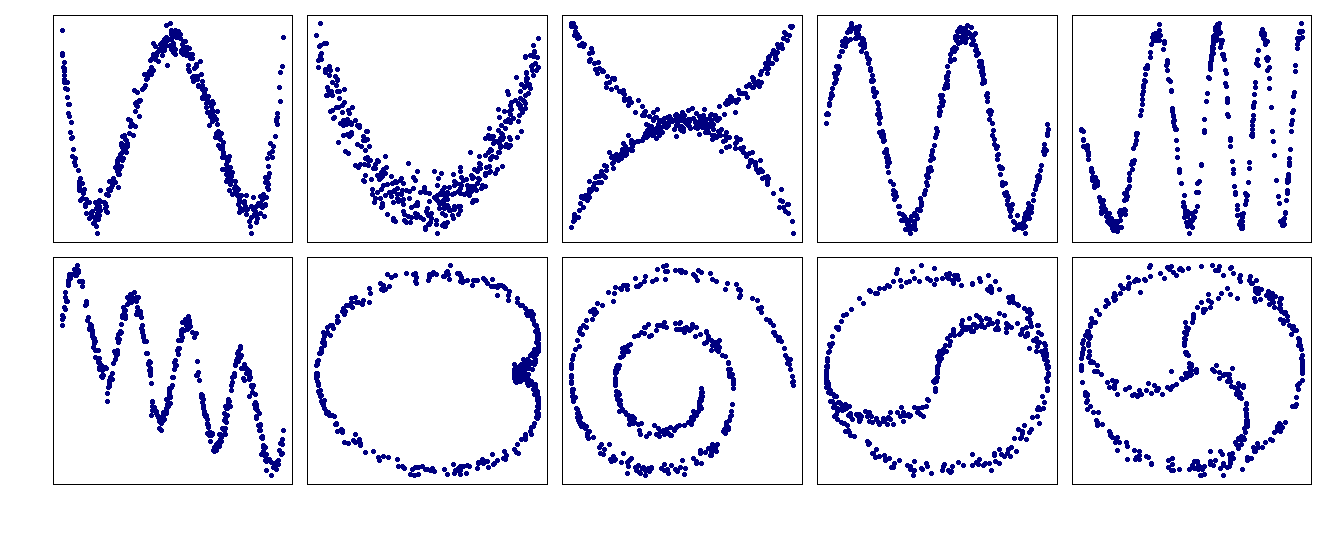}
	\end{center}
	\caption{Typical samples of size $n=400$ generated from the training models 11--20; from top left to bottom right: W-shape, Parabola, Two-parabola, Sine, Doppler, Heavy-sine, Heart, Spiral, Taegeuk, Samtaegeuk.} \label{fig:train2}
\end{figure}

\ppn The \(20{,}000\) dependent samples should nevertheless reflect as wide a range of {\it dependence structures} as possible, so that a neural network trained on them is unlikely to be caught unprepared when presented with a new sample. To this end, we considered 20 main dependence structures drawn from the existing literature on independence testing. Figures~\ref{fig:train1} and~\ref{fig:train2} display one representative sample of size \(n=400\) generated under each of these 20 models, at a low noise level so that the nature of the underlying relationship can be seen clearly. For each model, we generated \(1{,}000\) distinct samples for each of the sample sizes \(n=50\), \(100\), \(200\), and \(400\). Moreover, to maximise the diversity of the training samples, both the structural parameters of the models and the noise level, which governs the strength of dependence, were randomised. Whenever meaningful, a random rotation was also applied to the output. This ensures that even samples generated from the same underlying dependence model remain markedly different from one another and therefore each contribute genuinely new information about how dependence may manifest itself in a sample of that size. Full details on the 20 dependence models and on the randomisation of their parameters are provided in Appendix \ref{app:training}.

\subsection{Deep learning and computational details}\label{subsec:deep_learning}

\begin{figure}[h]
\centering
\includegraphics[width=0.5\textwidth]{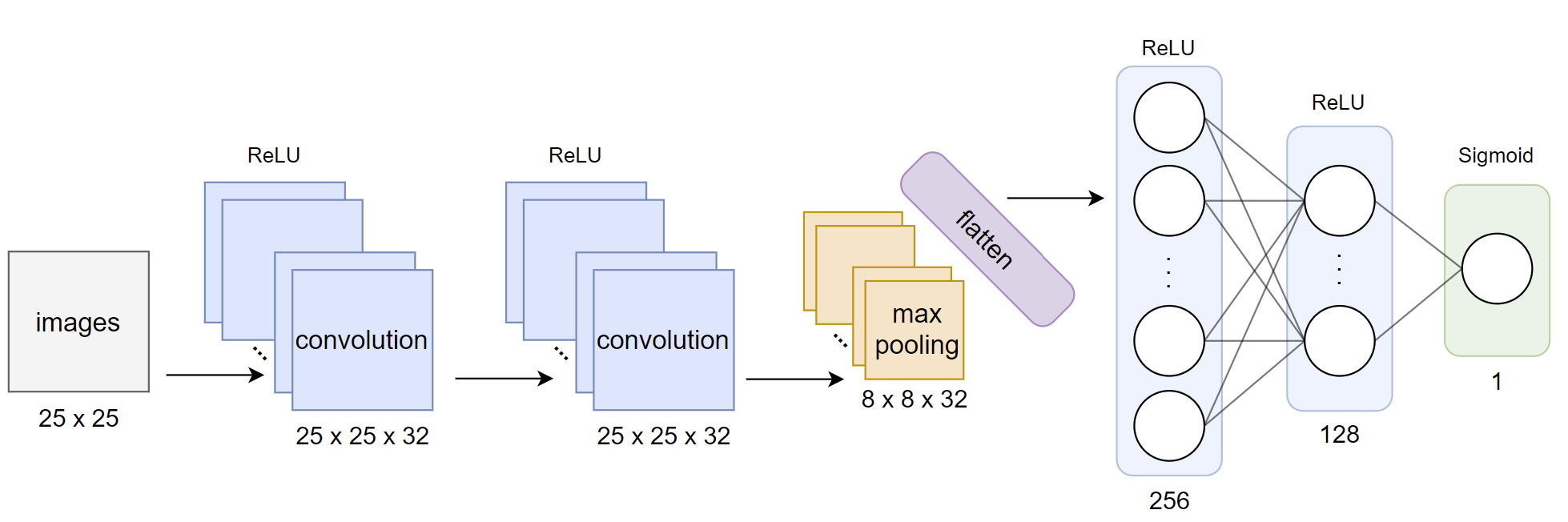}
\caption{Convolutional neural network architecture {\texttt{\scriptsize All-CNN}} for the analysis of images only (Scenario 1).}
\label{fig:cnn}
\end{figure}

\begin{figure}[h]
\centering
\includegraphics[width=0.4\textwidth]{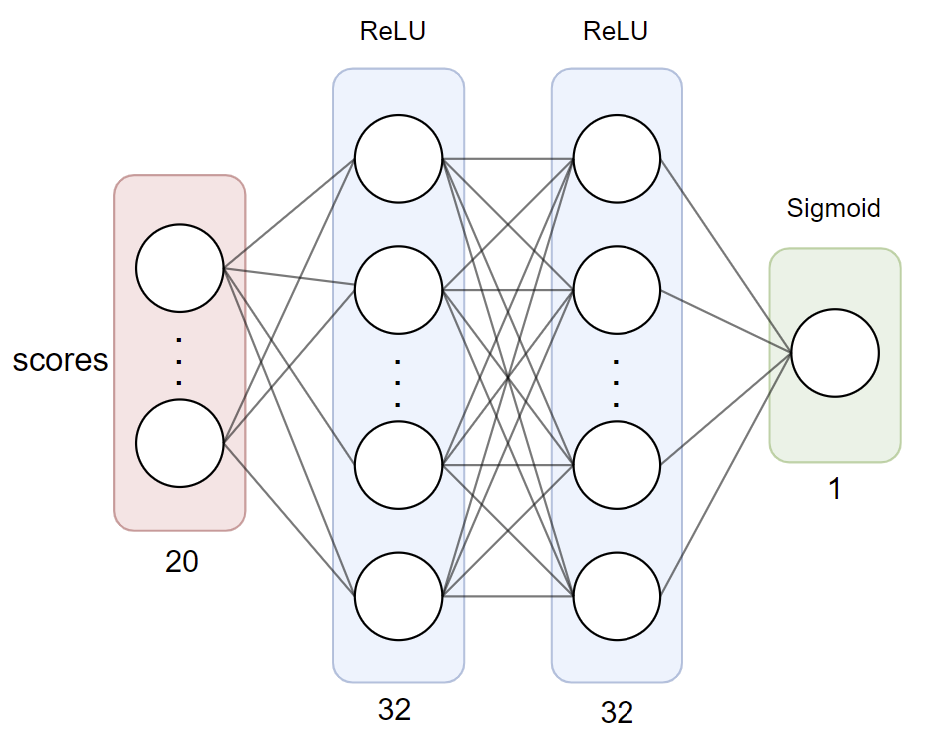}
\caption{Feed-forward neural network architecture {\texttt{\scriptsize All-MLP}} using only dependence indicators (and sample size) as input features (Scenario 2).}
\label{fig:mlp}
\end{figure}

\begin{figure}[h]
\centering
\includegraphics[width=0.5\textwidth]{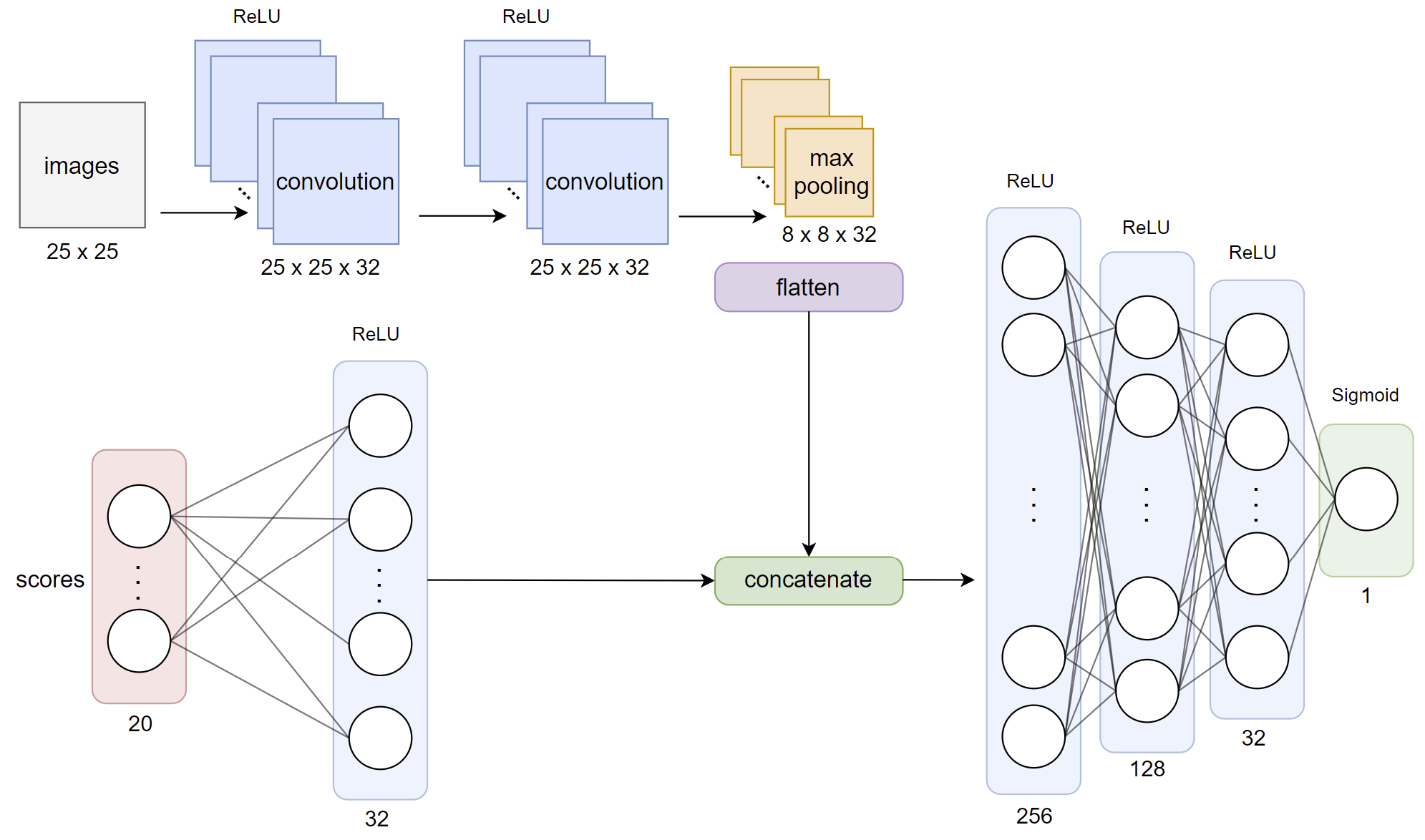}
\caption{Neural network architecture {\texttt{\scriptsize All-CNN-MLP}} using both dependence indicators (and sample size) and images as input features (Scenario 3).}
\label{fig:cnnmlp}
\end{figure}

We explored several neural network architectures and ultimately retained three, denoted \pkg{All-CNN}, \pkg{All-MLP}, and \pkg{All-CNN-MLP}. The prefix \pkg{All} signifies that the training set used for each of these networks comprised samples of all four considered sizes, \(n=50\), \(100\), \(200\), and \(400\).\footnote{An alternative approach would have been to train on samples of only one fixed size at a time, thus leading to a distinct classifier \(\hat{\phi}\) for each sample size.} The suffix indicates the corresponding deep learning architecture. The \pkg{All-CNN} network is a convolutional neural network (CNN), extending the standard architecture by incorporating convolutional and pooling layers. Such architectures have consistently proved effective for image-processing tasks \cite{Albawi17,CNN}, which motivates their use in Scenario~1.  The \pkg{All-MLP} network is a standard feed-forward multilayer perceptron (MLP), and is used for Scenario~2. Finally, the \pkg{All-CNN-MLP} network combines the previous two architectures and is used for Scenario~3. The exact architectures of these three networks are displayed in Figures~\ref{fig:cnn}, \ref{fig:mlp}, and~\ref{fig:cnnmlp}, respectively.

\ppn The \pkg{All-CNN} model takes as input \(25\times25\times1\) images and applies two \(3\times3\) convolutional layers (32 filters, ReLU activation, same padding), followed by a \(3\times3\) max-pooling layer and 20\% dropout. The resulting representation is then flattened and passed through dense layers of size 256 and 128 (each followed by 20\% dropout), before the final sigmoid output layer, yielding a total of \(567{,}137\) trainable parameters.  The \pkg{All-MLP} model processes a 20-dimensional input (19 score features and the sample size) through two dense layers of size 32 (with ReLU activation and 20\% dropout), followed by a sigmoid output layer, for a total of \(1{,}761\) trainable parameters. The hybrid \pkg{All-CNN-MLP} network concatenates a 2,048-dimensional image embedding with a 20-dimensional score embedding obtained through a single dense layer of size 32, and then feeds the resulting 2,080-dimensional vector through dense layers of size 256, 128, and 32 before the sigmoid output layer. This network has \(580{,}001\) trainable parameters.

\ppn For each scenario/network architecture, the optimal map \(\hat{\phi}\) was identified as in~\eqref{eqn:hatphiDL} by minimising the binary cross-entropy loss. The resulting maps \(\hat{\phi}_1\), \(\hat{\phi}_2\), and \(\hat{\phi}_3\) may therefore be viewed as test statistics that are strongly discriminative between independent and dependent samples for the corresponding input representation.

\ppn All experiments were conducted on a Dell Precision 5550 laptop equipped with an Intel Xeon W-10855M CPU at 2.80~GHz (6 cores, 12 threads), 31~GiB of RAM, an NVIDIA Quadro T2000 Mobile GPU (4~GiB, CUDA~12.0), and Intel UHD Graphics P630, running Debian GNU/Linux~12 (Bookworm, kernel~6.1.0-17-amd64). The software stack comprised R (v4.x), together with the \pkg{keras} and \pkg{reticulate} packages, interfacing Python~3.11.2 in the \pkg{r-reticulate} virtual environment, which provided TensorFlow and Keras as the backend. All three networks were implemented in R using \pkg{keras}, with \pkg{reticulate} providing the interface to Python's \pkg{tensorflow.keras} backend. They were trained under the same configuration: Adam optimiser (\pkg{learning\_rate=\(10^{-3}\)}, \(\beta_1=0.9\), \(\beta_2=0.999\), \(\epsilon=10^{-7}\), \pkg{amsgrad=FALSE}), binary cross-entropy loss, accuracy as the performance metric, batch size 128, and a maximum of 50 epochs with a 20\% validation split. In addition, EarlyStopping was applied to \pkg{val\_loss} (\pkg{patience=3}), together with ReduceLROnPlateau (\pkg{factor=0.5}, \pkg{patience=2}, \pkg{min\_lr=\(10^{-6}\)}).

\ppn The final evaluation on the full training set is summarised in Table~\ref{tab:all_models_metrics}, which reports several standard classification metrics -- namely, the area under the ROC curve (AUC), accuracy, precision, recall, and F1-score -- for each of the three network architectures/scenarios.

\begin{table}[ht]
\centering
\caption{Evaluation metrics for the three network architectures}
\label{tab:all_models_metrics}
\begin{tabular}{@{}l c | c c c c c@{}}
\toprule
Model & Scen. & AUC & Acc. & Prec. & Recall & F1-score \\
\midrule
{\texttt{\scriptsize All-CNN}}    & 1 & 0.996 & 0.976 & 0.988 & 0.963 & 0.975 \\
{\texttt{\scriptsize All-MLP}}    & 2 & 0.993 & 0.961 & 0.986 & 0.936 & 0.960 \\
{\texttt{\scriptsize All-CNN-MLP}} & 3 & 0.997 & 0.977 & 0.988 & 0.966 & 0.977 \\
\bottomrule
\end{tabular}
\end{table}

\noindent Table~\ref{tab:all_models_metrics} shows that all three architectures achieve excellent classification performance: AUC values above 0.99 confirm that each map $\hat{\phi}_1$, $\hat{\phi}_2$, and $\hat{\phi}_3$ can rank dependent samples above independent ones across a wide range of thresholds, while accuracy values above 0.96 show that the proportion of correctly classified samples is very high throughout. Precision values near 0.99 indicate that predicted dependence is almost always correct, and the recall values show that most dependent samples are identified as such -- though \pkg{All-MLP} misses slightly more than the other two architectures, as also reflected in its lower F1-score. 

\ppn Thus -- and not surprisingly -- increasing model complexity yields steadily better performance: the lightweight \pkg{All-MLP} ($\sim$1.8~K parameters) trails both convolutional models, while the \pkg{All-CNN} ($\sim$567~K parameters) outperforms it and the hybrid \pkg{All-CNN-MLP} ($\sim$580~K parameters) only marginally edges out the \pkg{All-CNN} -- consistent with its relatively small ($\sim$13~K) parameter increase. Nevertheless, the \pkg{All-MLP}, despite its much smaller parameter count, performs impressively well by fully leveraging the information embedded in all the 19 dependence indicators listed in Table \ref{tab:indicators}.

\ppn  The previous discussion describes the construction and the performance of the respective classifiers $\hat{\phi}_1, \hat{\phi}_2$ and $\hat{\phi}_3$ for each architecture and/or scenarios. We now explain how they are converted into a statistical test that controls the nominal level \(\alpha\).

\subsection{Near-exact critical values} \label{subsec:exactcrit}

When computed on a newly observed sample $\Zs^\text{new}$, the maps $\hat{\phi}_1, \hat{\phi}_2$ and $\hat{\phi}_3$ return a dependence score between 0 and 1, with a value `too large' being an indication of dependence in $\Zs^\text{new}$. What `too large' means here is to be fixed by the level $\alpha$ of the test (i.e., the maximum probability of type I error which we would allow). The threshold value which would maintain any level $\alpha \in (0,1)$ can be identified near-exactly (Section \ref{sec:pierre}). Indeed the copula transformation (\ref{eqn:coptrans}) turns the composite independence hypothesis (\ref{eqn:H0indep}) into a simple, equivalent one: 
\[ H_0: (F_{Z_1}(Z_1),F_{Z_2}(Z_2)) \sim \Us_{[0,1]^2}.\]
Hence, regardless of the sample size $n^\text{new}$ of $\Zs^\text{new}$, the exact sampling distribution of any copula-based test statistic (such as the maps $\hat{\phi}_1, \hat{\phi}_2$ and $\hat{\phi}_3$ in our case) can be approached arbitrarily close by Monte-Carlo simulations. 

\ppn For illustration, we consider below sample sizes $n^\text{new} \in\{30, 50, 100, 200, 300, 400\} $. For each of these, we generate $N'=50,000$ `independent samples' like in Section \ref{subsec:train} (i.e., bivariate uniform on the unit square) -- this is the null-calibration set described in Step \ref{it:nullcalib} of the summarised procedure (end of Section \ref{sec:deptect}). On each of these samples, we computed the vectors $\xx$ containing the greyscale density image, and $\tilde{\xx}$ containing both the values of the 19 dependence indicators (Table \ref{tab:indicators}) and the sample size. From these, we then computed the corresponding values returned by $\hat{\phi}_1, \hat{\phi}_2$ and $\hat{\phi}_3$. For each of these `deep-test-statistics', the empirical distribution of the observed $N'$ values reconstructs near-exactly the sampling distribution under $H_0$, despite the highly non-linear nature of these quantities.

\ppn For the nominal level $\alpha \in (0,1)$, we then extract the critical value $d_\alpha$ (Step \ref{it:crit} in the summarised procedure; end of Section \ref{sec:deptect}). The critical values, for the considered sample sizes $n^\text{new}$ and customary level $\alpha = 0.05$, are listed for the three deep-test statistics $\hat{\phi}_1, \hat{\phi}_2$ and $\hat{\phi}_3$ in Table~\ref{tab:critical_values}. 

\begin{table}[ht]
    \centering
    \caption{Critical values $d_\alpha$ at \(\alpha = 0.05\) for different sample sizes and models}
    \label{tab:critical_values}
    \begin{tabular}{cccc}
        \toprule
        Sample size & {\texttt{\scriptsize All-CNN}} & {\texttt{\scriptsize All-MLP}} & {\texttt{\scriptsize All-CNN-MLP}} \\
        $n^\text{new}$ & $\hat{\phi}_1$ & $\hat{\phi}_2$ & $\hat{\phi}_3$  \\
        \midrule
        30  & 0.817 & 0.737 & 0.683 \\
        50  & 0.481 & 0.459 & 0.454 \\
        100 & 0.182 & 0.301 & 0.150 \\
        200 & 0.034 & 0.108 & 0.019 \\
        300 & 0.011 & 0.075 & 0.005 \\
        400 & 0.006 & 0.022 & 0.003 \\
        \bottomrule
    \end{tabular}
\end{table}

\subsection{Testing samples}\label{subsec:valid}

Finally, we seek to assess the power of the proposed deep-testing procedures; that is, their ability to detect dependence in bivariate samples. To this end, we consider two empirical experiments, each involving a large number of {\it testing samples} generated as follows.  

\ppn {\bf Experiment 1:} We generated \(20{,}000\) bivariate samples from the same 20 dependence models used to generate the dependent training samples (Figures~\ref{fig:train1} and~\ref{fig:train2}), namely \(1{,}000\) samples from each model, {\it but using different parameter values}; see details in Appendix~\ref{app:testing}. Due to the randomness in the parameters, noise levels, and rotations, these samples remain substantially different from those used to train the networks.

\begin{figure}
	\begin{center}
		\includegraphics[width=0.5\textwidth]{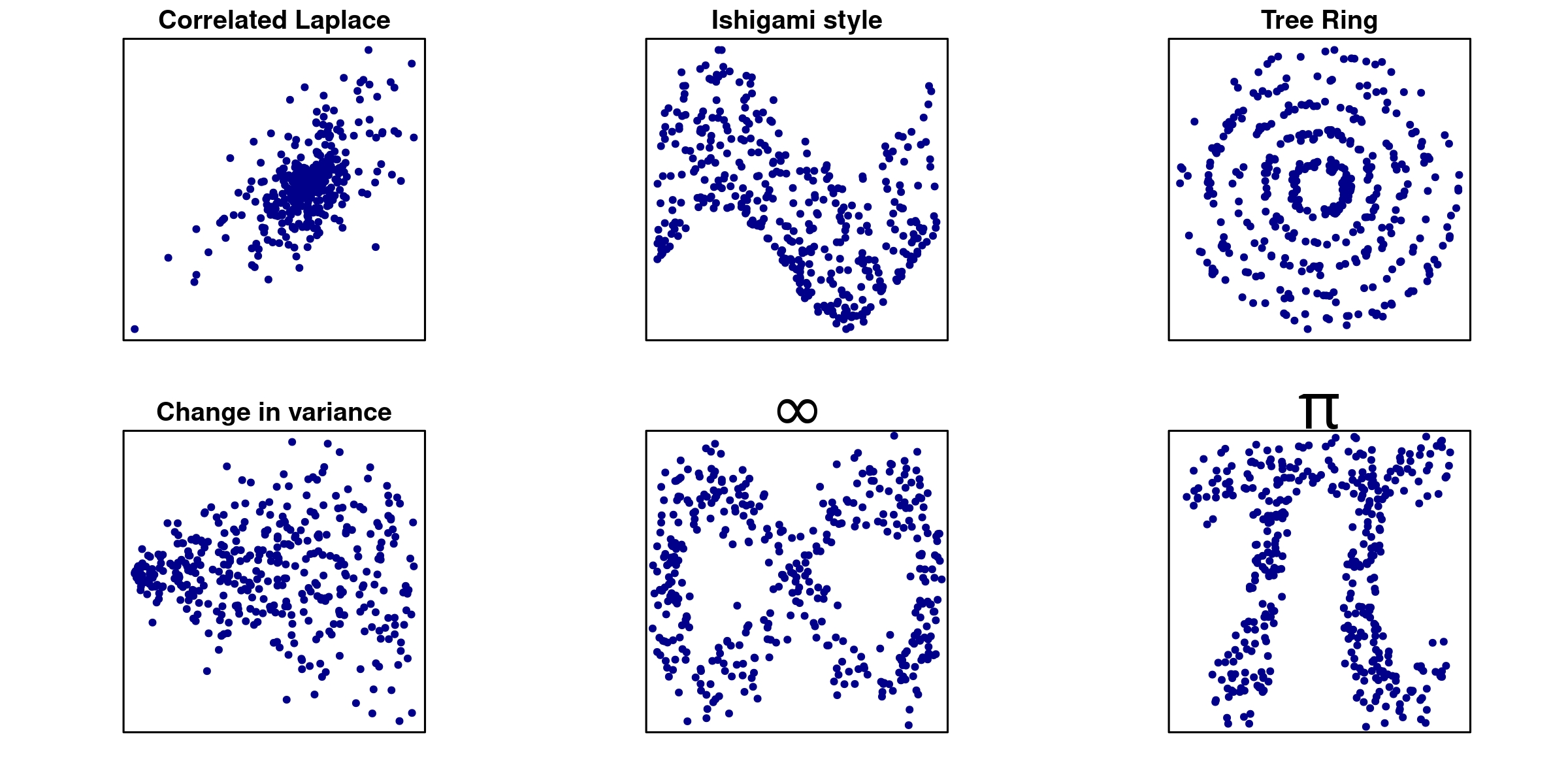}
	\end{center}
	\caption{Representative samples of size \(n=400\) generated from six novel dependence patterns; from top left to bottom right: Laplace, Ishigami, Tree Ring, Variance, Infinity, Pi.}
	\label{fig:test}
\end{figure}

\ppn {\bf Experiment 2:} We investigate the behaviour of the deep tests on dependent samples generated from dependence models not seen during training. Specifically, we generated 12,000 testing samples as follows (see Appendix~\ref{app:testing} for details):
\begin{enumerate}[($i$)]
	\item \(2{,}000\) samples from each of four novel dependence structures (`Laplace', `Ishigami', `Tree Ring', `Variance'), with two different noise levels (\(1{,}000\) samples per noise level for each model); typical samples of size \(n=400\) are shown in Figure~\ref{fig:test}.
	\item \(2{,}000\) samples from each of two `image-based'\footnote{This means that their dependence patterns are defined visually rather than analytically; see Appendix \ref{app:testing}.} dependence structures (`Infinity' and `Pi'), whose typical samples of size \(n=400\) are also shown in Figure~\ref{fig:test}.
\end{enumerate}

\ppn We repeated Experiments 1 and 2 on sample sizes \(n^{\text{new}} \in \{30,50,100,200,300,400\}\) -- that is, the same sizes as the training samples, together with the additional unseen sample sizes $30$ and $300$. Each of our three proposed deep-testing procedures was subsequently applied to all testing samples generated for each sample sizes, which allowed us to obtain Monte Carlo approximations of the power of each test for every dependence model as the sample size increases (see Figures \ref{fig:pow1} and \ref{fig:pow2}, and Appendix \ref{app:power}). For benchmarking purposes, we also applied to the same testing samples the independence tests based on each individual dependence indicator listed in Table~\ref{tab:indicators}. By the same argument as in Section~\ref{subsec:exactcrit}, these tests can likewise be made near-exact: the \(N'=50{,}000\) independent samples generated there for each sample size \(n^{\text{new}}\) may also be used to approximate arbitrarily closely the null distribution of each such indicator, and near-exact critical values for each of them may be deduced. 

\ppn This means that, by construction, all the procedures under comparison -- both our three deep-tests and the tests based on the individual indicators -- have rejection rates under \(H_0\) arbitrarily close to the prescribed level for all sample sizes considered. Consequently, the ability to maintain the nominal level is not, in itself, informative for comparing their performance in the present setting. The comparison carried out in the next section therefore rests exclusively on the power of the tests under the two experimental settings described above.

\section{Power results}\label{sec:res}

The full set of numerical results is reported in Appendix \ref{app:power} in the form of power tables and power curves. Figures~\ref{fig:pow1} and~\ref{fig:pow2} below show only selected representative results from Experiments~1 and~2, respectively, to facilitate exposition. Based on these results, the main empirical conclusions can be summarised as follows.

\ppn First, an internal comparison between the three deep-testing procedures across both experiments shows that the image-based \pkg{All-CNN} (Scenario 1) and the hybrid \pkg{All-CNN-MLP} (Scenario 3) are generally the most powerful, especially against highly non-linear or geometric alternatives for which the shape of the scatter plot is particularly informative (e.g., Diamond or Spiral; see Figures \ref{fig:train1}, \ref{fig:train2} and \ref{fig:pow1}). The dependence-indicator-based \pkg{All-MLP} (Scenario 2), however, also performs remarkably well, despite involving substantially fewer trainable parameters and substantially lower computational cost than the convolutional architectures (Section~\ref{subsec:deep_learning}). Overall, this broadly confirms the picture already suggested by Table~\ref{tab:all_models_metrics}.

\ppn Second, combining dependence-indicator information with image information does not degrade performance and often yields the best power. This again illustrates the well-known robustness of deep learning methods to overparameterisation. In practice, however, the gain of \pkg{All-CNN-MLP} over \pkg{All-CNN} is often modest, suggesting that the greyscale image representation already captures most of the discriminating signal. Nevertheless, the hybrid architecture provides an additional safeguard against alternatives that are better captured by specific indicators than by geometry alone.

\ppn Within Experiment~1, the proposed \pkg{All-CNN-MLP} procedure emerges as the strongest method {\it overall} across the full benchmark of 20 training dependence models. As detailed in Table \ref{tab:pow1} (Appendix \ref{app:power}), it achieves {\it the highest average power for every sample size} from \(n=30\) to \(n=300\), and ties for first place with \pkg{All-MLP} at \(n=400\). More precisely, its average power is \(0.528\), \(0.703\), \(0.861\), \(0.941\), and \(0.969\) for \(n=30,50,100,200,\) and \(300\), respectively, each value exceeding that of every competing procedure; at \(n=400\), \pkg{All-CNN-MLP} and \pkg{All-MLP} both attain \(0.976\).

\ppn By comparison, the strongest competitors on average are the `likelihood-sum' variant of the HHG test \cite{hhgtest} and the test based on mutual information \cite{Rentropy}, but even these remain clearly below the deep-testing procedures in the aggregate rows. For instance, at \(n=100\), the average power of \pkg{All-CNN-MLP} is \(0.861\), compared with \(0.820\) for \pkg{hhgGs} and \(0.795\) for \pkg{Info}. 

\ppn Importantly, essentially the same pattern persists for Experiment 2, where the testing samples were generated by the six dependence models not previously seen by the deep-testing classifiers (Table~\ref{tab:pow3}): the deep tests remain strongest overall. Indeed the power curves displayed in Figure \ref{fig:pow2} do not show any marked difference with those in Figure \ref{fig:pow1}. This confirms that the power of the deep tests observed above cannot be attributed to familiarity with the training models. In any case, as explained earlier, the substantial randomisation of the various parameters entering these data-generating processes ensures that even samples generated from the same model remain highly diverse.

\begin{figure}
	\begin{center}
		\includegraphics[height=0.23\textheight]{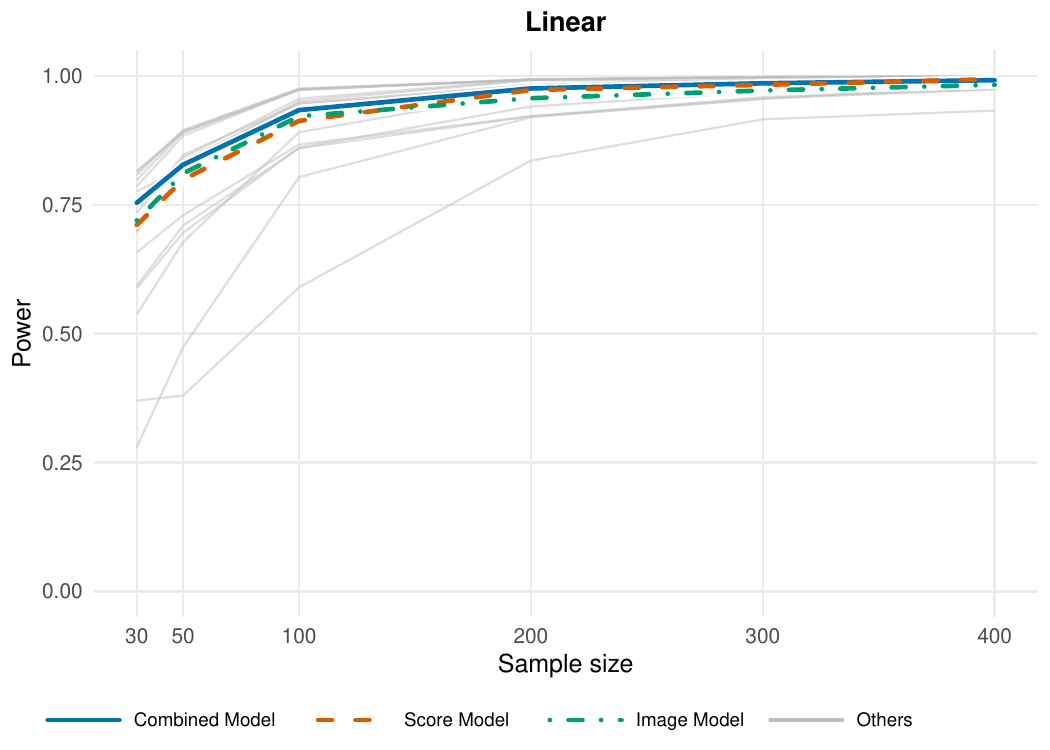} \\
        \includegraphics[height=0.23\textheight]{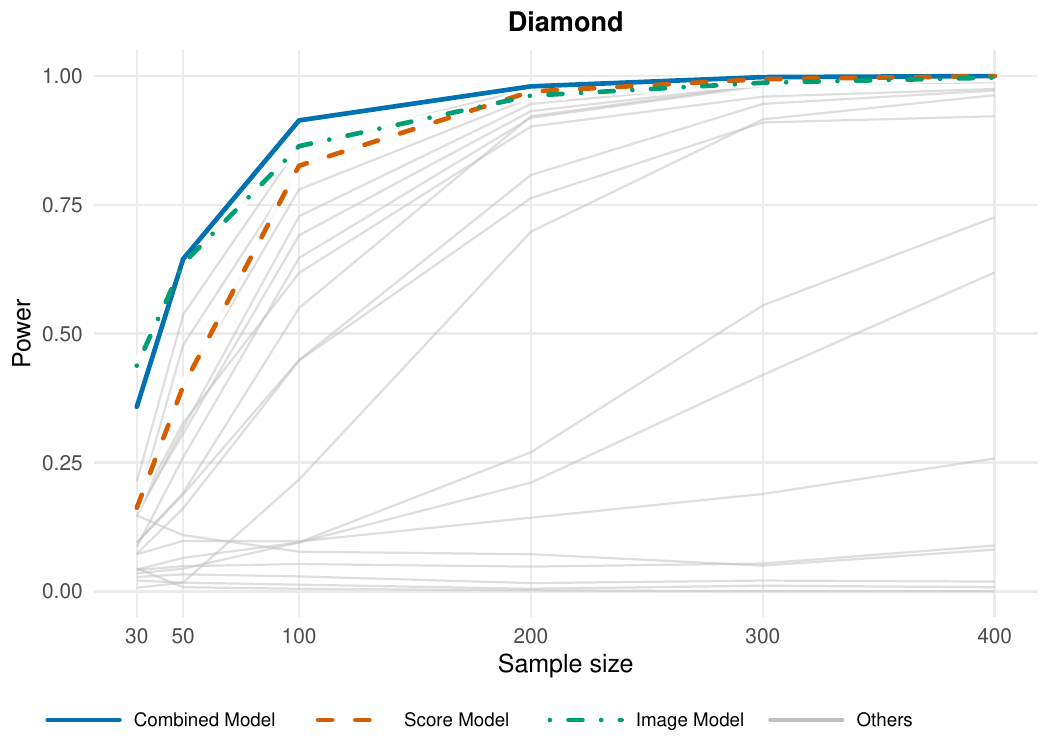} \\
        \includegraphics[height=0.23\textheight]{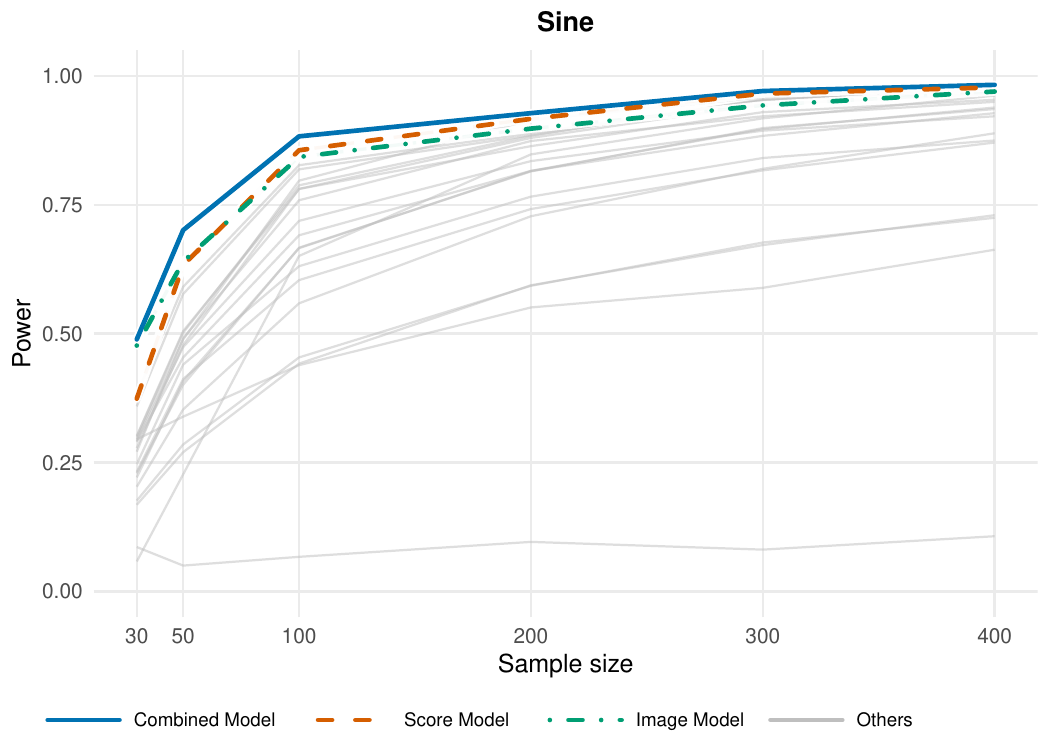} \\
        \includegraphics[height=0.23\textheight]{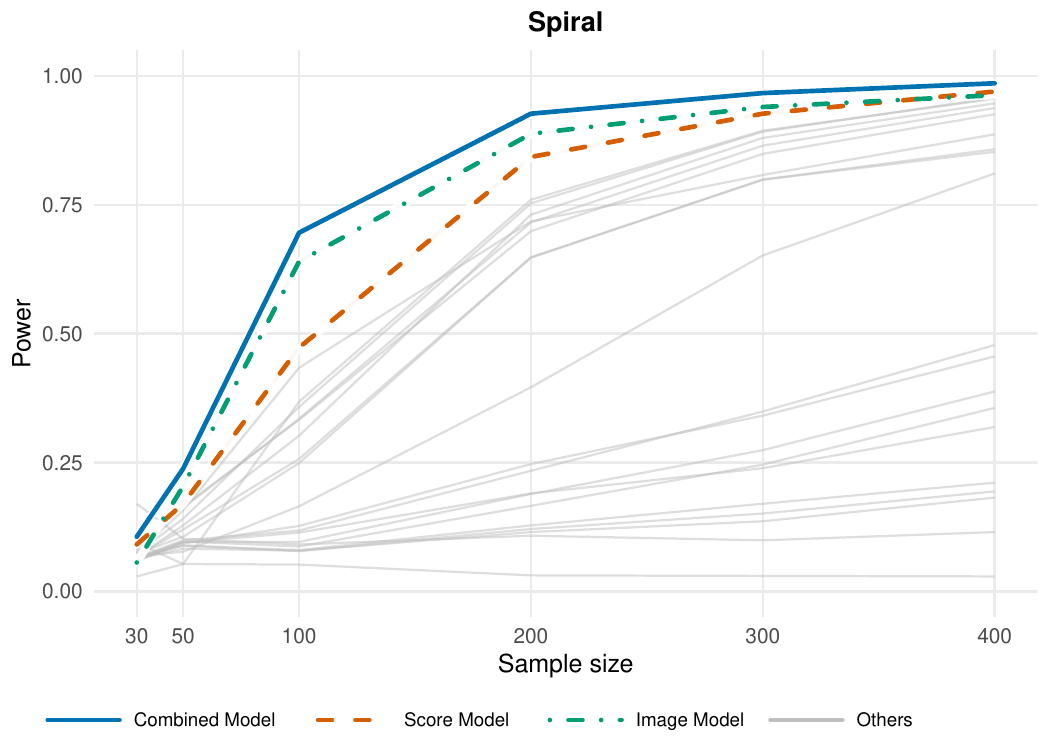}
    \end{center}
	\caption{(Monte-Carlo) Power of the proposed 3 deep-testing procedures and their 19 competitors for 4 training dependence models (Linear, Diamond, Sine, Spiral; see Figures \ref{fig:train1}--\ref{fig:train2}). Blue: \texttt{\scriptsize ALL-CNN-MLP}; Green: \texttt{\scriptsize ALL-CNN}; Orange: \texttt{\scriptsize ALL-MLP}; Grey: other competitors (Table \ref{tab:indicators}).}
	\label{fig:pow1}
\end{figure}

\begin{figure}
	\begin{center}
        \includegraphics[height=0.23\textheight]{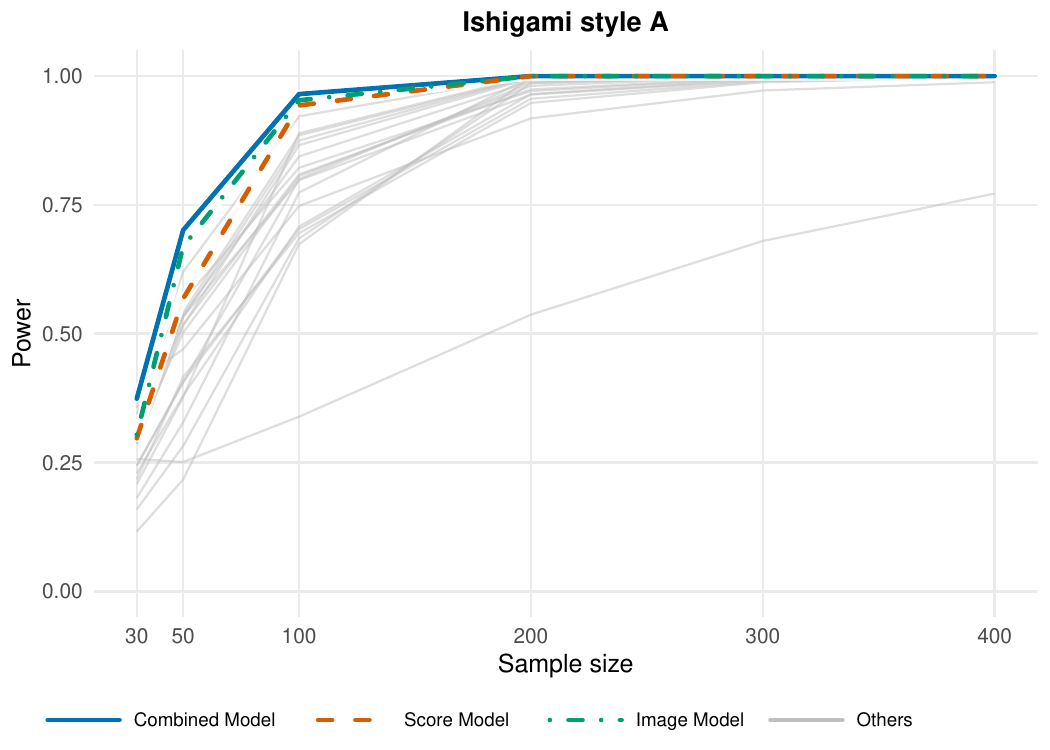} 
        \includegraphics[height=0.23\textheight]{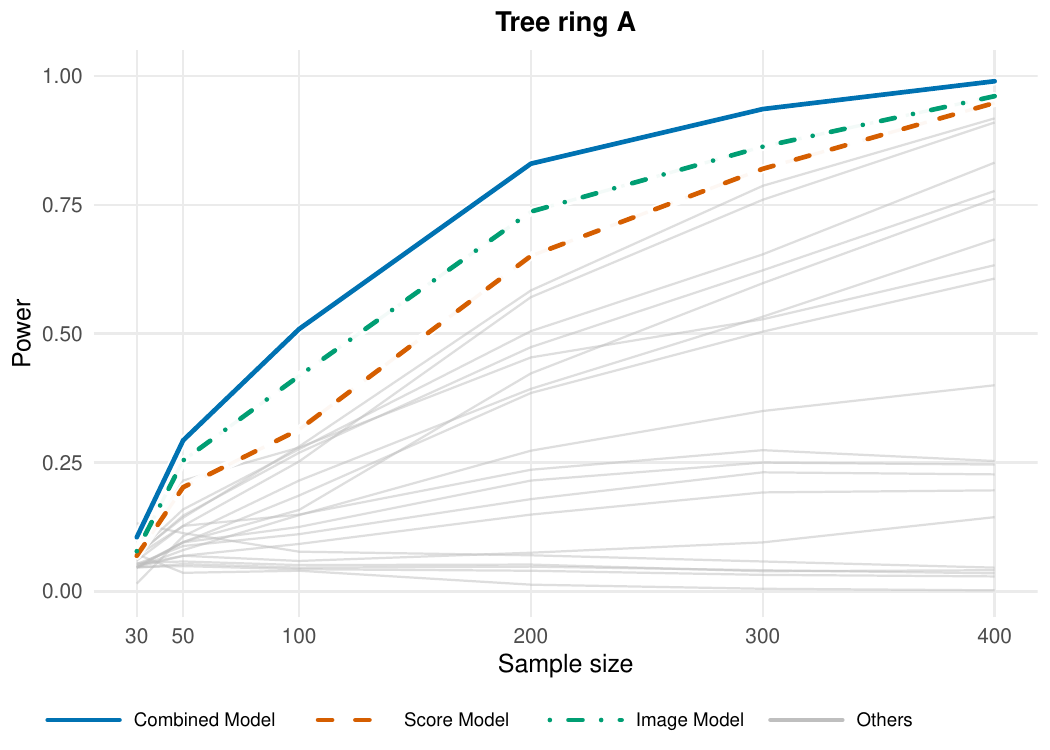} \\
    \end{center}
	\caption{(Monte-Carlo approximated) Power of the proposed 3 deep-testing procedures and their 19 competitors for 2 testing dependence models (Ishigami and Tree Ring; see Figure \ref{fig:test}). Blue: \texttt{\scriptsize ALL-CNN-MLP}; Green: \texttt{\scriptsize ALL-CNN}; Orange: \texttt{\scriptsize ALL-MLP}; Grey: other competitors (Table \ref{tab:indicators}).}
	\label{fig:pow2}
\end{figure}

\ppn The superiority of \pkg{ALL-CNN-MLP} is not limited to its mean performance. It also achieves the smallest average `power gap' relative to the best-performing method across all dependence models, for every sample size, in both Tables~\ref{tab:pow1} and~\ref{tab:pow3}. Thus, even in the relatively few settings where it is not the top-performing procedure, it remains, on average, closer to the best attainable power than any competing method, underscoring its robustness across heterogeneous dependence structures.

\ppn This conclusion is further supported by the `max-gap' criterion, defined as the maximum power loss observed at a given sample size across all dependence models. Under this criterion as well, \pkg{ALL-CNN-MLP} performs best for essentially all sample sizes in both Experiment~1 (Table~\ref{tab:pow1}) and Experiment~2 (Table~\ref{tab:pow3}).

\ppn The `Linear' model provides a clear illustration of this (Figure~\ref{fig:pow1}, top panel). In this setting, tests based on indicators that directly measure this specific type of concordance, such as Kendall's \(\tau\) or Spearman's \(\rho\), tend to be more powerful than the more {\it omnibus} deep tests. Intuitively, capturing a simple linear trend does not require the full flexibility of a deep neural network, and a more targeted method may be more efficient. Even so, the deep tests remain close behind. By contrast, in the other models displayed in Figures~\ref{fig:pow1}--\ref{fig:pow2}, the deep test \pkg{ALL-CNN-MLP}, and to a lesser extent \pkg{ALL-CNN}, consistently outperforms all competing methods, sometimes by a wide margin.

\ppn Overall, \pkg{ALL-CNN-MLP} emerges as the strongest choice in the benchmark: it achieves the highest average power, remains closest to the best-performing method across settings, and is the least prone to marked losses on more challenging dependence patterns. The separate \pkg{ALL-MLP} and \pkg{ALL-CNN} procedures show complementary strengths, with \pkg{ALL-MLP} performing particularly well on smoother or more regular relationships and \pkg{ALL-CNN} offering advantages for more geometric or shape-driven alternatives. Their combination, however, provides the best overall trade-off across models and sample sizes. The benchmark therefore identifies \pkg{ALL-CNN-MLP} as preferred default recommendation for independence testing problems.

\section{Conclusion}\label{sec:ccl}

This paper introduced \emph{deep-testing}, a deep learning perspective on hypothesis testing, in which a test statistic is learned from simulated data through binary classification and then calibrated under the null by Monte Carlo to obtain a near-exact test. As a proof of concept, the approach was investigated in detail for testing independence between two continuous univariate random variables, using three input representations: a greyscale image of the scatter plot, a vector of dependence indicators, and a hybrid combination of the two. Crucially, because these representations are margin-free, a single universal network can be trained to evaluate samples of any size, requiring only a rapid, sample-size-specific null calibration at test time.

\ppn The empirical results show that this strategy is highly effective. In a large-scale power study, the hybrid procedure \pkg{ALL-CNN-MLP} emerges as the strongest overall method: it attains the highest average power across sample sizes, remains closest on average to the best-performing competitor across dependence models, and is the least vulnerable to substantial losses on more difficult alternatives. These findings support \pkg{ALL-CNN-MLP} as a natural default recommendation in the independence-testing setting considered here. The results further indicate that the other deep-tests are very competitive as well, confirming that even the lightweight \pkg{ALL-MLP}, with fewer than 2,000 parameters, delivers strong performance by fully exploiting the dependence indicator representation.

\ppn More broadly, these findings validate the proof of concept. Once trained on a sufficiently rich collection of synthetic dependence structures, a deep-learning classifier can be turned into a near-exact test of independence with strong and broadly reliable power.

\ppn Several directions for future work follow naturally from this study. A priority is to develop a more general theoretical understanding of deep-testing, so as to clarify its statistical foundations and support the development of a broader methodological framework. This should naturally include extensions to multivariate margins, discrete or mixed-type data, and the notoriously difficult problem of conditional independence testing. Ultimately, this research program could help establish deep-testing as a versatile and computationally efficient toolbox for modern statistical inference.

\onecolumn

%==================================================
%==================================================
%==================================================
%==================================================
%==================================================
%==================================================
%==================================================
%==================================================
%==================================================

\appendix

\subsection{Generation of the training sets} \label{app:training} 

Here we provide the essential algorithms -- that is, the data-generating-process (DGP) formulae -- used to generate independent samples (`units') of \(n\) pairs of i.i.d.\ observations \(\{(z_{1k},z_{2k})\}_{k=1}^n\) from each of the 20 dependence models depicted in Figures~\ref{fig:train1} and~\ref{fig:train2}. The training sets are generated randomly from these DGPs using different initial seeds. Although they share the same underlying dependence structures, the generated samples largely differ through the randomisation of the model parameters, and, in general, their exposure to different noise levels. The sample sizes considered are $n\in\{50, 100, 200, 400\}$. 

\ppn For each of the following 20 models of dependence, $\tilde{N}=1,000$ training samples were generated (which makes up both the $20,000$ samples generated `under $H_1$' mentioned in Section~\ref{subsec:train}. These $\tilde{N}$ samples are independent, but not identically distributed: we actually generated them using three different types of additive noise structure (implemented via the R functions \pkg{depstats::normnoise()} and \pkg{depstats::varnormnoise()}): $\tilde{N}_1=600$ with noise structure 1, $\tilde{N}_2=200$ with noise structure 2 and $\tilde{N}_3=200$ with noise structure 3 (so that $\tilde{N}=\tilde{N}_1+\tilde{N}_2+\tilde{N}_3$). For some models, we apply a random rotation implemented in the \pkg{depstats::rotate()} function. All details about these `noise structures' are to be found below. Let $\mathcal{L}_1=\{1,\ldots,\tilde{N}_1\}$, $\mathcal{L}_2=\{\tilde{N}_1+1,\ldots,\tilde{N}_1+\tilde{N}_2\}$ and $\mathcal{L}_3=\{\tilde{N}_1+\tilde{N}_2+1,\ldots,\tilde{N}\}$. For every $\ell\in\mathcal{L}:=\{1,\ldots,\tilde{N}\}=\mathcal{L}_1\cup\mathcal{L}_2\cup\mathcal{L}_3$, we thus generated the $n$ i.i.d.\ pairs of observations  $(Z_{1,k}^{(\ell)},Z_{2,k}^{(\ell)})$, $k=1,\ldots,n$ as follows. (Unless stated otherwise, all intermediate random variables used to obtain values of $(Z_{1,k}^{(\ell)},Z_{2,k}^{(\ell)})$ are generated independently from one another.)

\begin{itemize}
\vspace{1mm}
\item \textbf{Model 1 -- Linear.} (Top left of Figure~\ref{fig:train1}.) For $\ell\in\mathcal{L}$, we take
$$
Z_{1,k}^{(\ell)} \sim \Us_{[-1,1]} \text{ and } Z_{2,k}^{(\ell)} = \beta^{(\ell)}Z_{1,k}^{(\ell)} + \epsilon_k^{(\ell)},
$$
where the $\beta^{(\ell)}$'s are i.i.d.\ from the mixture distribution $0.5\Us_{[-2,-0.5]}+0.5\Us_{[0.5,2]}$ and where the $\epsilon_k^{(\ell)}$'s are generated independently from a $\Ns\left(0, \sigma_k^{(\ell)}\right)$ distribution. For a training set (resp. a test set), we take
$$
\sigma_k^{(\ell)}=\left\{
\begin{array}{ll}
\sigma^{(\ell)}, \text{ generated from a }\Us_{[0,1]} \,(\text{resp. } \Us_{[0,2]}) \text{ distribution}, & \text{if } \ell\in\mathcal{L}_1, \\
0.5|Z_{1,k}^{(\ell)}|\quad(\text{resp. }|0.75 + 0.75Z_{1,k}^{(\ell)}|), & \text{if } \ell\in\mathcal{L}_2, \\
\left|0.5- 0.5|Z_{1,k}^{(\ell)}|\right|\quad(\text{resp. }\left|1.5- 0.75|Z_{1,k}^{(\ell)}|\right|), & \text{if } \ell\in\mathcal{L}_3.
\end{array}
\right.
$$
\vspace{1mm}

The structural portion $Z_2=\beta Z_1$ is implemented in the R function  \pkg{depstats::randlin()}.

\vspace{1mm}

\item \textbf{Model 2 -- Diamond.} ($2^\text{nd}$ on top row of Figure~\ref{fig:train1}.) For $\ell\in\mathcal{L}$, we take
$$
Z_{1,k}^{(\ell)} = U_k^{(\ell)} \cos \big(\theta^{(\ell)}\big) - V_k^{(\ell)} \sin \big(\theta^{(\ell)}\big)\text{ and } Z_{2,k}^{(\ell)} = U_k^{(\ell)} \sin \big(\theta^{(\ell)}\big) + V_k^{(\ell)} \cos \big(\theta^{(\ell)}\big) + \epsilon_k^{(\ell)},
$$
where the $U_k^{(\ell)},V_k^{(\ell)}$'s are i.i.d.\ $\Us_{[-1,1]}$,   the $\theta^{(\ell)}$'s are i.i.d.\ $\Us_{[\pi/6,\pi/3]}$ and where the $\epsilon_k^{(\ell)}$'s are generated independently from a $\mathcal{N}\left(0, \sigma_k^{(\ell)}\right)$ distribution. For a training set (resp. a test set), we take
$$
\sigma_k^{(\ell)}=\left\{
\begin{array}{ll}
\sigma^{\ell}, \text{ generated from a }\Us_{[0,0.25]} \,(\text{resp. } \Us_{[0,0.5]}) \text{ distribution}, & \text{if } \ell\in\mathcal{L}_1,\\
0.2|Z_{1,k}^{(\ell)}|\quad(\text{resp. }|0.25 + 0.25Z_{1,k}^{(\ell)}|), & \text{if } \ell\in\mathcal{L}_2, \\
\left|0.2- 0.2|Z_{1,k}^{(\ell)}|\right|\quad(\text{resp. }\left|0.5- 0.25|Z_{1,k}^{(\ell)}|\right|), & \text{if } \ell\in\mathcal{L}_3.
\end{array}
\right.
$$
\vspace{1mm}
The structural portion $(Z_1,Z_2)^\top=\left(\begin{array}{cc}\cos(\theta) & -\sin(\theta)\\
\sin(\theta) & \cos(\theta)\end{array}\right)(U,V)^\top$ is implemented in the R function  \pkg{depstats::diam.cld()}.

\vspace{1mm}

\item \textbf{Model 3 -- Triangle.} ($3^\text{rd}$ on top row of Figure~\ref{fig:train1}.) For $\ell\in\mathcal{L}$, we take
$$
\left(Z_{1,k}^{(\ell)},Z_{2,k}^{(\ell)}\right)^\top = \Theta^{(\ell)} \left(U_k^{(\ell)}V_k^{(\ell)}, U_k^{(\ell)} + \epsilon_k^{(\ell)}\right)^\top,
$$
where $\Theta^{(\ell)}$ is the 2D rotation matrix  with angle $\theta^{(\ell)}\sim\Us_{[0,2\pi]}$, and where $U_k^{(\ell)}\sim\Us_{[0,1]}$,  $V_k^{(\ell)}\sim\Us_{[-l^{(\ell)},r^{(\ell)}]}$,  $l^{(\ell)}\sim\Us_{[0,1]}$, $r^{(\ell)}\sim\Us_{[0,1]}$ and where $\epsilon_k^{(\ell)}\sim\mathcal{N}\left(0, \sigma_{k}^{(\ell)}\right)$. For a training set (resp. a test set), we take
$$
\sigma_{k}^{(\ell)}=\left\{
\begin{array}{ll}
\sigma^{(\ell)}, \text{ generated from a }\Us_{[0,0.5]} \,(\text{resp. } \Us_{[0,0.5]}) \text{ distribution}, & \text{if } \ell\in\mathcal{L}_1, \\
0.3|Z_{1,k}^{(\ell)}|\quad(\text{resp. }|0.25 + 0.25Z_{1,k}^{(\ell)}|), & \text{if } \ell\in\mathcal{L}_2, \\
\left|0.3- 0.3|Z_{1,k}^{(\ell)}|\right|\quad(\text{resp. }\left|0.5- 0.25|Z_{1,k}^{(\ell)}|\right|), & \text{if } \ell\in\mathcal{L}_3.
\end{array}
\right.
$$
\vspace{1mm}
The structural portion is implemented in the R function \pkg{depstats::tri.cld()}. 

\vspace{1mm}

\item \textbf{Model 4 -- Crescent.} ($4^\text{th}$ on top row of Figure~\ref{fig:train1}.) For a training set (resp. a test set), we take
$$
\left(Z_{1,k}^{(\ell)},Z_{2,k}^{(\ell)}\right)^\top = \Theta^{(\ell)} \left(U_k^{(\ell)}, V_k^{(\ell)} + \epsilon_k^{(\ell)}\right)^\top\quad\left(\text{resp. }\left(Z_{1,k}^{(\ell)},Z_{2,k}^{(\ell)}\right)^\top = \Theta^{(\ell)} \left(U_k^{(\ell)}, V_k^{(\ell)}\right)^\top+\left(0,\epsilon_k^{(\ell)}\right)^\top\right),
$$
where $\Theta^{(\ell)}$ is the 2D rotation matrix  with angle $\theta^{(\ell)}\sim\Us_{[0,2\pi]}$,  $V_k^{(\ell)}\sim\Us_{[-1,c^{(\ell)}/2]}$ with $c^{(\ell)}\sim\Us_{[0.5,1]}$ (resp.\ $\Us_{[0.25,1]}$),
$$
U_k^{(\ell)}\sim\left\{
\begin{array}{ll}
\Us_{\left[-\sqrt{1-V_k^{(\ell)^2}},\sqrt{1-V_k^{(\ell)^2}}\right]}, & \text{if }V_k^{(\ell)}<c^{(\ell)}-1,\\
\text{the mixture }0.5\Us_{\left[-\sqrt{1-V_k^{(\ell)^2}},-\sqrt{1-\left(V_k^{(\ell)}-c^{(\ell)}\right)^2}\right]}+0.5\Us_{\left[\sqrt{1-\left(V_k^{(\ell)}-c^{(\ell)}\right)^2},\sqrt{1-V_k^{(\ell)^2}}\right]}, & \text{if }V_k^{(\ell)}\geq c^{(\ell)}-1,\\
\end{array}
\right.
$$
and $\epsilon_k^{(\ell)}\sim
\mathcal{N}\left(0, \sigma_{k}^{(\ell)}\right)$. We take
$$
\sigma_{k}^{(\ell)}=\left\{
\begin{array}{ll}
\sigma^{(\ell)}, \text{ generated from a }\Us_{[0,0.5]} \,(\text{resp. } \Us_{[0,1]}) \text{ distribution}, & \text{if } \ell\in\mathcal{L}_1, \\
0.3|Z_{1,k}^{(\ell)}|\quad(\text{resp. }|0.25 + 0.25Z_{1,k}^{(\ell)}|), & \text{if } \ell\in\mathcal{L}_2, \\
\left|0.3- 0.3|Z_{1,k}^{(\ell)}|\right|\quad(\text{resp. }\left|0.5- 0.25|Z_{1,k}^{(\ell)}|\right|), & \text{if } \ell\in\mathcal{L}_3.
\end{array}
\right.
$$
\vspace{1mm}
The structural portion is implemented in the R function \pkg{depstats::cre.cld()}.

\item \textbf{Model 5 -- Point clouds.} ($5^\text{th}$ on top row of Figure~\ref{fig:train1}.) We take
$$
\left(Z_{1,k}^{(\ell)},Z_{2,k}^{(\ell)}\right)^\top = \Theta^{(\ell)}\left(U_k^{(\ell)} + \epsilon_{1,k}^{(\ell)}, V_k^{(\ell)} + \epsilon_{2,k}^{(\ell)}\right)^\top,
$$
where $\Theta^{(\ell)}$ is the 2D rotation matrix  with angle $\theta^{(\ell)}\sim\Us_{[0,2\pi]}$,  
$U_k^{(\ell)}$ is sampled uniformly with replacement from $\left\{\tilde{U}_1^{(\ell)},\ldots,\tilde{U}_n^{(\ell)}\right\}$, $V_k^{(\ell)}$ is sampled uniformly with replacement from $\left\{\tilde{V}_1^{(\ell)},\ldots,\tilde{V}_n^{(\ell)}\right\}$, and $\left(\tilde{U}_k^{(\ell)}, \tilde{V}_k^{(\ell)}\right)$ is 
sampled uniformly from $S$ where 
$$
S = \left\{
\begin{array}{ll}
\left\{ (1,0), (-1,0), (0,1) \right\}, &\text{w.p.} \, \, \frac{1}{3}, \\
\left\{ (1,0), (-1,0), (0,0), (0,1), (0,-1) \right\}, & \text{w.p.} \, \, \frac{1}{3}, \\
\left\{ (1,1), (-1,1), (0,0), (1,-1), (-1,-1) \right\}, & \text{w.p.} \, \, \frac{1}{3}.
\end{array}
\right.
$$
We take $\epsilon_{1,k}^{(\ell)}\sim
\mathcal{N}\left(0, \sigma^{(\ell)}\right)$ and $\epsilon_{2,k}^{(\ell)}\sim
\mathcal{N}\left(0, \sigma^{(\ell)}\right)$ where
$\sigma^{(\ell)}$ is generated from a $\Us_{[0.05,0.5]}$ distribution. For this model, there is only one single structure of noise which is used for both the training and test sets.

\vspace{1mm}

The structural portion is implemented in the R function \pkg{depstats::pt.cld()}.

\vspace{1mm}

\item \textbf{Model 6 -- Exponential.} ($1^\text{st}$ on bottom row of Figure~\ref{fig:train1}.) For a training set (resp. a test set), we take
$$
\left(Z_{1,k}^{(\ell)},Z_{2,k}^{(\ell)}\right)^\top = \Theta^{(\ell)} \left(U_k^{(\ell)}, V_k^{(\ell)} + \epsilon_k^{(\ell)}\right)^\top\quad\left(\text{resp. }\left(Z_{1,k}^{(\ell)},Z_{2,k}^{(\ell)}\right)^\top = \Theta^{(\ell)} \left(U_k^{(\ell)}, V_k^{(\ell)}\right)^\top+\left(0,\epsilon_k^{(\ell)}\right)^\top\right),
$$
where $\Theta^{(\ell)}$ is the 2D rotation matrix  with angle $\theta^{(\ell)}\sim\Us_{[0,2\pi]}$,  $U_k^{(\ell)}\sim\Us_{[-1,1]}$, $V_k^{(\ell)}=b^{(\ell)^{U_k^{(\ell)}}}$ 
and $\epsilon_k^{(\ell)}\sim
\mathcal{N}\left(0, \sigma_{k}^{(\ell)}\right)$. We take
$$
b^{(\ell)}\sim\left\{
\begin{array}{ll}
\Us_{[2,4]}\quad(\text{resp. }\Us_{[2,4]}), & \text{if } \ell\in\mathcal{L}_1\cup\mathcal{L}_2, \\
\Us_{[1.5,3]}\quad(\text{resp. }\Us_{[2,4]}), & \text{if } \ell\in\mathcal{L}_3,
\end{array}
\right.
$$
and
$$
\sigma_{k}^{(\ell)}=\left\{
\begin{array}{ll}
\sigma^{(\ell)}, \text{ generated from a }\Us_{[0,0.5]} \,(\text{resp. } \Us_{[0,2}) \text{ distribution}, & \text{if } \ell\in\mathcal{L}_1, \\
0.3|Z_{1,k}^{(\ell)}|\quad(\text{resp. }|0.5 + 0.5Z_{1,k}^{(\ell)}|), & \text{if } \ell\in\mathcal{L}_2, \\
\left|0.3- 0.3|Z_{1,k}^{(\ell)}|\right|\quad(\text{resp. }\left|1- 0.5|Z_{1,k}^{(\ell)}|\right|), & \text{if } \ell\in\mathcal{L}_3.
\end{array}
\right.
$$
\vspace{1mm}
The structural portion is implemented in the R function \pkg{depstats::exp.cv()}.

\vspace{1mm}

\item \textbf{Model 7 -- Circles.} ($2^\text{nd}$ on bottom row of Figure~\ref{fig:train1}.) We take
$$
\left(Z_{1,k}^{(\ell)},Z_{2,k}^{(\ell)}\right)^\top = \Theta^{(\ell)} \left(\sin\left(\pi W_k^{(\ell)}\right) + U_k^{(\ell)}, \cos\left(\pi W_k^{(\ell)}\right) + V_k^{(\ell)} + \epsilon_{k}^{(\ell)}\right)^\top,
$$
where $\Theta^{(\ell)}$ is the 2D rotation matrix  with angle $\theta^{(\ell)}\sim\Us_{[0,2\pi]}$, $W_k^{(\ell)}\sim\Us_{[-1,1]}$,  
$U_k^{(\ell)}$ is sampled uniformly with replacement from $\left\{\tilde{U}_1^{(\ell)},\ldots,\tilde{U}_n^{(\ell)}\right\}$, $V_k^{(\ell)}$ is sampled uniformly with replacement from $\left\{\tilde{V}_1^{(\ell)},\ldots,\tilde{V}_n^{(\ell)}\right\}$, and $\left(\tilde{U}_k^{(\ell)}, \tilde{V}_k^{(\ell)}\right)$ is 
sampled uniformly from $S$ where 
$$
S = \left\{
\begin{array}{ll}
\left\{ (0,0) \right\}, &\text{w.p.} \, \, \frac{1}{3}, \\
\left\{ (1,0), (-1,0), (0,1)\right\}, & \text{w.p.} \, \, \frac{1}{3}, \\
\left\{ (1,1), (-1,1), (1,-1), (-1,-1) \right\}, & \text{w.p.} \, \, \frac{1}{3}.
\end{array}
\right.
$$
We take $\epsilon_{k}^{(\ell)}\sim
\mathcal{N}\left(0, \sigma_{k}^{(\ell)}\right)$ with  for a training set (resp. a test set), 
$$
\sigma_{k}^{(\ell)}=\left\{
\begin{array}{ll}
\sigma^{(\ell)}, \text{ generated from a }\Us_{[0,0.5]} \,(\text{resp. } \Us_{[0,1}) \text{ distribution}, & \text{if } \ell\in\mathcal{L}_1, \\
0.3|Z_{1,k}^{(\ell)}|\quad(\text{resp. } 0.5|Z_{1,k}^{(\ell)}|), & \text{if } \ell\in\mathcal{L}_2, \\
\left|0.3- 0.3|Z_{1,k}^{(\ell)}|\right|\quad(\text{resp. }\left|0.5- 0.5|Z_{1,k}^{(\ell)}|\right|), & \text{if } \ell\in\mathcal{L}_3.
\end{array}
\right.
$$

\vspace{1mm}

The structural portion is implemented in the R function \pkg{depstats::circ.cv()}.

\vspace{1mm}

\item \textbf{Model 8 -- Cross.} ($3^\text{rd}$ on bottom row of Figure~\ref{fig:train1}.) We take
$$
\left(Z_{1,k}^{(\ell)},Z_{2,k}^{(\ell)}\right)^\top = \Theta^{(\ell)} \left(U_k^{(\ell)}, V_k^{(\ell)}\right)^\top+\left(0,\epsilon_k^{(\ell)}\right)^\top,
$$
where $\Theta^{(\ell)}$ is the 2D rotation matrix  with angle $\theta^{(\ell)}\sim\Us_{[-\pi/6,\pi/6]}$,  $U_k^{(\ell)}\sim\Us_{[-1,1]}$, $V_k^{(\ell)}=W_k^{(\ell)} U_k^{(\ell)}$, $W_k^{(\ell)}\sim2\mathcal{B}\textrm{er}(0.5)-1$ 
and $\epsilon_k^{(\ell)}\sim
\mathcal{N}\left(0, \sigma_{k}^{(\ell)}\right)$.  For a training set (resp. a test set), we take
$$
\sigma_{k}^{(\ell)}=\left\{
\begin{array}{ll}
\sigma^{(\ell)}, \text{ generated from a }\Us_{[0,0.5]} \,(\text{resp. } \Us_{[0,1]}) \text{ distribution}, & \text{if } \ell\in\mathcal{L}_1, \\
0.3|Z_{1,k}^{(\ell)}|\quad(\text{resp. }|0.25 + 0.25Z_{1,k}^{(\ell)}|), & \text{if } \ell\in\mathcal{L}_2, \\
\left|0.3- 0.3|Z_{1,k}^{(\ell)}|\right|\quad(\text{resp. }\left|0.5- 0.25|Z_{1,k}^{(\ell)}|\right|), & \text{if } \ell\in\mathcal{L}_3.
\end{array}
\right.
$$
\vspace{1mm}
The structural portion is implemented in the R function \pkg{depstats::cross.cv()}.

\vspace{1mm}

\item \textbf{Model 9 -- Wedge.} ($4^\text{th}$ on bottom row of Figure~\ref{fig:train1}.)  For a training set (resp. a test set), we take
$$
\left(Z_{1,k}^{(\ell)},Z_{2,k}^{(\ell)}\right)^\top = \Theta^{(\ell)} \left(U_k^{(\ell)}, V_k^{(\ell)} + \epsilon_k^{(\ell)}\right)^\top\quad\left(\text{resp. }\left(Z_{1,k}^{(\ell)},Z_{2,k}^{(\ell)}\right)^\top = \Theta^{(\ell)} \left(U_k^{(\ell)}, V_k^{(\ell)}\right)^\top+\left(0,\epsilon_k^{(\ell)}\right)^\top\right),
$$
where $\Theta^{(\ell)}$ is the 2D rotation matrix  with angle $\theta^{(\ell)}\sim\Us_{[-\pi/6,\pi/6]}$,  $U_k^{(\ell)}\sim\Us_{[-1,1]}$, $V_k^{(\ell)}=2\left(U_k^{(\ell)}+1\right)W_k^{(\ell)}$, $W_k^{(\ell)}\sim2\mathcal{B}\textrm{er}(0.5)-1$ 
and $\epsilon_k^{(\ell)}\sim
\mathcal{N}\left(0, \sigma_{k}^{(\ell)}\right)$.  We take
$$
\sigma_{k}^{(\ell)}=\left\{
\begin{array}{ll}
\sigma^{(\ell)}, \text{ generated from a }\Us_{[0,1]} \,(\text{resp. } \Us_{[2,4}) \text{ distribution}, & \text{if } \ell\in\mathcal{L}_1, \\
0.5|Z_{1,k}^{(\ell)}|\quad(\text{resp. }|2 + Z_{1,k}^{(\ell)}|), & \text{if } \ell\in\mathcal{L}_2, \\
\left|0.5- 0.5|Z_{1,k}^{(\ell)}|\right|\quad(\text{resp. }\left|2- |Z_{1,k}^{(\ell)}|\right|), & \text{if } \ell\in\mathcal{L}_3.
\end{array}
\right.
$$
\vspace{1mm}
The structural portion is implemented in the R function \pkg{depstats::wedge.cv()}. 

\vspace{1mm}

\item \textbf{Model 10 -- Cubic.} ($5^\text{th}$ on bottom row of Figure~\ref{fig:train1}.)  For a training set (resp. a test set), we take
$$
\left(Z_{1,k}^{(\ell)},Z_{2,k}^{(\ell)}\right)^\top = \Theta^{(\ell)} \left(U_k^{(\ell)}, V_k^{(\ell)} + \epsilon_k^{(\ell)}\right)^\top\quad\left(\text{resp. }\left(Z_{1,k}^{(\ell)},Z_{2,k}^{(\ell)}\right)^\top = \Theta^{(\ell)} \left(U_k^{(\ell)}, V_k^{(\ell)}\right)^\top+\left(0,\epsilon_k^{(\ell)}\right)^\top\right),
$$
where $\Theta^{(\ell)}$ is the 2D rotation matrix  with angle $\theta^{(\ell)}\sim\Us_{[0,2\pi]}$,  $U_k^{(\ell)}\sim\Us_{[-1,1]}$, $V_k^{(\ell)}=U_k^{(\ell)^3}-U_k^{(\ell)}$, 
and $\epsilon_k^{(\ell)}\sim
\mathcal{N}\left(0, \sigma_{k}^{(\ell)}\right)$.  We take
$$
\sigma_{k}^{(\ell)}=\left\{
\begin{array}{ll}
\sigma^{(\ell)}, \text{ generated from a }\Us_{[0,0.25]} \,(\text{resp. } \Us_{[0,1}) \text{ distribution}, & \text{if } \ell\in\mathcal{L}_1, \\
0.2|Z_{1,k}^{(\ell)}|\quad(\text{resp. }|0.25 + 0.25Z_{1,k}^{(\ell)}|), & \text{if } \ell\in\mathcal{L}_2, \\
\left|0.2- 0.2|Z_{1,k}^{(\ell)}|\right|\quad(\text{resp. }\left|0.5- 0.25|Z_{1,k}^{(\ell)}|\right|), & \text{if } \ell\in\mathcal{L}_3.
\end{array}
\right.
$$
\vspace{1mm}
The structural portion is implemented in the R function \pkg{depstats::cubic.cv()}. 

\vspace{1mm}

\item \textbf{Model 11 -- W-shape.} ($1^\text{st}$ on top row of Figure~\ref{fig:train2}.)  For a training set (resp. a test set), we take
$$
(Z_{1,i},Z_{2,i})^\top = \Theta_\ell (U_i, V_i + \epsilon_i)^\top\quad\left(\text{resp. }(Z_{1,i},Z_{2,i})^\top = \Theta_\ell (U_i, V_i)^\top+(0,\epsilon_i)^\top\right),
$$
where $\Theta_\ell$ is the 2D rotation matrix  with angle $\theta_\ell\sim\Us_{[0,2\pi]}$,  $U_i\sim\Us_{[-1,1]}$, $V_i=4\left(U_i^2-0.5\right)^2$, 
and $\epsilon_i\sim
\text{N}(0, \sigma_{\ell_i})$.  We take
$$
\sigma_{\ell_i}=\left\{
\begin{array}{ll}
\sigma_{\ell}, \text{ generated from a }\Us_{[0,0.3]} \,(\text{resp. } \Us_{[0,1}) \text{ distribution}, & \text{for } \ell\in\mathcal{L}_1, \\
0.2|Z_{1,i}|\quad(\text{resp. }|0.5 + 0.5Z_{1,i}|), & \text{for } \ell\in\mathcal{L}_2, \\
\left|0.2- 0.2|Z_{1,i}|\right|\quad(\text{resp. }\left|1- 0.5|Z_{1,i}|\right|), & \text{for } \ell\in\mathcal{L}_3.
\end{array}
\right.
$$
\vspace{1mm}
The structural portion is implemented in the R function \pkg{depstats::w.cv()}. 

\vspace{1mm}

\item \textbf{Model 12 -- Parabola.} ($2^\text{nd}$ on top row of Figure~\ref{fig:train2}.)  For a training set (resp. a test set), we take
$$
(Z_{1,i},Z_{2,i})^\top = \Theta_\ell (U_i, V_i + \epsilon_i)^\top\quad\left(\text{resp. }(Z_{1,i},Z_{2,i})^\top = \Theta_\ell (U_i, V_i)^\top+(0,\epsilon_i)^\top\right),
$$
where $\Theta_\ell$ is the 2D rotation matrix  with angle $\theta_\ell\sim\Us_{[0,2\pi]}$,  $U_i\sim\Us_{[-1,1]}$, $V_i=0.5U_i^2$, 
and $\epsilon_i\sim
\text{N}(0, \sigma_{\ell_i})$.  We take
$$
\sigma_{\ell_i}=\left\{
\begin{array}{ll}
\sigma_{\ell}, \text{ generated from a }\Us_{[0,0.3]} \,(\text{resp. } \Us_{[0,1}) \text{ distribution}, & \text{for } \ell\in\mathcal{L}_1, \\
0.2|Z_{1,i}|\quad(\text{resp. }|0.25 + 0.25Z_{1,i}|), & \text{for } \ell\in\mathcal{L}_2, \\
\left|0.2- 0.2|Z_{1,i}|\right|\quad(\text{resp. }\left|0.5- 0.25|Z_{1,i}|\right|), & \text{for } \ell\in\mathcal{L}_3.
\end{array}
\right.
$$
\vspace{1mm}
The structural portion is implemented in the R function \pkg{depstats::parabola.cv()}. 

\vspace{1mm}

\item \textbf{Model 13 -- Two-parabola.} ($3^\text{rd}$ on top row of Figure~\ref{fig:train2}.)  For a training set (resp. a test set), we take
$$
(Z_{1,i},Z_{2,i})^\top = \Theta_\ell (U_i, V_i + \epsilon_i)^\top\quad\left(\text{resp. }(Z_{1,i},Z_{2,i})^\top = \Theta_\ell (U_i, V_i)^\top+(0,\epsilon_i)^\top\right),
$$
where $\Theta_\ell$ is the 2D rotation matrix  with angle $\theta_\ell\sim\Us_{[0,2\pi]}$,  $U_i\sim\Us_{[-1,1]}$, $V_i=W_iU_i^2$, $W_i\sim2\mathcal{B}\textrm{er}(0.5)-1$ 
and $\epsilon_i\sim
\text{N}(0, \sigma_{\ell_i})$.  We take
$$
\sigma_{\ell_i}=\left\{
\begin{array}{ll}
\sigma_{\ell}, \text{ generated from a }\Us_{[0,0.4]} \,(\text{resp. } \Us_{[0,1}) \text{ distribution}, & \text{for } \ell\in\mathcal{L}_1, \\
0.3|Z_{1,i}|\quad(\text{resp. }|0.25 + 0.25Z_{1,i}|), & \text{for } \ell\in\mathcal{L}_2, \\
\left|0.3- 0.3|Z_{1,i}|\right|\quad(\text{resp. }\left|0.5- 0.25|Z_{1,i}|\right|), & \text{for } \ell\in\mathcal{L}_3.
\end{array}
\right.
$$
\vspace{1mm}
The structural portion is implemented in the R function \pkg{depstats::twoparabola.cv()}. 

\vspace{1mm}

\item \textbf{Model 14 -- Sine.} ($4^\text{th}$ on top row of Figure~\ref{fig:train2}.)  For a training set (resp. a test set), we take
$$
(Z_{1,i},Z_{2,i})^\top = \Theta_\ell (U_i, V_i + \epsilon_i)^\top\quad\left(\text{resp. }(Z_{1,i},Z_{2,i})^\top = \Theta_\ell (U_i, V_i)^\top+(0,\epsilon_i)^\top\right),
$$
where $\Theta_\ell$ is the 2D rotation matrix  with angle $\theta_\ell\sim\Us_{[0,2\pi]}$,  $U_i\sim\Us_{[-1,1]}$, $V_i=\sin(S_\ell\pi U_i)$, $S_\ell\sim\Us_{[1,4]}$ (resp. $S_\ell\sim\Us_{[1,4]}$) 
and $\epsilon_i\sim
\text{N}(0, \sigma_{\ell_i})$.  We take
$$
\sigma_{\ell_i}=\left\{
\begin{array}{ll}
\sigma_{\ell}, \text{ generated from a }\Us_{[0,1]} \,(\text{resp. } \Us_{[0,1}) \text{ distribution}, & \text{for } \ell\in\mathcal{L}_1, \\
0.5|Z_{1,i}|\quad(\text{resp. }|0.25 + 0.25Z_{1,i}|), & \text{for } \ell\in\mathcal{L}_2, \\
\left|0.5- 0.5|Z_{1,i}|\right|\quad(\text{resp. }\left|0.5- 0.25|Z_{1,i}|\right|), & \text{for } \ell\in\mathcal{L}_3.
\end{array}
\right.
$$
\vspace{1mm}
The structural portion is implemented in the R function \pkg{depstats::sine.cv()}. 

\vspace{1mm}

\item \textbf{Model 15 -- Doppler.} ($5^\text{th}$ on top row of Figure~\ref{fig:train2}.)  For a training set (resp. a test set), we take
$$
(Z_{1,i},Z_{2,i})^\top = \Theta_\ell (U_i, V_i + \epsilon_i)^\top\quad\left(\text{resp. }(Z_{1,i},Z_{2,i})^\top = \Theta_\ell (U_i, V_i)^\top+(0,\epsilon_i)^\top\right),
$$
where $\Theta_\ell$ is the 2D rotation matrix  with angle $\theta_\ell\sim\Us_{[0,2\pi]}$,  $U_i\sim\Us_{[-1,1]}$, $V_i=\sin\left(-f_\ell(1+U_i)(2+U_i)\right)$, $f_\ell\sim\Us_{[2,4]}$ (resp. $f_\ell\sim\Us_{[2,4]}$) 
and $\epsilon_i\sim
\text{N}(0, \sigma_{\ell_i})$.  We take
$$
\sigma_{\ell_i}=\left\{
\begin{array}{ll}
\sigma_{\ell}, \text{ generated from a }\Us_{[0,1]} \,(\text{resp. } \Us_{[0,1}) \text{ distribution}, & \text{for } \ell\in\mathcal{L}_1, \\
0.5|Z_{1,i}|\quad(\text{resp. }|0.25 + 0.25Z_{1,i}|), & \text{for } \ell\in\mathcal{L}_2, \\
\left|0.5- 0.5|Z_{1,i}|\right|\quad(\text{resp. }\left|0.5- 0.25|Z_{1,i}|\right|), & \text{for } \ell\in\mathcal{L}_3.
\end{array}
\right.
$$
\vspace{1mm}
The structural portion is implemented in the R function \pkg{depstats::doppler.cv()}. 

\vspace{1mm}

\item \textbf{Model 16 -- Heavy-sine.} ($1^\text{st}$ on bottom row of Figure~\ref{fig:train2}.)  For a training set (resp. a test set), we take
$$
(Z_{1,i},Z_{2,i})^\top = \Theta_\ell (U_i, V_i + \epsilon_i)^\top\quad\left(\text{resp. }(Z_{1,i},Z_{2,i})^\top = \Theta_\ell (U_i, V_i)^\top+(0,\epsilon_i)^\top\right),
$$
where $\Theta_\ell$ is the 2D rotation matrix  with angle $\theta_\ell\sim\Us_{[0,2\pi]}$,  $U_i\sim\Us_{[-1,1]}$, $V_i=\sin(4\pi U_i)/2 - \textrm{sign}(U_i+0.6)/8 - \textrm{sign}(U_i+0.2)/8 - \textrm{sign}(U_i-0.2)/8 - \textrm{sign}(U_i-0.6)/8$  (where $\textrm{sign}(x)=-1$ if $x<0$, $0$ if $x=0$ and $1$ if $x>0$) 
and $\epsilon_i\sim
\text{N}(0, \sigma_{\ell_i})$.  We take
$$
\sigma_{\ell_i}=\left\{
\begin{array}{ll}
\sigma_{\ell}, \text{ generated from a }\Us_{[0,1]} \,(\text{resp. } \Us_{[0,1}) \text{ distribution}, & \text{for } \ell\in\mathcal{L}_1, \\
0.5|Z_{1,i}|\quad(\text{resp. }|0.25 + 0.25Z_{1,i}|), & \text{for } \ell\in\mathcal{L}_2, \\
\left|0.5- 0.5|Z_{1,i}|\right|\quad(\text{resp. }\left|0.5- 0.25|Z_{1,i}|\right|), & \text{for } \ell\in\mathcal{L}_3.
\end{array}
\right.
$$
\vspace{1mm}
The structural portion is implemented in the R function \pkg{depstats::heavysine.cv()}. 

\vspace{1mm}

\item \textbf{Model 17 -- Heart.} ($2^\text{nd}$ on bottom row of Figure~\ref{fig:train2}.)  For a training set (resp. a test set), we take
$$
(Z_{1,i},Z_{2,i})^\top = \Theta_\ell (U_i, V_i + \epsilon_i)^\top\quad\left(\text{resp. }(Z_{1,i},Z_{2,i})^\top = \Theta_\ell (U_i, V_i)^\top+(0,\epsilon_i)^\top\right),
$$
where $\Theta_\ell$ is the 2D rotation matrix  with angle $\theta_\ell\sim\Us_{[0,2\pi]}$,  $W_i\sim\Us_{[-1,1]}$, $U_i=0.5\left(2\cos(\pi W_i)-\cos(2\pi W_i)\right)$, $V_i=0.5\left(2\sin(\pi W_i)-\sin(2\pi W_i)\right)$
and $\epsilon_i\sim
\text{N}(0, \sigma_{\ell_i})$.  We take
$$
\sigma_{\ell_i}=\left\{
\begin{array}{ll}
\sigma_{\ell}, \text{ generated from a }\Us_{[0,0.5]} \,(\text{resp. } \Us_{[0,1}) \text{ distribution}, & \text{for } \ell\in\mathcal{L}_1, \\
0.3|Z_{1,i}|\quad(\text{resp. }|0.25 + 0.25Z_{1,i}|), & \text{for } \ell\in\mathcal{L}_2, \\
\left|0.3- 0.3|Z_{1,i}|\right|\quad(\text{resp. }\left|0.5- 0.25|Z_{1,i}|\right|), & \text{for } \ell\in\mathcal{L}_3.
\end{array}
\right.
$$
\vspace{1mm}
The structural portion is implemented in the R function \pkg{depstats::heart.cv()}. 

\vspace{1mm}

\item \textbf{Model 18 -- Spiral.} ($3^\text{nd}$ on bottom row of Figure~\ref{fig:train2}.)  For a training set (resp. a test set), we take
$$
(Z_{1,i},Z_{2,i})^\top = \Theta_\ell (U_i, V_i + \epsilon_i)^\top\quad\left(\text{resp. }(Z_{1,i},Z_{2,i})^\top = \Theta_\ell (U_i, V_i)^\top+(0,\epsilon_i)^\top\right),
$$
where $\Theta_\ell$ is the 2D rotation matrix  with angle $\theta_\ell\sim\Us_{[0,2\pi]}$,  $W_i\sim\Us_{[-k_\ell \pi,k_\ell \pi]}$, $U_i=\exp\left(-W_i/10\right)\cos(W_i)$, $V_i=\exp\left(-W_i/10\right)\sin(W_i)$, $k_\ell\sim\Us_{[2.5,3.5]}$ (resp.\ $k_\ell\sim\Us_{[2.5,3.5]}$)
and $\epsilon_i\sim
\text{N}(0, \sigma_{\ell_i})$.  We take
$$
\sigma_{\ell_i}=\left\{
\begin{array}{ll}
\sigma_{\ell}, \text{ generated from a }\Us_{[0,0.3]} \,(\text{resp. } \Us_{[0,0.5}) \text{ distribution}, & \text{for } \ell\in\mathcal{L}_1, \\
0.2|Z_{1,i}|\quad(\text{resp. }|0.1 + 0.1Z_{1,i}|), & \text{for } \ell\in\mathcal{L}_2, \\
\left|0.2- 0.2|Z_{1,i}|\right|\quad(\text{resp. }\left|0.2- 0.1|Z_{1,i}|\right|), & \text{for } \ell\in\mathcal{L}_3.
\end{array}
\right.
$$
\vspace{1mm}
The structural portion is implemented in the R function \pkg{depstats::spiral.cv()}. 

\vspace{1mm}

\item \textbf{Model 19 -- Taegeuk.} ($4^\text{th}$ on bottom row of Figure~\ref{fig:train2}.)  For a training set (resp. a test set), we take
$$
(Z_{1,i},Z_{2,i})^\top = \Theta_\ell (U_i, V_i + \epsilon_i)^\top\quad\left(\text{resp. }(Z_{1,i},Z_{2,i})^\top = \Theta_\ell (U_i, V_i)^\top+(0,\epsilon_i)^\top\right),
$$
where $\Theta_\ell$ is the 2D rotation matrix  with angle $\theta_\ell\sim\Us_{[0,2\pi]}$,  $W_i\sim\Us_{[-1,1]}$, $U_i=0.5^{|\delta_i|}\sin(\pi W_i)  + 0.5\delta_i$, $V_i=| 0.5\cos(\pi W_i)| \delta_i + \cos(\pi W_i) (1 - |\delta_i|)$, $\delta_i=0$ w.p.\ $1/2$ and $-1$ or $1$ w.p.\ $1/4$, and $\epsilon_i\sim
\text{N}(0, \sigma_{\ell_i})$.  We take
$$
\sigma_{\ell_i}=\left\{
\begin{array}{ll}
\sigma_{\ell}, \text{ generated from a }\Us_{[0,0.3]} \,(\text{resp. } \Us_{[0,1}) \text{ distribution}, & \text{for } \ell\in\mathcal{L}_1, \\
0.2|Z_{1,i}|\quad(\text{resp. }|0.25 + 0.25Z_{1,i}|), & \text{for } \ell\in\mathcal{L}_2, \\
\left|0.2- 0.2|Z_{1,i}|\right|\quad(\text{resp. }\left|0.5- 0.25|Z_{1,i}|\right|), & \text{for } \ell\in\mathcal{L}_3.
\end{array}
\right.
$$
\vspace{1mm}
The structural portion is implemented in the R function \pkg{depstats::taegeuk.cv()}. 

\vspace{1mm}

\item \textbf{Model 20 -- Samtaegeuk.} ($5^\text{th}$ on bottom row of Figure~\ref{fig:train2}.)  For a training set (resp. a test set), we take
$$
(Z_{1,i},Z_{2,i})^\top = \Theta_\ell (U_i, V_i + \epsilon_i)^\top\quad\left(\text{resp. }(Z_{1,i},Z_{2,i})^\top = \Theta_\ell (U_i, V_i)^\top+(0,\epsilon_i)^\top\right),
$$
where $\Theta_\ell$ is the 2D rotation matrix  with angle $\theta_\ell\sim\Us_{[0,2\pi]}$,  $W_i\sim\Us_{[-1,1]}$, $U_i=\cos(W_i\pi) \mathbb{1}_{\delta_i = 0} +
    0.5\cos(W_i\pi/2) \mathbb{1}_{\delta_i = 1} +
    (0.5\cos(W_i\pi/2 + \pi/2 + \pi/6) + \sqrt{3}/4) \mathbb{1}_{\delta_i = 2} +
    (0.5\cos(-W_i\pi/2 - \pi/2 - \pi/6) - \sqrt{3}/4) \mathbb{1}_{\delta_i = 3}$, $V_i=\sin(W_i\pi) \mathbb{1}_{\delta_i = 0} +
    (0.5\sin(W_i\pi/2) - 0.5) \mathbb{1}_{\delta_i = 1} +
    (0.5\sin(W_i\pi/2 + \pi/2 + \pi/6) + 1/4 ) \mathbb{1}_{\delta_i = 2} +
    (0.5\sin(-W_i\pi/2 - \pi/2 - \pi/6) + 1/4) \mathbb{1}_{\delta_i = 3}$, $\delta_i=0$ w.p.\ $1/2$ and 
 $1, 2$ or $3$ w.p.\ $1/6$, and $\epsilon_i\sim
\text{N}(0, \sigma_{\ell_i})$.  We take
$$
\sigma_{\ell_i}=\left\{
\begin{array}{ll}
\sigma_{\ell}, \text{ generated from a }\Us_{[0,0.3]} \,(\text{resp. } \Us_{[0,1}) \text{ distribution}, & \text{for } \ell\in\mathcal{L}_1, \\
0.2|Z_{1,i}|\quad(\text{resp. }|0.25 + 0.25Z_{1,i}|), & \text{for } \ell\in\mathcal{L}_2, \\
\left|0.2- 0.2|Z_{1,i}|\right|\quad(\text{resp. }\left|0.5- 0.25|Z_{1,i}|\right|), & \text{for } \ell\in\mathcal{L}_3.
\end{array}
\right.
$$
In the above, $\mathbb{1}$ denotes the indicator function.
\vspace{1mm}
The structural portion is implemented in the R function \pkg{depstats::samtaegeuk.cv()}. 

\vspace{1mm}
\end{itemize}

\subsection{Generation of the testing sets} \label{app:testing} 

For Experiment 1 (Section \ref{subsec:valid}), we generated 20,000 testing samples from the 20 dependence models described in the previous Appendix \ref{app:training} (1,000 samples per model). Although generated by the same underlying dependence structures, these testing samples may be markedly different to the testing samples through the randomisation of the model parameters, and, in general, their exposure to higher noise levels. Also, we consider the additional unseen sample sizes $n=30$ and $n=300$. 

\ppn For Experiment 2, we generated 12,000 testing samples from 6 additional dependence models, of different nature to those used to generate the training samples -- typical samples of size $n=400$ are shown in Figure~\ref{fig:test}. We first considered Models 21--24 below, and for each of them we generated $\tilde{N}=1,000$ samples under two different parameter settings (A and B), yielding a total of $8,000$ samples generated \textit{under $H_1$} as referenced in Section~\ref{subsec:valid}.  More specifically, for each $\ell\in\mathcal{L}:=\{1,\ldots,\tilde{N}\}$, we generated $n\in\{30, 50, 100, 200, 300, 400\}$ i.i.d.\ observation pairs $(Z_{1,k}^{(\ell)},Z_{2,k}^{(\ell)})$ for $k=1,\ldots,n$, as follows. Unless stated otherwise, all intermediate random variables used in constructing these pairs were generated independently.

\begin{itemize}
\vspace{1mm}

\item \textbf{Model 21 -- Bivariate Laplace.} ($1^{\text{st}}$ on top row of Figure~\ref{fig:test}.) For $\ell\in\mathcal{L}$, we take
      \[
        \bigl(Z_{1,k}^{(\ell)},Z_{2,k}^{(\ell)}\bigr)^{\!\top}
        \;\sim\;\textrm{Bivariate Laplace}(\boldsymbol{0},\Sigma^{(\ell)})
      \]
with scale matrix
\[
        \Sigma^{(\ell)}
        =\begin{pmatrix}
            1 & \rho^{(\ell)}\\[2pt]
            \rho^{(\ell)} & 1
          \end{pmatrix},
      \]
where $\rho^{(\ell)}\sim \Us_{[0.4,0.5]}$ (resp. $\rho^{(\ell)}\sim \Us_{[0.5,0.6]}$).

%\item \textbf{Model 4 -- Tanh.} ($4^{\text{th}}$ on top row of Figure~\ref{fig:test}.) For $\ell\in\mathcal{L}$, we take
%$$
%Z_{1,k}^{(\ell)} \sim \Ns(0,1) \text{ and } Z_{2,k}^{(\ell)} = \tanh(Z_{1,k}^{(\ell)}) + \epsilon_k^{(\ell)},
%$$
%where the $\epsilon_k^{(\ell)}$'s are generated independently from a $\Ns\left(0, \sigma^{(\ell)}\right)$ distribution with $\sigma^{(\ell)}\sim\Us(2,3)$ (resp. $\sigma^{(\ell)}\sim\Us(3,4)$).

\item \textbf{Model 22 -- Ishigami Style.} ($2^{\text{nd}}$ on top row of Figure~\ref{fig:test}.) For $\ell\in\mathcal{L}$, we take
$$
Z_{1,k}^{(\ell)} \sim \Us(0,1) \text{ and } Z_{2,k}^{(\ell)} = \sin(Z_{1,k}^{(\ell)}) + 4\sin^2(V_{k}^{(\ell)}) + 0.5 \left(W_{k}^{(\ell)}\right)^4\sin(Z_{1,k}^{(\ell)}) + \epsilon_k^{(\ell)},
$$
where the $V_k^{(\ell)}$'s and $W_k^{(\ell)}$'s are generated independently from a $\Us(0,1)$ and where the $\epsilon_k^{(\ell)}$'s are generated independently from a $\Ns\left(0, \sigma^{(\ell)}\right)$ distribution with $\sigma^{(\ell)}\sim\Us(0.5,1.5)$ (resp. $\sigma^{(\ell)}\sim\Us(\pi,3\pi)$).

\item \textbf{Model 23 -- Annual Ring.} ($3^{\text{rd}}$ on top row of Figure~\ref{fig:test}.) For $\ell\in\mathcal{L}$, we take
$$
Z_{1,k}^{(\ell)} = L^{(\ell)}\cos(\theta_{k}^{(\ell)}) + V_{1,k}^{(\ell)} / 4 \text{ and } Z_{2,k}^{(\ell)} = L^{(\ell)}\sin(\theta_{k}^{(\ell)}) + V_{2,k}^{(\ell)} / 4
$$
where $L^{(\ell)}\sim\Us\{1,\ldots,R^{(\ell)}\}$ with $R^{(\ell)}\sim\Us\{2,\ldots,10\}$, and where the $V_{1,k}^{(\ell)}$'s and the $V_{2,k}^{(\ell)}$'s are generated independently from a $\Ns(0,\sigma^{(\ell)})$ with $\sigma^{(\ell)}\sim\Us(0,0.5)$ (resp. $\sigma^{(\ell)}\sim\Us(0,1)$) and where $\theta_{k}^{(\ell)}\sim\Us(0,2\pi)$.

\item \textbf{Model 24 -- Escalating variance.} ($1^{\text{st}}$ on bottom row of Figure~\ref{fig:test}.) For $\ell\in\mathcal{L}$, we take
$$
Z_{1,k}^{(\ell)} \sim\Us(0,1) \text{ and } Z_{2,k}^{(\ell)} = \left|Z_{1,k}^{(\ell)}\right|^{p^{(\ell)}}V_{k}^{(\ell)},
$$
where the $V_k^{(\ell)}$'s are generated independently from a $\Ns(0,1)$ and where $p^{(\ell)}\sim\Us(0.4,0.5)$ (resp. $p^{(\ell)}\sim\Us(0.5,0.6)$).

\end{itemize}

\ppn In addition, we used two `image-based' data generation process, referred to `infinity' and `pi', and typical samples of sizes $n=400$ drawn from which are shown in Figure \ref{fig:test}. Using our function \pkg{depstats::randimage()}, each one of these two images  (of width \(p\) and height \(q\)) is first converted to grayscale via
\[
  v : \{1,\ldots,p\}\times\{1,\ldots,q\}\;\longrightarrow\;[0,1],
  \qquad
  (i,j)\mapsto v_{ij}.
\]
We then set for each pixel
\[
  x_{ij} \;=\;\frac{i - \tfrac p2}{p},
  \qquad
  y_{ij} \;=\;-\frac{j - \tfrac q2}{q},
\]
so that \((x_{ij},y_{ij})\) is centered at \((0,0)\) and lies in \((x_{ij},y_{ij})\in\bigl[-\tfrac12+\tfrac1p,\;\tfrac12\bigr]\times\bigl[-\tfrac12,\;\tfrac12-\tfrac1q\bigr]\) which is roughly \([-\tfrac12,\tfrac12]^2\) when $p$ and $q$ are large. Next we define the sampling probabilities
\[
  p_{ij}
  = \frac{1 - v_{ij}}
         {\displaystyle\sum_{a=1}^p\sum_{b=1}^q(1 - v_{ab})},
  \qquad
  (i,j)\in\{1,\ldots,p\}\times\{1,\ldots,q\},
\]
and draw
\[
  \{(i_k,j_k)\}_{k=1}^n
  \;\overset{\mathrm{i.i.d.}}{\sim}\;\{p_{ij}\},
\]
for the sample sizes in $\{30,50,100,200,300,400\}$. 
Our returned points are 
\[
  (x_{i_k j_k},\, y_{i_k j_k}),
  \quad
  k=1,\ldots,n.
\]
Finally, let $\sigma$ be a fixed value generated from  a $\text{Unif}[0,v]$ distribution ($v$ is equal to $v_1=0.1$ or $v_2=0.2$ for the `infinity' image and $v$ is equal to $v_1=0.5$ or $v_2=1$ for the `pi' image). Then, our returned dependent samples of \(n\) points are
\[
  \Theta_k(x_{i_k j_k},\, y_{i_k j_k} + \epsilon_k),
  \quad
  k=1,\ldots,n,
\]
where, for $k=1,\ldots,n$, a vertical additive noise $\epsilon_k\overset{\text{i.i.d.}}{\sim}\mathcal{N}(0,\sigma)$ is added (using our function \pkg{depstats::normnoise()}) and $\Theta_k$ is a 2D rotation of random angle $\theta_k\overset{\text{i.i.d.}}{\sim}\text{Unif}[0,2\pi]$ applied thanks to the argument \pkg{randrotate = TRUE} of our function \pkg{depstats::depgen()}. These samples are then processed further using the function \pkg{depstats::sampleapply()} to form $n$ input features $\xx_k$ and $\tilde{\xx}_k$ (as explained in Section~\ref{subsec:scatinp}) having a structure of dependence never seen during the training stage. For each sample size $n\in\{30, 50, 100, 200, 300, 400\}$, we generated independently $1,000$ samples for $v=v_1$ and $1,000$ samples for $v=v_2$ (for each one of the two images, as described above), for a total of $4,000$ samples.

\subsection{Power analysis} \label{app:power}

\begin{landscape} 
 \begin{table}\caption{Experiment 1: the 20 dependence models (Appendix \ref{app:training}) \\ Columns \texttt{ALL-CNN-MLP}, \texttt{ALL-MLP}, \texttt{ALL-CNN} refer to the three deep-tests; other columns refer to the 19 dependence indicators listed in Table \ref{tab:indicators}. \\ Bold values highlight the highest power for a given dependence model and a given sample size.} 
\tiny \centering
\begin{tabular}{ m{1.5cm} m{0.75cm}   c c c c c c c c c c c c c c c c c c c c c c }
\hline
  & n & \texttt{ALL-CNN-MLP} & \texttt{ALL-MLP} & \texttt{ALL-CNN} & ACE & AUK & Blom & dcor & Hell & Hoeff & HSIC & Info & Ken & Martdiff & MIC & Rand & Spear & ddrV & ddrTS2 & hhgPs & hhgGs & hhgPm & hhgGm \\
\cline{3-24}
\cline{3-24}

\multirow{6}{1.5cm}{Linear} &  30 &  0.754 &  0.711 &  0.720 &  0.537 &  0.370 &  0.776 &  0.798 &  0.734 &  0.786 &  0.753 &  0.759 &  \textbf{ 0.815 } &  0.808 &  0.658 &  0.280 &  \textbf{ 0.815 } &  0.700 &  0.696 &  0.725 &  0.727 &  0.593 &  0.588 \\ 
&  50 &  0.828 &  0.799 &  0.811 &  0.678 &  0.380 &  0.822 &  0.891 &  0.847 &  0.884 &  0.835 &  0.825 &  0.894 &  \textbf{ 0.895 } &  0.730 &  0.474 &  0.890 &  0.827 &  0.842 &  0.823 &  0.820 &  0.710 &  0.696 \\ 
&  100 &  0.934 &  0.913 &  0.922 &  0.891 &  0.590 &  0.934 &  0.974 &  0.947 &  0.974 &  0.947 &  0.921 &  0.974 &  \textbf{ 0.976 } &  0.867 &  0.804 &  0.973 &  0.951 &  0.956 &  0.938 &  0.937 &  0.860 &  0.861 \\ 
&  200 &  0.976 &  0.973 &  0.957 &  0.970 &  0.836 &  0.976 &  \textbf{ 0.994 } &  0.979 &  0.993 &  0.983 &  0.961 &  0.993 &  0.993 &  0.922 &  0.920 &  0.993 &  0.992 &  0.993 &  0.980 &  0.978 &  0.922 &  0.941 \\ 
&  300 &  0.986 &  0.983 &  0.972 &  0.990 &  0.916 &  0.984 &  \textbf{ 0.998 } &  0.989 &  \textbf{ 0.998 } &  0.994 &  0.973 &  \textbf{ 0.998 } &  \textbf{ 0.998 } &  0.956 &  0.956 &  \textbf{ 0.998 } &  0.991 &  \textbf{ 0.998 } &  0.991 &  0.992 &  0.959 &  0.969 \\ 
&  400 &  0.992 &  0.994 &  0.983 &  0.996 &  0.933 &  0.996 &  \textbf{ 0.999 } &  0.995 &  \textbf{ 0.999 } &  \textbf{ 0.999 } &  0.988 &  \textbf{ 0.999 } &  \textbf{ 0.999 } &  0.975 &  0.982 &  \textbf{ 0.999 } &  \textbf{ 0.999 } &  \textbf{ 0.999 } &  0.997 &  0.996 &  0.974 &  0.983 \\ 
\hline 
\multirow{6}{1.5cm}{Diamond} &  30 &  0.358 &  0.162 &  \textbf{ 0.438 } &  0.095 &  0.045 &  0.147 &  0.042 &  0.073 &  0.035 &  0.094 &  0.086 &  0.021 &  0.041 &  0.072 &  0.007 &  0.028 &  0.163 &  0.213 &  0.146 &  0.173 &  0.159 &  0.154 \\ 
&  50 &  \textbf{ 0.645 } &  0.398 &  0.636 &  0.191 &  0.008 &  0.109 &  0.065 &  0.161 &  0.044 &  0.188 &  0.260 &  0.017 &  0.049 &  0.098 &  0.017 &  0.033 &  0.478 &  0.537 &  0.317 &  0.396 &  0.327 &  0.306 \\ 
&  100 &  \textbf{ 0.914 } &  0.826 &  0.864 &  0.550 &  0.005 &  0.077 &  0.095 &  0.448 &  0.095 &  0.449 &  0.647 &  0.013 &  0.053 &  0.097 &  0.217 &  0.029 &  0.824 &  0.864 &  0.728 &  0.780 &  0.618 &  0.691 \\ 
&  200 &  \textbf{ 0.980 } &  0.970 &  0.962 &  0.922 &  0.002 &  0.072 &  0.211 &  0.763 &  0.270 &  0.808 &  0.919 &  0.005 &  0.048 &  0.143 &  0.698 &  0.016 &  0.971 &  0.976 &  0.946 &  0.958 &  0.902 &  0.932 \\ 
&  300 &  0.998 &  0.994 &  0.987 &  0.983 &  0.000 &  0.050 &  0.420 &  0.910 &  0.555 &  0.946 &  0.981 &  0.011 &  0.054 &  0.189 &  0.916 &  0.021 &  0.997 &  \textbf{ 0.999 } &  0.988 &  0.990 &  0.960 &  0.983 \\ 
&  400 &  \textbf{ 1.000 } &  \textbf{ 1.000 } &  0.997 &  0.995 &  0.000 &  0.081 &  0.619 &  0.922 &  0.726 &  0.972 &  0.992 &  0.008 &  0.089 &  0.258 &  0.963 &  0.019 &  0.999 &  0.999 &  0.998 &  0.998 &  0.975 &  0.988 \\ 
\hline 

\multirow{6}{1.5cm}{Triangle} &  30 &  0.388 &  0.368 &  0.320 &  0.236 &  0.180 &  0.393 &  0.388 &  0.340 &  0.383 &  0.383 &  0.406 &  0.356 &  0.363 &  0.301 &  0.100 &  0.353 &  0.379 &  0.262 &  0.405 &  \textbf{ 0.417 } &  0.290 &  0.298 \\ 
&  50 &  0.545 &  0.540 &  0.518 &  0.366 &  0.158 &  0.401 &  0.548 &  0.493 &  0.541 &  0.564 &  0.557 &  0.431 &  0.475 &  0.394 &  0.204 &  0.438 &  0.585 &  0.377 &  0.578 &  \textbf{ 0.587 } &  0.437 &  0.438 \\ 
&  100 &  0.738 &  0.741 &  0.702 &  0.680 &  0.202 &  0.535 &  0.749 &  0.651 &  0.756 &  0.764 &  0.741 &  0.576 &  0.682 &  0.566 &  0.483 &  0.568 &  0.790 &  0.523 &  0.804 &  \textbf{ 0.807 } &  0.642 &  0.653 \\ 
&  200 &  0.888 &  0.889 &  0.850 &  0.887 &  0.308 &  0.646 &  0.878 &  0.803 &  0.875 &  0.894 &  0.867 &  0.676 &  0.829 &  0.723 &  0.766 &  0.674 &  0.904 &  0.688 &  0.910 &  \textbf{ 0.912 } &  0.819 &  0.837 \\ 
&  300 &  0.951 &  0.952 &  0.920 &  0.960 &  0.352 &  0.671 &  0.933 &  0.863 &  0.937 &  0.946 &  0.931 &  0.715 &  0.896 &  0.821 &  0.895 &  0.706 &  0.951 &  0.738 &  \textbf{ 0.965 } &  0.964 &  0.913 &  0.912 \\ 
&  400 &  0.955 &  0.960 &  0.924 &  0.962 &  0.386 &  0.736 &  0.951 &  0.902 &  0.951 &  0.957 &  0.942 &  0.768 &  0.935 &  0.845 &  0.929 &  0.770 &  \textbf{ 0.973 } &  0.803 &  0.969 &  0.970 &  0.924 &  0.933 \\ 
\hline 

\multirow{6}{1.5cm}{Crescent} &  30 &  0.650 &  0.624 &  0.613 &  0.470 &  0.069 &  0.482 &  0.558 &  0.530 &  0.504 &  0.653 &  \textbf{ 0.664 } &  0.337 &  0.469 &  0.578 &  0.111 &  0.386 &  0.599 &  0.292 &  0.624 &  0.626 &  0.544 &  0.554 \\ 
&  50 &  0.812 &  0.788 &  0.792 &  0.654 &  0.061 &  0.527 &  0.770 &  0.692 &  0.732 &  0.822 &  \textbf{ 0.827 } &  0.482 &  0.676 &  0.689 &  0.333 &  0.529 &  0.798 &  0.485 &  0.802 &  0.813 &  0.741 &  0.730 \\ 
&  100 &  0.899 &  0.914 &  0.880 &  0.883 &  0.077 &  0.603 &  0.897 &  0.839 &  0.890 &  0.918 &  \textbf{ 0.919 } &  0.634 &  0.860 &  0.828 &  0.759 &  0.677 &  0.912 &  0.638 &  0.916 &  0.918 &  0.878 &  0.885 \\ 
&  200 &  0.981 &  \textbf{ 0.990 } &  0.969 &  0.987 &  0.067 &  0.697 &  0.977 &  0.929 &  0.975 &  0.982 &  0.977 &  0.745 &  0.954 &  0.912 &  0.944 &  0.770 &  0.987 &  0.774 &  0.988 &  0.989 &  0.974 &  0.977 \\ 
&  300 &  0.992 &  0.993 &  0.985 &  0.996 &  0.081 &  0.733 &  0.992 &  0.956 &  0.991 &  0.994 &  0.994 &  0.799 &  0.980 &  0.953 &  0.977 &  0.822 &  0.994 &  0.829 &  \textbf{ 0.997 } &  0.995 &  0.988 &  0.991 \\ 
&  400 &  0.996 &  0.997 &  0.991 &  0.998 &  0.095 &  0.796 &  0.997 &  0.981 &  0.997 &  0.998 &  0.997 &  0.852 &  0.989 &  0.980 &  0.990 &  0.867 &  0.998 &  0.889 &  \textbf{ 0.999 } &  \textbf{ 0.999 } &  0.992 &  0.996 \\ 
\hline 

\multirow{6}{1.5cm}{Points} &  30 &  \textbf{ 0.669 } &  0.617 &  0.662 &  0.504 &  0.051 &  0.288 &  0.418 &  0.466 &  0.301 &  0.559 &  0.575 &  0.154 &  0.280 &  0.531 &  0.155 &  0.181 &  0.367 &  0.197 &  0.620 &  0.633 &  0.548 &  0.598 \\ 
&  50 &  0.823 &  0.791 &  \textbf{ 0.826 } &  0.647 &  0.019 &  0.242 &  0.569 &  0.616 &  0.486 &  0.728 &  0.780 &  0.205 &  0.448 &  0.632 &  0.465 &  0.233 &  0.504 &  0.375 &  0.780 &  0.780 &  0.737 &  0.746 \\ 
&  100 &  \textbf{ 0.934 } &  0.924 &  0.931 &  0.872 &  0.022 &  0.285 &  0.735 &  0.755 &  0.696 &  0.874 &  0.924 &  0.274 &  0.616 &  0.755 &  0.762 &  0.304 &  0.661 &  0.630 &  0.901 &  0.899 &  0.884 &  0.885 \\ 
&  200 &  \textbf{ 0.972 } &  0.967 &  0.969 &  0.950 &  0.031 &  0.327 &  0.841 &  0.875 &  0.828 &  0.942 &  0.966 &  0.310 &  0.696 &  0.814 &  0.922 &  0.327 &  0.797 &  0.782 &  0.956 &  0.959 &  0.938 &  0.945 \\ 
&  300 &  \textbf{ 0.987 } &  0.975 &  0.979 &  0.972 &  0.039 &  0.336 &  0.904 &  0.896 &  0.902 &  0.961 &  0.976 &  0.343 &  0.781 &  0.868 &  0.953 &  0.346 &  0.816 &  0.839 &  0.971 &  0.971 &  0.956 &  0.964 \\ 
&  400 &  \textbf{ 0.991 } &  0.987 &  0.987 &  0.985 &  0.034 &  0.339 &  0.922 &  0.929 &  0.922 &  0.981 &  0.990 &  0.336 &  0.783 &  0.875 &  0.970 &  0.338 &  0.880 &  0.898 &  0.981 &  0.981 &  0.970 &  0.973 \\ 
\hline 

\multirow{6}{1.5cm}{Expo-nential} &  30 &  0.684 &  0.656 &  0.627 &  0.530 &  0.313 &  0.665 &  \textbf{ 0.699 } &  0.641 &  0.682 &  0.675 &  0.687 &  0.669 &  0.678 &  0.603 &  0.282 &  0.665 &  0.653 &  0.595 &  0.694 &  0.689 &  0.562 &  0.563 \\ 
&  50 &  0.783 &  0.767 &  0.756 &  0.663 &  0.321 &  0.686 &  \textbf{ 0.797 } &  0.743 &  0.790 &  0.773 &  0.775 &  0.750 &  0.766 &  0.671 &  0.498 &  0.751 &  0.789 &  0.713 &  0.789 &  0.785 &  0.689 &  0.693 \\ 
&  100 &  0.858 &  0.849 &  0.842 &  0.835 &  0.426 &  0.766 &  0.889 &  0.849 &  \textbf{ 0.891 } &  0.869 &  0.855 &  0.828 &  0.866 &  0.785 &  0.772 &  0.821 &  0.869 &  0.812 &  0.873 &  0.869 &  0.805 &  0.802 \\ 
&  200 &  0.917 &  0.911 &  0.903 &  0.917 &  0.629 &  0.845 &  \textbf{ 0.930 } &  0.906 &  0.926 &  0.921 &  0.908 &  0.871 &  0.922 &  0.852 &  0.870 &  0.866 &  0.925 &  0.886 &  0.919 &  0.920 &  0.879 &  0.892 \\ 
&  300 &  0.947 &  0.941 &  0.920 &  0.953 &  0.705 &  0.875 &  \textbf{ 0.960 } &  0.933 &  0.958 &  0.950 &  0.930 &  0.915 &  0.952 &  0.902 &  0.910 &  0.908 &  0.955 &  0.921 &  0.958 &  0.958 &  0.912 &  0.916 \\ 
&  400 &  0.940 &  0.946 &  0.925 &  0.952 &  0.733 &  0.892 &  0.957 &  0.934 &  0.958 &  0.958 &  0.937 &  0.926 &  0.958 &  0.890 &  0.918 &  0.927 &  \textbf{ 0.960 } &  0.939 &  0.959 &  0.959 &  0.913 &  0.926 \\ 
\hline 

\multirow{6}{1.5cm}{Circles} &  30 &  \textbf{ 0.438 } &  0.314 &  0.390 &  0.289 &  0.067 &  0.174 &  0.096 &  0.242 &  0.129 &  0.211 &  0.223 &  0.080 &  0.097 &  0.145 &  0.091 &  0.087 &  0.204 &  0.270 &  0.281 &  0.295 &  0.213 &  0.228 \\ 
&  50 &  \textbf{ 0.712 } &  0.582 &  0.690 &  0.498 &  0.051 &  0.113 &  0.138 &  0.398 &  0.240 &  0.313 &  0.370 &  0.095 &  0.108 &  0.299 &  0.305 &  0.095 &  0.339 &  0.409 &  0.494 &  0.526 &  0.387 &  0.413 \\ 
&  100 &  \textbf{ 0.910 } &  0.848 &  0.868 &  0.765 &  0.057 &  0.117 &  0.314 &  0.621 &  0.436 &  0.410 &  0.728 &  0.143 &  0.157 &  0.644 &  0.748 &  0.142 &  0.425 &  0.506 &  0.782 &  0.797 &  0.711 &  0.735 \\ 
&  200 &  \textbf{ 0.984 } &  0.966 &  0.967 &  0.925 &  0.076 &  0.152 &  0.550 &  0.831 &  0.636 &  0.599 &  0.948 &  0.249 &  0.248 &  0.839 &  0.934 &  0.225 &  0.578 &  0.661 &  0.936 &  0.952 &  0.908 &  0.924 \\ 
&  300 &  \textbf{ 0.997 } &  0.991 &  0.987 &  0.952 &  0.089 &  0.133 &  0.678 &  0.900 &  0.757 &  0.715 &  0.977 &  0.293 &  0.312 &  0.906 &  0.975 &  0.274 &  0.705 &  0.770 &  0.971 &  0.979 &  0.958 &  0.969 \\ 
&  400 &  0.999 &  \textbf{ 1.000 } &  0.999 &  0.981 &  0.128 &  0.192 &  0.812 &  0.967 &  0.876 &  0.803 &  0.996 &  0.391 &  0.424 &  0.950 &  0.994 &  0.371 &  0.793 &  0.837 &  0.995 &  0.995 &  0.992 &  0.997 \\ 
\hline 

\multirow{6}{1.5cm}{Cross} &  30 &  0.556 &  0.578 &  0.552 &  0.456 &  0.040 &  0.207 &  0.182 &  0.430 &  0.234 &  0.424 &  0.506 &  0.126 &  0.104 &  0.310 &  0.186 &  0.096 &  0.298 &  0.295 &  \textbf{ 0.584 } &  0.571 &  0.430 &  0.468 \\ 
&  50 &  0.766 &  0.759 &  \textbf{ 0.781 } &  0.651 &  0.014 &  0.143 &  0.340 &  0.568 &  0.379 &  0.626 &  0.729 &  0.115 &  0.140 &  0.515 &  0.439 &  0.091 &  0.417 &  0.408 &  0.716 &  0.714 &  0.608 &  0.624 \\ 
&  100 &  0.910 &  0.901 &  \textbf{ 0.911 } &  0.881 &  0.002 &  0.164 &  0.652 &  0.761 &  0.648 &  0.826 &  0.892 &  0.152 &  0.331 &  0.741 &  0.817 &  0.104 &  0.533 &  0.554 &  0.872 &  0.870 &  0.827 &  0.827 \\ 
&  200 &  \textbf{ 0.986 } &  0.980 &  0.970 &  0.977 &  0.001 &  0.262 &  0.821 &  0.852 &  0.814 &  0.932 &  0.965 &  0.160 &  0.563 &  0.855 &  0.938 &  0.115 &  0.637 &  0.689 &  0.961 &  0.956 &  0.927 &  0.922 \\ 
&  300 &  \textbf{ 0.993 } &  0.990 &  0.990 &  0.992 &  0.000 &  0.295 &  0.917 &  0.930 &  0.907 &  0.970 &  0.989 &  0.207 &  0.673 &  0.908 &  0.984 &  0.139 &  0.688 &  0.777 &  0.986 &  0.986 &  0.968 &  0.974 \\ 
&  400 &  0.993 &  0.991 &  0.990 &  \textbf{ 0.994 } &  0.000 &  0.352 &  0.910 &  0.929 &  0.906 &  0.971 &  0.984 &  0.227 &  0.708 &  0.915 &  0.978 &  0.142 &  0.739 &  0.836 &  0.990 &  0.990 &  0.971 &  0.974 \\ 
\hline 

\multirow{6}{1.5cm}{Wedge} &  30 &  0.588 &  0.568 &  0.540 &  0.421 &  0.230 &  0.558 &  0.592 &  0.532 &  0.573 &  0.602 &  \textbf{ 0.616 } &  0.539 &  0.556 &  0.527 &  0.206 &  0.527 &  0.563 &  0.425 &  0.605 &  0.607 &  0.500 &  0.478 \\ 
&  50 &  0.732 &  0.725 &  0.715 &  0.588 &  0.230 &  0.602 &  \textbf{ 0.760 } &  0.672 &  0.752 &  0.754 &  0.753 &  0.658 &  0.703 &  0.616 &  0.379 &  0.666 &  0.753 &  0.601 &  0.750 &  0.748 &  0.651 &  0.636 \\ 
&  100 &  0.856 &  0.852 &  0.817 &  0.827 &  0.293 &  0.710 &  0.880 &  0.810 &  0.867 &  0.878 &  0.856 &  0.770 &  0.847 &  0.760 &  0.706 &  0.774 &  \textbf{ 0.882 } &  0.765 &  0.872 &  0.871 &  0.793 &  0.796 \\ 
&  200 &  0.913 &  0.915 &  0.889 &  0.919 &  0.445 &  0.772 &  0.923 &  0.880 &  0.924 &  0.928 &  0.904 &  0.843 &  0.917 &  0.831 &  0.853 &  0.850 &  \textbf{ 0.943 } &  0.847 &  0.925 &  0.923 &  0.875 &  0.873 \\ 
&  300 &  0.958 &  0.962 &  0.928 &  0.969 &  0.476 &  0.820 &  0.961 &  0.931 &  0.961 &  0.970 &  0.950 &  0.875 &  0.946 &  0.890 &  0.931 &  0.878 &  \textbf{ 0.976 } &  0.896 &  0.967 &  0.966 &  0.928 &  0.937 \\ 
&  400 &  0.967 &  0.976 &  0.937 &  0.982 &  0.517 &  0.857 &  0.970 &  0.936 &  0.969 &  0.980 &  0.960 &  0.903 &  0.958 &  0.911 &  0.947 &  0.903 &  \textbf{ 0.985 } &  0.919 &  0.978 &  0.977 &  0.951 &  0.956 \\ 
\hline 

\multirow{6}{1.5cm}{Cubic} &  30 &  \textbf{ 0.732 } &  0.698 &  0.686 &  0.581 &  0.328 &  0.710 &  0.729 &  0.699 &  0.723 &  0.719 &  0.723 &  0.725 &  0.725 &  0.659 &  0.223 &  0.724 &  0.671 &  0.656 &  0.713 &  0.712 &  0.592 &  0.595 \\ 
&  50 &  \textbf{ 0.819 } &  0.813 &  0.805 &  0.733 &  0.351 &  0.746 &  \textbf{ 0.819 } &  0.795 &  0.805 &  0.818 &  0.813 &  0.773 &  0.786 &  0.762 &  0.535 &  0.770 &  0.768 &  0.764 &  0.813 &  0.817 &  0.740 &  0.733 \\ 
&  100 &  \textbf{ 0.916 } &  0.899 &  0.897 &  0.895 &  0.397 &  0.798 &  0.909 &  0.874 &  0.902 &  0.904 &  0.909 &  0.854 &  0.890 &  0.869 &  0.808 &  0.859 &  0.853 &  0.861 &  0.904 &  0.906 &  0.863 &  0.854 \\ 
&  200 &  \textbf{ 0.972 } &  0.968 &  0.964 &  0.971 &  0.608 &  0.868 &  0.944 &  0.936 &  0.946 &  0.955 &  0.963 &  0.898 &  0.931 &  0.930 &  0.942 &  0.901 &  0.917 &  0.914 &  0.960 &  0.961 &  0.942 &  0.937 \\ 
&  300 &  0.981 &  0.975 &  0.971 &  \textbf{ 0.983 } &  0.656 &  0.899 &  0.976 &  0.958 &  0.974 &  0.975 &  0.973 &  0.912 &  0.964 &  0.945 &  0.956 &  0.913 &  0.940 &  0.945 &  0.978 &  0.979 &  0.952 &  0.954 \\ 
&  400 &  0.985 &  0.978 &  0.978 &  \textbf{ 0.990 } &  0.693 &  0.882 &  0.975 &  0.965 &  0.976 &  0.979 &  0.980 &  0.924 &  0.964 &  0.958 &  0.972 &  0.923 &  0.951 &  0.957 &  0.982 &  0.982 &  0.960 &  0.967 \\ 
\hline 

\multirow{6}{1.5cm}{W-shape} &  30 &  \textbf{ 0.437 } &  0.414 &  0.392 &  0.304 &  0.086 &  0.329 &  0.329 &  0.320 &  0.295 &  0.419 &  0.415 &  0.194 &  0.274 &  0.373 &  0.102 &  0.223 &  0.304 &  0.142 &  0.421 &  0.428 &  0.344 &  0.342 \\ 
&  50 &  \textbf{ 0.637 } &  0.619 &  0.606 &  0.496 &  0.065 &  0.331 &  0.523 &  0.479 &  0.484 &  0.588 &  0.632 &  0.315 &  0.433 &  0.527 &  0.279 &  0.357 &  0.518 &  0.315 &  0.592 &  0.599 &  0.544 &  0.535 \\ 
&  100 &  0.810 &  0.828 &  0.789 &  0.775 &  0.106 &  0.417 &  0.769 &  0.658 &  0.743 &  0.808 &  \textbf{ 0.837 } &  0.515 &  0.702 &  0.701 &  0.605 &  0.551 &  0.717 &  0.491 &  0.803 &  0.808 &  0.742 &  0.750 \\ 
&  200 &  0.948 &  \textbf{ 0.956 } &  0.922 &  0.955 &  0.108 &  0.530 &  0.931 &  0.836 &  0.920 &  0.943 &  0.953 &  0.687 &  0.870 &  0.861 &  0.884 &  0.707 &  0.916 &  0.716 &  0.940 &  0.944 &  0.917 &  0.924 \\ 
&  300 &  0.984 &  \textbf{ 0.990 } &  0.962 &  0.987 &  0.152 &  0.593 &  0.973 &  0.917 &  0.967 &  0.979 &  0.983 &  0.774 &  0.945 &  0.938 &  0.965 &  0.799 &  0.966 &  0.826 &  0.982 &  0.984 &  0.963 &  0.976 \\ 
&  400 &  \textbf{ 0.996 } &  \textbf{ 0.996 } &  0.981 &  0.995 &  0.187 &  0.665 &  0.983 &  0.951 &  0.983 &  0.987 &  0.993 &  0.810 &  0.967 &  0.963 &  0.987 &  0.827 &  0.987 &  0.862 &  0.994 &  0.993 &  0.981 &  0.994 \\ 
\hline 

\multirow{6}{1.5cm}{Parabola} &  30 &  0.718 &  0.693 &  0.660 &  0.549 &  0.252 &  0.665 &  0.729 &  0.685 &  0.712 &  0.726 &  \textbf{ 0.730 } &  0.666 &  0.693 &  0.648 &  0.270 &  0.662 &  0.709 &  0.565 &  0.711 &  0.719 &  0.587 &  0.588 \\ 
&  50 &  0.837 &  0.821 &  0.818 &  0.726 &  0.255 &  0.715 &  0.835 &  0.798 &  0.828 &  0.837 &  0.849 &  0.758 &  0.814 &  0.732 &  0.553 &  0.760 &  \textbf{ 0.850 } &  0.730 &  0.820 &  0.825 &  0.750 &  0.738 \\ 
&  100 &  0.907 &  0.912 &  0.897 &  0.895 &  0.308 &  0.770 &  \textbf{ 0.924 } &  0.881 &  0.919 &  0.923 &  0.914 &  0.824 &  0.901 &  0.833 &  0.813 &  0.832 &  0.920 &  0.819 &  0.916 &  0.917 &  0.875 &  0.875 \\ 
&  200 &  0.955 &  0.956 &  0.935 &  0.963 &  0.466 &  0.829 &  0.952 &  0.928 &  0.954 &  0.958 &  0.950 &  0.868 &  0.939 &  0.892 &  0.916 &  0.867 &  \textbf{ 0.969 } &  0.881 &  0.962 &  0.960 &  0.932 &  0.938 \\ 
&  300 &  0.971 &  0.974 &  0.953 &  0.979 &  0.524 &  0.852 &  0.969 &  0.948 &  0.970 &  0.976 &  0.967 &  0.888 &  0.959 &  0.919 &  0.962 &  0.889 &  \textbf{ 0.984 } &  0.916 &  0.972 &  0.973 &  0.952 &  0.965 \\ 
&  400 &  0.989 &  0.991 &  0.975 &  0.996 &  0.558 &  0.879 &  0.984 &  0.969 &  0.985 &  0.991 &  0.980 &  0.921 &  0.975 &  0.945 &  0.974 &  0.923 &  \textbf{ 0.998 } &  0.941 &  0.992 &  0.992 &  0.969 &  0.978 \\ 
\hline 
\end{tabular} \label{tab:pow1} 
\end{table}

\addtocounter{table}{-1}
\begin{table} \label{tab:pow2} \caption{Experiment 1: the 20 dependence models  (Appendix \ref{app:training}) \\ Columns \texttt{ALL-CNN-MLP}, \texttt{ALL-MLP}, \texttt{ALL-CNN} refer to the three deep-tests; other columns refer to the 19 dependence indicators listed in Table \ref{tab:indicators}. \\ Bold values highlight the highest power for a given dependence model and a given sample size.}
\tiny \centering
\begin{tabular}{ m{1.5cm} m{0.75cm}   c c c c c c c c c c c c c c c c c c c c c c }
\hline
  & n & \texttt{ALL-CNN-MLP} & \texttt{ALL-MLP} & \texttt{ALL-CNN} & ACE & AUK & Blom & dcor & Hell & Hoeff & HSIC & Info & Ken & Martdiff & MIC & Rand & Spear & ddrV & ddrTS2 & hhgPs & hhgGs & hhgPm & hhgGm \\
\cline{3-24}
\cline{3-24}

\multirow{6}{1.5cm}{Two-parabola} &  30 &  \textbf{ 0.536 } &  0.528 &  0.504 &  0.434 &  0.061 &  0.381 &  0.304 &  0.390 &  0.317 &  0.457 &  0.483 &  0.224 &  0.248 &  0.347 &  0.166 &  0.210 &  0.369 &  0.345 &  0.527 &  0.523 &  0.420 &  0.421 \\ 
&  50 &  0.726 &  0.701 &  \textbf{ 0.738 } &  0.615 &  0.021 &  0.385 &  0.462 &  0.543 &  0.485 &  0.639 &  0.685 &  0.314 &  0.366 &  0.507 &  0.391 &  0.303 &  0.522 &  0.528 &  0.695 &  0.692 &  0.610 &  0.623 \\ 
&  100 &  \textbf{ 0.862 } &  0.859 &  0.854 &  0.834 &  0.017 &  0.465 &  0.674 &  0.726 &  0.680 &  0.787 &  0.842 &  0.444 &  0.506 &  0.702 &  0.742 &  0.417 &  0.675 &  0.698 &  0.841 &  0.840 &  0.801 &  0.797 \\ 
&  200 &  \textbf{ 0.943 } &  0.936 &  0.928 &  0.917 &  0.017 &  0.560 &  0.797 &  0.825 &  0.792 &  0.881 &  0.913 &  0.529 &  0.624 &  0.789 &  0.885 &  0.500 &  0.811 &  0.845 &  0.927 &  0.922 &  0.890 &  0.892 \\ 
&  300 &  \textbf{ 0.979 } &  0.974 &  0.975 &  0.971 &  0.016 &  0.588 &  0.877 &  0.879 &  0.865 &  0.932 &  0.968 &  0.582 &  0.699 &  0.882 &  0.959 &  0.545 &  0.876 &  0.905 &  0.963 &  0.963 &  0.943 &  0.952 \\ 
&  400 &  \textbf{ 0.988 } &  0.985 &  0.983 &  0.982 &  0.018 &  0.638 &  0.916 &  0.910 &  0.909 &  0.945 &  0.980 &  0.614 &  0.747 &  0.918 &  0.971 &  0.569 &  0.901 &  0.929 &  0.974 &  0.973 &  0.961 &  0.965 \\ 
\hline 

\multirow{6}{1.5cm}{Sine} &  30 &  \textbf{ 0.489 } &  0.374 &  0.477 &  0.296 &  0.086 &  0.295 &  0.221 &  0.292 &  0.229 &  0.290 &  0.300 &  0.168 &  0.203 &  0.271 &  0.058 &  0.176 &  0.232 &  0.247 &  0.359 &  0.382 &  0.278 &  0.302 \\ 
&  50 &  \textbf{ 0.701 } &  0.633 &  0.640 &  0.489 &  0.050 &  0.339 &  0.408 &  0.474 &  0.401 &  0.454 &  0.502 &  0.270 &  0.353 &  0.491 &  0.227 &  0.285 &  0.412 &  0.440 &  0.577 &  0.591 &  0.478 &  0.505 \\ 
&  100 &  \textbf{ 0.883 } &  0.856 &  0.843 &  0.797 &  0.067 &  0.439 &  0.666 &  0.719 &  0.667 &  0.691 &  0.788 &  0.442 &  0.559 &  0.759 &  0.651 &  0.454 &  0.604 &  0.631 &  0.819 &  0.827 &  0.781 &  0.781 \\ 
&  200 &  \textbf{ 0.928 } &  0.917 &  0.898 &  0.915 &  0.096 &  0.551 &  0.816 &  0.835 &  0.815 &  0.816 &  0.884 &  0.594 &  0.728 &  0.874 &  0.848 &  0.593 &  0.742 &  0.766 &  0.885 &  0.889 &  0.864 &  0.880 \\ 
&  300 &  \textbf{ 0.971 } &  0.966 &  0.943 &  0.959 &  0.081 &  0.589 &  0.897 &  0.894 &  0.899 &  0.884 &  0.955 &  0.672 &  0.820 &  0.922 &  0.918 &  0.677 &  0.817 &  0.841 &  0.953 &  0.953 &  0.930 &  0.945 \\ 
&  400 &  \textbf{ 0.983 } &  0.978 &  0.970 &  0.981 &  0.107 &  0.663 &  0.939 &  0.922 &  0.936 &  0.928 &  0.972 &  0.730 &  0.889 &  0.950 &  0.962 &  0.725 &  0.871 &  0.875 &  0.975 &  0.975 &  0.954 &  0.966 \\ 
\hline 

\multirow{6}{1.5cm}{Doppler} &  30 &  \textbf{ 0.446 } &  0.376 &  0.408 &  0.240 &  0.069 &  0.310 &  0.227 &  0.263 &  0.226 &  0.308 &  0.296 &  0.154 &  0.198 &  0.277 &  0.036 &  0.173 &  0.249 &  0.187 &  0.344 &  0.369 &  0.278 &  0.299 \\ 
&  50 &  \textbf{ 0.679 } &  0.621 &  0.618 &  0.477 &  0.037 &  0.327 &  0.436 &  0.452 &  0.409 &  0.510 &  0.535 &  0.253 &  0.365 &  0.482 &  0.171 &  0.288 &  0.465 &  0.403 &  0.584 &  0.605 &  0.518 &  0.539 \\ 
&  100 &  \textbf{ 0.854 } &  \textbf{ 0.854 } &  0.828 &  0.804 &  0.048 &  0.432 &  0.689 &  0.685 &  0.675 &  0.776 &  0.819 &  0.441 &  0.568 &  0.732 &  0.634 &  0.463 &  0.719 &  0.652 &  0.835 &  0.840 &  0.785 &  0.791 \\ 
&  200 &  \textbf{ 0.949 } &  0.948 &  0.917 &  0.928 &  0.049 &  0.588 &  0.897 &  0.833 &  0.880 &  0.908 &  0.938 &  0.589 &  0.778 &  0.877 &  0.862 &  0.616 &  0.869 &  0.825 &  0.929 &  0.935 &  0.906 &  0.915 \\ 
&  300 &  \textbf{ 0.965 } &  \textbf{ 0.965 } &  0.944 &  0.962 &  0.070 &  0.639 &  0.927 &  0.897 &  0.928 &  0.935 &  0.963 &  0.669 &  0.874 &  0.924 &  0.916 &  0.697 &  0.916 &  0.885 &  0.959 &  0.960 &  0.939 &  0.948 \\ 
&  400 &  0.978 &  0.978 &  0.965 &  0.978 &  0.080 &  0.705 &  0.956 &  0.934 &  0.959 &  0.964 &  0.977 &  0.736 &  0.932 &  0.954 &  0.968 &  0.752 &  0.956 &  0.949 &  0.977 &  \textbf{ 0.979 } &  0.960 &  0.967 \\ 
\hline 

\multirow{6}{1.5cm}{Heavy-sine} &  30 &  \textbf{ 0.619 } &  0.554 &  0.579 &  0.339 &  0.294 &  0.581 &  0.596 &  0.565 &  0.581 &  0.524 &  0.540 &  0.602 &  0.610 &  0.447 &  0.095 &  0.610 &  0.507 &  0.514 &  0.533 &  0.544 &  0.323 &  0.327 \\ 
&  50 &  \textbf{ 0.769 } &  0.700 &  0.754 &  0.522 &  0.318 &  0.631 &  0.723 &  0.706 &  0.717 &  0.683 &  0.688 &  0.707 &  0.726 &  0.579 &  0.232 &  0.720 &  0.717 &  0.744 &  0.707 &  0.718 &  0.533 &  0.549 \\ 
&  100 &  \textbf{ 0.894 } &  0.859 &  0.869 &  0.808 &  0.384 &  0.781 &  0.859 &  0.847 &  0.856 &  0.856 &  0.849 &  0.830 &  0.855 &  0.784 &  0.650 &  0.836 &  0.873 &  0.877 &  0.874 &  0.880 &  0.791 &  0.798 \\ 
&  200 &  0.949 &  0.936 &  0.935 &  0.937 &  0.619 &  0.861 &  0.930 &  0.924 &  0.935 &  0.936 &  0.931 &  0.879 &  0.907 &  0.891 &  0.851 &  0.888 &  0.947 &  \textbf{ 0.956 } &  0.948 &  0.949 &  0.898 &  0.911 \\ 
&  300 &  0.952 &  0.948 &  0.933 &  \textbf{ 0.955 } &  0.708 &  0.875 &  0.945 &  0.931 &  0.948 &  0.944 &  0.944 &  0.899 &  0.925 &  0.929 &  0.911 &  0.908 &  0.947 &  \textbf{ 0.955 } &  0.951 &  0.950 &  0.931 &  0.934 \\ 
&  400 &  0.957 &  0.962 &  0.943 &  0.963 &  0.742 &  0.900 &  0.958 &  0.951 &  0.960 &  0.960 &  0.958 &  0.916 &  0.937 &  0.949 &  0.934 &  0.922 &  0.961 &  \textbf{ 0.969 } &  0.966 &  0.966 &  0.945 &  0.955 \\ 
\hline 

\multirow{6}{1.5cm}{Heart} &  30 &  0.643 &  0.573 &  \textbf{ 0.662 } &  0.493 &  0.021 &  0.211 &  0.159 &  0.363 &  0.206 &  0.490 &  0.484 &  0.073 &  0.094 &  0.367 &  0.105 &  0.059 &  0.325 &  0.208 &  0.514 &  0.539 &  0.433 &  0.433 \\ 
&  50 &  0.810 &  0.776 &  \textbf{ 0.820 } &  0.696 &  0.015 &  0.140 &  0.342 &  0.568 &  0.389 &  0.720 &  0.751 &  0.077 &  0.111 &  0.515 &  0.462 &  0.058 &  0.591 &  0.382 &  0.738 &  0.743 &  0.670 &  0.693 \\ 
&  100 &  0.906 &  0.908 &  0.896 &  0.859 &  0.002 &  0.116 &  0.679 &  0.738 &  0.696 &  0.877 &  \textbf{ 0.909 } &  0.138 &  0.252 &  0.736 &  0.777 &  0.091 &  0.836 &  0.636 &  0.884 &  0.896 &  0.863 &  0.865 \\ 
&  200 &  0.966 &  0.970 &  0.947 &  0.951 &  0.001 &  0.132 &  0.884 &  0.835 &  0.878 &  0.961 &  \textbf{ 0.971 } &  0.258 &  0.585 &  0.828 &  0.895 &  0.124 &  0.943 &  0.812 &  0.963 &  0.966 &  0.935 &  0.955 \\ 
&  300 &  0.990 &  \textbf{ 0.994 } &  0.980 &  0.987 &  0.000 &  0.124 &  0.940 &  0.890 &  0.933 &  0.990 &  \textbf{ 0.994 } &  0.367 &  0.767 &  0.876 &  0.955 &  0.200 &  0.978 &  0.893 &  \textbf{ 0.994 } &  \textbf{ 0.994 } &  0.973 &  0.983 \\ 
&  400 &  0.999 &  \textbf{ 1.000 } &  0.993 &  \textbf{ 1.000 } &  0.000 &  0.136 &  0.976 &  0.924 &  0.969 &  0.999 &  0.999 &  0.436 &  0.865 &  0.917 &  0.985 &  0.264 &  0.995 &  0.924 &  0.999 &  \textbf{ 1.000 } &  0.990 &  0.998 \\ 
\hline 

\multirow{6}{1.5cm}{Spiral} &  30 &  0.106 &  0.091 &  0.056 &  0.075 &  0.094 &  \textbf{ 0.170 } &  0.064 &  0.073 &  0.060 &  0.066 &  0.068 &  0.063 &  0.065 &  0.068 &  0.029 &  0.067 &  0.059 &  0.063 &  0.080 &  0.088 &  0.067 &  0.074 \\ 
&  50 &  \textbf{ 0.238 } &  0.170 &  0.204 &  0.157 &  0.053 &  0.102 &  0.097 &  0.106 &  0.096 &  0.099 &  0.120 &  0.083 &  0.089 &  0.077 &  0.053 &  0.089 &  0.093 &  0.095 &  0.165 &  0.162 &  0.128 &  0.141 \\ 
&  100 &  \textbf{ 0.696 } &  0.473 &  0.640 &  0.434 &  0.052 &  0.091 &  0.087 &  0.249 &  0.096 &  0.118 &  0.257 &  0.079 &  0.080 &  0.165 &  0.368 &  0.078 &  0.127 &  0.114 &  0.333 &  0.335 &  0.302 &  0.358 \\ 
&  200 &  \textbf{ 0.927 } &  0.843 &  0.888 &  0.718 &  0.031 &  0.108 &  0.166 &  0.648 &  0.189 &  0.234 &  0.648 &  0.115 &  0.128 &  0.396 &  0.760 &  0.121 &  0.247 &  0.190 &  0.699 &  0.716 &  0.731 &  0.753 \\ 
&  300 &  \textbf{ 0.967 } &  0.927 &  0.940 &  0.808 &  0.030 &  0.099 &  0.246 &  0.799 &  0.274 &  0.349 &  0.799 &  0.136 &  0.170 &  0.652 &  0.894 &  0.151 &  0.341 &  0.239 &  0.849 &  0.865 &  0.880 &  0.892 \\ 
&  400 &  \textbf{ 0.986 } &  0.970 &  0.963 &  0.887 &  0.029 &  0.115 &  0.356 &  0.858 &  0.388 &  0.478 &  0.853 &  0.182 &  0.211 &  0.811 &  0.956 &  0.194 &  0.456 &  0.319 &  0.926 &  0.938 &  0.947 &  0.958 \\ 
\hline 

\multirow{6}{1.5cm}{Taegeuk} &  30 &  0.387 &  0.266 &  \textbf{ 0.424 } &  0.185 &  0.057 &  0.347 &  0.184 &  0.136 &  0.170 &  0.247 &  0.199 &  0.090 &  0.166 &  0.190 &  0.037 &  0.116 &  0.196 &  0.227 &  0.277 &  0.303 &  0.259 &  0.261 \\ 
&  50 &  \textbf{ 0.622 } &  0.494 &  0.600 &  0.341 &  0.026 &  0.370 &  0.331 &  0.271 &  0.299 &  0.414 &  0.400 &  0.156 &  0.271 &  0.275 &  0.121 &  0.210 &  0.365 &  0.425 &  0.495 &  0.526 &  0.434 &  0.415 \\ 
&  100 &  \textbf{ 0.786 } &  0.736 &  0.741 &  0.640 &  0.021 &  0.501 &  0.530 &  0.516 &  0.542 &  0.622 &  0.687 &  0.307 &  0.461 &  0.475 &  0.476 &  0.377 &  0.624 &  0.680 &  0.711 &  0.730 &  0.678 &  0.682 \\ 
&  200 &  \textbf{ 0.877 } &  0.866 &  0.851 &  0.831 &  0.020 &  0.636 &  0.751 &  0.708 &  0.763 &  0.836 &  0.857 &  0.521 &  0.652 &  0.646 &  0.737 &  0.579 &  0.816 &  0.839 &  0.862 &  0.868 &  0.830 &  0.846 \\ 
&  300 &  \textbf{ 0.931 } &  0.929 &  0.904 &  0.901 &  0.041 &  0.703 &  0.865 &  0.794 &  0.854 &  0.902 &  0.919 &  0.647 &  0.755 &  0.743 &  0.822 &  0.696 &  0.894 &  0.920 &  0.927 &  \textbf{ 0.931 } &  0.879 &  0.899 \\ 
&  400 &  0.939 &  \textbf{ 0.942 } &  0.919 &  0.916 &  0.042 &  0.719 &  0.873 &  0.822 &  0.882 &  0.914 &  0.929 &  0.702 &  0.788 &  0.804 &  0.866 &  0.735 &  0.911 &  0.924 &  0.933 &  0.935 &  0.892 &  0.904 \\ 
\hline 

\multirow{6}{1.5cm}{Sam-taegeuk} &  30 &  0.353 &  0.192 &  \textbf{ 0.386 } &  0.143 &  0.036 &  0.167 &  0.055 &  0.084 &  0.051 &  0.127 &  0.110 &  0.030 &  0.049 &  0.104 &  0.015 &  0.038 &  0.133 &  0.167 &  0.204 &  0.218 &  0.195 &  0.208 \\ 
&  50 &  \textbf{ 0.571 } &  0.404 &  0.555 &  0.253 &  0.010 &  0.115 &  0.082 &  0.171 &  0.084 &  0.197 &  0.229 &  0.026 &  0.056 &  0.137 &  0.067 &  0.037 &  0.344 &  0.392 &  0.339 &  0.371 &  0.321 &  0.343 \\ 
&  100 &  \textbf{ 0.747 } &  0.698 &  0.719 &  0.546 &  0.011 &  0.079 &  0.140 &  0.431 &  0.153 &  0.401 &  0.604 &  0.017 &  0.055 &  0.283 &  0.402 &  0.025 &  0.588 &  0.624 &  0.639 &  0.665 &  0.605 &  0.625 \\ 
&  200 &  \textbf{ 0.814 } &  0.812 &  0.775 &  0.758 &  0.009 &  0.083 &  0.391 &  0.637 &  0.408 &  0.685 &  0.779 &  0.025 &  0.087 &  0.428 &  0.670 &  0.034 &  0.779 &  0.768 &  0.782 &  0.790 &  0.757 &  0.781 \\ 
&  300 &  \textbf{ 0.884 } &  \textbf{ 0.884 } &  0.838 &  0.836 &  0.002 &  0.059 &  0.554 &  0.707 &  0.585 &  0.775 &  0.845 &  0.018 &  0.140 &  0.586 &  0.748 &  0.025 &  0.858 &  0.850 &  0.859 &  0.867 &  0.815 &  0.835 \\ 
&  400 &  0.895 &  \textbf{ 0.898 } &  0.844 &  0.858 &  0.004 &  0.073 &  0.680 &  0.710 &  0.688 &  0.827 &  0.881 &  0.026 &  0.231 &  0.670 &  0.798 &  0.039 &  0.864 &  0.844 &  0.887 &  0.891 &  0.837 &  0.855 \\ 
\hline 
\hline 

\multirow{6}{1.5cm}{Average Power} &  30 &  \textbf{ 0.528 } &  0.468 &  0.505 &  0.359 &  0.137 &  0.393 &  0.368 &  0.393 &  0.360 &  0.436 &  0.444 &  0.304 &  0.336 &  0.374 &  0.128 &  0.310 &  0.384 &  0.328 &  0.468 &  0.478 &  0.381 &  0.389 \\ 
&  50 &  \textbf{ 0.703 } &  0.645 &  0.684 &  0.522 &  0.122 &  0.392 &  0.497 &  0.528 &  0.492 &  0.578 &  0.604 &  0.369 &  0.431 &  0.486 &  0.310 &  0.380 &  0.557 &  0.498 &  0.629 &  0.641 &  0.551 &  0.555 \\ 
&  100 &  \textbf{ 0.861 } &  0.833 &  0.836 &  0.774 &  0.154 &  0.454 &  0.656 &  0.700 &  0.659 &  0.735 &  0.795 &  0.463 &  0.561 &  0.654 &  0.650 &  0.469 &  0.719 &  0.667 &  0.812 &  0.820 &  0.755 &  0.766 \\ 
&  200 &  \textbf{ 0.941 } &  0.933 &  0.920 &  0.915 &  0.221 &  0.525 &  0.779 &  0.838 &  0.786 &  0.855 &  0.910 &  0.541 &  0.670 &  0.765 &  0.855 &  0.538 &  0.835 &  0.790 &  0.919 &  0.922 &  0.887 &  0.899 \\ 
&  300 &  \textbf{ 0.969 } &  0.965 &  0.951 &  0.955 &  0.247 &  0.546 &  0.847 &  0.896 &  0.858 &  0.904 &  0.951 &  0.586 &  0.731 &  0.835 &  0.925 &  0.580 &  0.879 &  0.847 &  0.959 &  0.961 &  0.935 &  0.945 \\ 
&  400 &  0.976 &  \textbf{ 0.976 } &  0.962 &  0.970 &  0.264 &  0.581 &  0.887 &  0.921 &  0.897 &  0.930 &  0.964 &  0.620 &  0.767 &  0.872 &  0.952 &  0.610 &  0.909 &  0.881 &  0.974 &  0.974 &  0.953 &  0.962 \\ 
\hline 
\hline 

\multirow{6}{1.5cm}{Average Gap} &  30 &  \textbf{ 0.021 } &  0.081 &  0.044 &  0.190 &  0.411 &  0.156 &  0.180 &  0.156 &  0.189 &  0.112 &  0.105 &  0.244 &  0.212 &  0.175 &  0.421 &  0.239 &  0.165 &  0.220 &  0.080 &  0.070 &  0.168 &  0.160 \\ 
&  50 &  \textbf{ 0.011 } &  0.069 &  0.030 &  0.192 &  0.592 &  0.321 &  0.217 &  0.186 &  0.221 &  0.136 &  0.110 &  0.345 &  0.282 &  0.227 &  0.403 &  0.334 &  0.157 &  0.215 &  0.085 &  0.073 &  0.163 &  0.159 \\ 
&  100 &  \textbf{ 0.012 } &  0.040 &  0.037 &  0.099 &  0.718 &  0.419 &  0.217 &  0.172 &  0.214 &  0.138 &  0.078 &  0.410 &  0.312 &  0.219 &  0.223 &  0.404 &  0.153 &  0.206 &  0.060 &  0.053 &  0.117 &  0.107 \\ 
&  200 &  \textbf{ 0.006 } &  0.014 &  0.028 &  0.033 &  0.727 &  0.423 &  0.168 &  0.109 &  0.162 &  0.093 &  0.038 &  0.407 &  0.278 &  0.182 &  0.093 &  0.409 &  0.113 &  0.157 &  0.029 &  0.025 &  0.060 &  0.049 \\ 
&  300 &  \textbf{ 0.005 } &  0.008 &  0.023 &  0.019 &  0.727 &  0.428 &  0.127 &  0.078 &  0.116 &  0.069 &  0.023 &  0.388 &  0.243 &  0.138 &  0.049 &  0.394 &  0.094 &  0.127 &  0.015 &  0.013 &  0.039 &  0.029 \\ 
&  400 &  0.005 &  \textbf{ 0.005 } &  0.019 &  0.012 &  0.717 &  0.401 &  0.095 &  0.061 &  0.085 &  0.052 &  0.017 &  0.361 &  0.214 &  0.110 &  0.029 &  0.371 &  0.073 &  0.101 &  0.008 &  0.007 &  0.029 &  0.020 \\ 
\hline 
\hline 

\multirow{6}{1.5cm}{Max Gap} &  30 &  \textbf{ 0.080 } &  0.276 &  0.114 &  0.343 &  0.641 &  0.451 &  0.503 &  0.365 &  0.456 &  0.344 &  0.352 &  0.589 &  0.568 &  0.366 &  0.557 &  0.603 &  0.337 &  0.472 &  0.292 &  0.265 &  0.296 &  0.292 \\ 
&  50 &  \textbf{ 0.067 } &  0.247 &  0.084 &  0.454 &  0.807 &  0.680 &  0.580 &  0.484 &  0.601 &  0.457 &  0.385 &  0.743 &  0.709 &  0.547 &  0.628 &  0.762 &  0.373 &  0.451 &  0.328 &  0.249 &  0.325 &  0.339 \\ 
&  100 &  \textbf{ 0.069 } &  0.223 &  0.105 &  0.364 &  0.912 &  0.837 &  0.819 &  0.466 &  0.819 &  0.578 &  0.439 &  0.901 &  0.861 &  0.817 &  0.697 &  0.885 &  0.569 &  0.582 &  0.363 &  0.361 &  0.394 &  0.338 \\ 
&  200 &  \textbf{ 0.030 } &  0.084 &  0.062 &  0.209 &  0.985 &  0.908 &  0.769 &  0.279 &  0.738 &  0.693 &  0.279 &  0.975 &  0.932 &  0.837 &  0.282 &  0.964 &  0.680 &  0.737 &  0.228 &  0.211 &  0.196 &  0.174 \\ 
&  300 &  \textbf{ 0.018 } &  0.040 &  0.048 &  0.159 &  0.999 &  0.949 &  0.721 &  0.177 &  0.693 &  0.618 &  0.168 &  0.988 &  0.945 &  0.810 &  0.136 &  0.978 &  0.626 &  0.728 &  0.118 &  0.102 &  0.087 &  0.075 \\ 
&  400 &  0.020 &  \textbf{ 0.016 } &  0.054 &  0.099 &  1.000 &  0.919 &  0.630 &  0.188 &  0.598 &  0.508 &  0.133 &  0.992 &  0.911 &  0.742 &  0.100 &  0.981 &  0.530 &  0.667 &  0.060 &  0.048 &  0.061 &  0.043 \\ 
\hline 
\end{tabular}
\end{table}

\begin{table} \caption{Experiment 2: the 4 additional dependence models and the two images (Appendix \ref{app:testing}) \\ Columns \texttt{ALL-CNN-MLP}, \texttt{ALL-MLP}, \texttt{ALL-CNN} refer to the three deep-tests; other columns refer to the 19 dependence indicators listed in Table \ref{tab:indicators}. \\ Bold values highlight the highest power for a given dependence model and a given sample size.}
\tiny \centering
\begin{tabular}{ m{1.5cm} m{0.75cm}   c c c c c c c c c c c c c c c c c c c c c c }
\hline
  & n & \texttt{ALL-CNN-MLP} & \texttt{ALL-MLP} & \texttt{ALL-CNN} & ACE & AUK & Blom & dcor & Hell & Hoeff & HSIC & Info & Ken & Martdiff & MIC & Rand & Spear & ddrV & ddrTS2 & hhgPs & hhgGs & hhgPm & hhgGm \\
\cline{3-24}
\cline{3-24}

\multirow{6}{1.5cm}{Laplace A} &  30 &  0.458 &  0.439 &  0.382 &  0.304 &  0.621 &  0.545 &  0.559 &  0.401 &  0.563 &  0.508 &  0.563 &  \textbf{ 0.657 } &  0.555 &  0.317 &  0.231 &  0.604 &  0.601 &  0.627 &  0.606 &  0.571 &  0.304 &  0.312 \\ 
&  50 &  0.709 &  0.687 &  0.688 &  0.474 &  0.757 &  0.660 &  0.814 &  0.611 &  0.820 &  0.756 &  0.775 &  0.851 &  0.811 &  0.390 &  0.401 &  0.832 &  0.872 &  \textbf{ 0.891 } &  0.841 &  0.817 &  0.555 &  0.513 \\ 
&  100 &  0.942 &  0.921 &  0.935 &  0.898 &  0.962 &  0.880 &  0.978 &  0.820 &  0.977 &  0.972 &  0.967 &  0.985 &  0.974 &  0.612 &  0.813 &  0.980 &  0.984 &  \textbf{ 0.992 } &  0.986 &  0.983 &  0.897 &  0.872 \\ 
&  200 &  \textbf{ 1.000 } &  0.998 &  0.998 &  0.997 &  \textbf{ 1.000 } &  0.980 &  \textbf{ 1.000 } &  0.980 &  \textbf{ 1.000 } &  \textbf{ 1.000 } &  \textbf{ 1.000 } &  \textbf{ 1.000 } &  \textbf{ 1.000 } &  0.870 &  0.987 &  \textbf{ 1.000 } &  \textbf{ 1.000 } &  \textbf{ 1.000 } &  \textbf{ 1.000 } &  \textbf{ 1.000 } &  0.994 &  0.991 \\ 
&  300 &  \textbf{ 1.000 } &  \textbf{ 1.000 } &  \textbf{ 1.000 } &  \textbf{ 1.000 } &  0.999 &  0.999 &  \textbf{ 1.000 } &  0.998 &  \textbf{ 1.000 } &  \textbf{ 1.000 } &  \textbf{ 1.000 } &  \textbf{ 1.000 } &  \textbf{ 1.000 } &  0.963 &  \textbf{ 1.000 } &  \textbf{ 1.000 } &  \textbf{ 1.000 } &  \textbf{ 1.000 } &  \textbf{ 1.000 } &  \textbf{ 1.000 } &  \textbf{ 1.000 } &  \textbf{ 1.000 } \\ 
&  400 &  \textbf{ 1.000 } &  \textbf{ 1.000 } &  \textbf{ 1.000 } &  \textbf{ 1.000 } &  \textbf{ 1.000 } &  \textbf{ 1.000 } &  \textbf{ 1.000 } &  \textbf{ 1.000 } &  \textbf{ 1.000 } &  \textbf{ 1.000 } &  \textbf{ 1.000 } &  \textbf{ 1.000 } &  \textbf{ 1.000 } &  0.983 &  \textbf{ 1.000 } &  \textbf{ 1.000 } &  \textbf{ 1.000 } &  \textbf{ 1.000 } &  \textbf{ 1.000 } &  \textbf{ 1.000 } &  \textbf{ 1.000 } &  \textbf{ 1.000 } \\ 
\hline 

\multirow{6}{1.5cm}{Laplace B} &  30 &  0.677 &  0.622 &  0.609 &  0.445 &  0.791 &  0.705 &  0.781 &  0.562 &  0.773 &  0.705 &  0.743 &  \textbf{ 0.823 } &  0.771 &  0.475 &  0.264 &  0.795 &  0.761 &  0.793 &  0.744 &  0.724 &  0.492 &  0.483 \\ 
&  50 &  0.881 &  0.851 &  0.871 &  0.675 &  0.893 &  0.859 &  0.946 &  0.787 &  0.951 &  0.919 &  0.925 &  0.958 &  0.938 &  0.603 &  0.535 &  0.948 &  0.952 &  \textbf{ 0.968 } &  0.938 &  0.926 &  0.748 &  0.709 \\ 
&  100 &  0.987 &  0.988 &  0.987 &  0.978 &  0.993 &  0.967 &  0.998 &  0.947 &  0.998 &  0.997 &  0.995 &  0.998 &  0.998 &  0.842 &  0.938 &  0.998 &  \textbf{ 1.000 } &  0.999 &  \textbf{ 1.000 } &  0.999 &  0.964 &  0.957 \\ 
&  200 &  \textbf{ 1.000 } &  0.999 &  \textbf{ 1.000 } &  \textbf{ 1.000 } &  \textbf{ 1.000 } &  \textbf{ 1.000 } &  \textbf{ 1.000 } &  0.999 &  \textbf{ 1.000 } &  \textbf{ 1.000 } &  \textbf{ 1.000 } &  \textbf{ 1.000 } &  \textbf{ 1.000 } &  0.980 &  0.999 &  \textbf{ 1.000 } &  \textbf{ 1.000 } &  \textbf{ 1.000 } &  \textbf{ 1.000 } &  \textbf{ 1.000 } &  \textbf{ 1.000 } &  \textbf{ 1.000 } \\ 
&  300 &  \textbf{ 1.000 } &  \textbf{ 1.000 } &  \textbf{ 1.000 } &  \textbf{ 1.000 } &  \textbf{ 1.000 } &  \textbf{ 1.000 } &  \textbf{ 1.000 } &  \textbf{ 1.000 } &  \textbf{ 1.000 } &  \textbf{ 1.000 } &  \textbf{ 1.000 } &  \textbf{ 1.000 } &  \textbf{ 1.000 } &  \textbf{ 1.000 } &  \textbf{ 1.000 } &  \textbf{ 1.000 } &  \textbf{ 1.000 } &  \textbf{ 1.000 } &  \textbf{ 1.000 } &  \textbf{ 1.000 } &  \textbf{ 1.000 } &  \textbf{ 1.000 } \\ 
&  400 &  \textbf{ 1.000 } &  \textbf{ 1.000 } &  \textbf{ 1.000 } &  \textbf{ 1.000 } &  \textbf{ 1.000 } &  \textbf{ 1.000 } &  \textbf{ 1.000 } &  \textbf{ 1.000 } &  \textbf{ 1.000 } &  \textbf{ 1.000 } &  \textbf{ 1.000 } &  \textbf{ 1.000 } &  \textbf{ 1.000 } &  \textbf{ 1.000 } &  \textbf{ 1.000 } &  \textbf{ 1.000 } &  \textbf{ 1.000 } &  \textbf{ 1.000 } &  \textbf{ 1.000 } &  \textbf{ 1.000 } &  \textbf{ 1.000 } &  \textbf{ 1.000 } \\ 
\hline 

\multirow{6}{1.5cm}{Ishigami A} &  30 &  0.374 &  0.297 &  0.304 &  0.208 &  \textbf{ 0.406 } &  0.257 &  0.301 &  0.356 &  0.324 &  0.229 &  0.285 &  0.384 &  0.322 &  0.244 &  0.116 &  0.344 &  0.244 &  0.217 &  0.288 &  0.289 &  0.159 &  0.181 \\ 
&  50 &  \textbf{ 0.701 } &  0.568 &  0.669 &  0.377 &  0.470 &  0.251 &  0.502 &  0.620 &  0.536 &  0.380 &  0.516 &  0.564 &  0.518 &  0.405 &  0.217 &  0.533 &  0.407 &  0.416 &  0.532 &  0.541 &  0.282 &  0.328 \\ 
&  100 &  \textbf{ 0.965 } &  0.943 &  0.953 &  0.889 &  0.748 &  0.339 &  0.806 &  0.922 &  0.844 &  0.708 &  0.866 &  0.822 &  0.809 &  0.800 &  0.674 &  0.798 &  0.703 &  0.695 &  0.875 &  0.885 &  0.685 &  0.774 \\ 
&  200 &  \textbf{ 1.000 } &  \textbf{ 1.000 } &  \textbf{ 1.000 } &  \textbf{ 1.000 } &  0.918 &  0.537 &  0.981 &  \textbf{ 1.000 } &  0.990 &  0.965 &  0.996 &  0.971 &  0.974 &  0.987 &  0.998 &  0.964 &  0.956 &  0.948 &  0.998 &  0.998 &  0.986 &  0.998 \\ 
&  300 &  \textbf{ 1.000 } &  \textbf{ 1.000 } &  \textbf{ 1.000 } &  \textbf{ 1.000 } &  0.972 &  0.680 &  0.999 &  \textbf{ 1.000 } &  \textbf{ 1.000 } &  0.994 &  \textbf{ 1.000 } &  0.996 &  0.997 &  0.999 &  0.999 &  0.992 &  0.990 &  0.988 &  \textbf{ 1.000 } &  \textbf{ 1.000 } &  0.999 &  \textbf{ 1.000 } \\ 
&  400 &  \textbf{ 1.000 } &  \textbf{ 1.000 } &  \textbf{ 1.000 } &  \textbf{ 1.000 } &  0.988 &  0.772 &  \textbf{ 1.000 } &  \textbf{ 1.000 } &  \textbf{ 1.000 } &  \textbf{ 1.000 } &  \textbf{ 1.000 } &  \textbf{ 1.000 } &  \textbf{ 1.000 } &  \textbf{ 1.000 } &  \textbf{ 1.000 } &  0.999 &  0.997 &  0.998 &  \textbf{ 1.000 } &  \textbf{ 1.000 } &  \textbf{ 1.000 } &  \textbf{ 1.000 } \\ 
\hline 

\multirow{6}{1.5cm}{Ishigami B} &  30 &  0.705 &  \textbf{ 0.723 } &  0.668 &  0.520 &  0.005 &  0.523 &  0.552 &  0.624 &  0.552 &  0.660 &  0.688 &  0.436 &  0.499 &  0.586 &  0.083 &  0.448 &  0.553 &  0.330 &  0.626 &  0.629 &  0.494 &  0.528 \\ 
&  50 &  0.928 &  \textbf{ 0.954 } &  0.917 &  0.881 &  0.000 &  0.555 &  0.786 &  0.871 &  0.806 &  0.876 &  0.949 &  0.560 &  0.653 &  0.748 &  0.399 &  0.567 &  0.818 &  0.526 &  0.880 &  0.882 &  0.833 &  0.855 \\ 
&  100 &  0.997 &  0.999 &  0.989 &  \textbf{ 1.000 } &  0.000 &  0.647 &  0.970 &  0.993 &  0.967 &  0.987 &  0.999 &  0.630 &  0.923 &  0.904 &  0.996 &  0.632 &  0.977 &  0.648 &  0.996 &  0.995 &  0.994 &  0.998 \\ 
&  200 &  \textbf{ 1.000 } &  \textbf{ 1.000 } &  \textbf{ 1.000 } &  \textbf{ 1.000 } &  0.014 &  0.724 &  0.996 &  \textbf{ 1.000 } &  0.996 &  \textbf{ 1.000 } &  \textbf{ 1.000 } &  0.700 &  0.983 &  0.952 &  \textbf{ 1.000 } &  0.710 &  0.999 &  0.814 &  \textbf{ 1.000 } &  \textbf{ 1.000 } &  \textbf{ 1.000 } &  \textbf{ 1.000 } \\ 
&  300 &  \textbf{ 1.000 } &  \textbf{ 1.000 } &  \textbf{ 1.000 } &  \textbf{ 1.000 } &  0.054 &  0.774 &  \textbf{ 1.000 } &  \textbf{ 1.000 } &  \textbf{ 1.000 } &  \textbf{ 1.000 } &  \textbf{ 1.000 } &  0.755 &  0.999 &  0.982 &  \textbf{ 1.000 } &  0.760 &  \textbf{ 1.000 } &  0.903 &  \textbf{ 1.000 } &  \textbf{ 1.000 } &  \textbf{ 1.000 } &  \textbf{ 1.000 } \\ 
&  400 &  \textbf{ 1.000 } &  \textbf{ 1.000 } &  \textbf{ 1.000 } &  \textbf{ 1.000 } &  0.067 &  0.779 &  \textbf{ 1.000 } &  \textbf{ 1.000 } &  \textbf{ 1.000 } &  \textbf{ 1.000 } &  \textbf{ 1.000 } &  0.763 &  \textbf{ 1.000 } &  0.991 &  \textbf{ 1.000 } &  0.766 &  \textbf{ 1.000 } &  0.928 &  \textbf{ 1.000 } &  \textbf{ 1.000 } &  \textbf{ 1.000 } &  \textbf{ 1.000 } \\ 
\hline 

\multirow{6}{1.5cm}{Tree ring A} &  30 &  0.105 &  0.069 &  0.078 &  0.069 &  0.072 &  \textbf{ 0.132 } &  0.051 &  0.046 &  0.047 &  0.053 &  0.046 &  0.047 &  0.055 &  0.047 &  0.015 &  0.046 &  0.049 &  0.060 &  0.070 &  0.075 &  0.069 &  0.073 \\ 
&  50 &  \textbf{ 0.293 } &  0.202 &  0.254 &  0.215 &  0.036 &  0.113 &  0.069 &  0.095 &  0.069 &  0.088 &  0.096 &  0.049 &  0.058 &  0.080 &  0.107 &  0.053 &  0.095 &  0.127 &  0.147 &  0.159 &  0.127 &  0.143 \\ 
&  100 &  \textbf{ 0.509 } &  0.314 &  0.418 &  0.279 &  0.040 &  0.077 &  0.059 &  0.186 &  0.092 &  0.111 &  0.158 &  0.043 &  0.051 &  0.148 &  0.215 &  0.046 &  0.125 &  0.149 &  0.269 &  0.276 &  0.252 &  0.281 \\ 
&  200 &  \textbf{ 0.830 } &  0.651 &  0.737 &  0.454 &  0.013 &  0.070 &  0.075 &  0.385 &  0.149 &  0.179 &  0.423 &  0.040 &  0.052 &  0.273 &  0.393 &  0.048 &  0.215 &  0.236 &  0.474 &  0.505 &  0.571 &  0.584 \\ 
&  300 &  \textbf{ 0.936 } &  0.820 &  0.863 &  0.528 &  0.005 &  0.058 &  0.095 &  0.504 &  0.192 &  0.231 &  0.598 &  0.032 &  0.039 &  0.350 &  0.533 &  0.041 &  0.250 &  0.274 &  0.623 &  0.654 &  0.760 &  0.787 \\ 
&  400 &  \textbf{ 0.990 } &  0.948 &  0.961 &  0.633 &  0.002 &  0.046 &  0.144 &  0.607 &  0.196 &  0.227 &  0.762 &  0.029 &  0.041 &  0.400 &  0.683 &  0.035 &  0.246 &  0.253 &  0.777 &  0.832 &  0.910 &  0.918 \\ 
\hline 

\multirow{6}{1.5cm}{Tree ring B} &  30 &  0.113 &  0.064 &  0.104 &  0.075 &  0.077 &  \textbf{ 0.142 } &  0.041 &  0.040 &  0.051 &  0.046 &  0.045 &  0.040 &  0.041 &  0.051 &  0.014 &  0.040 &  0.044 &  0.057 &  0.059 &  0.068 &  0.078 &  0.070 \\ 
&  50 &  \textbf{ 0.214 } &  0.129 &  0.187 &  0.150 &  0.040 &  0.085 &  0.055 &  0.074 &  0.059 &  0.072 &  0.067 &  0.042 &  0.051 &  0.066 &  0.058 &  0.046 &  0.077 &  0.095 &  0.109 &  0.123 &  0.109 &  0.114 \\ 
&  100 &  \textbf{ 0.440 } &  0.264 &  0.362 &  0.248 &  0.035 &  0.072 &  0.072 &  0.126 &  0.079 &  0.106 &  0.154 &  0.047 &  0.059 &  0.099 &  0.183 &  0.051 &  0.131 &  0.153 &  0.192 &  0.205 &  0.199 &  0.209 \\ 
&  200 &  \textbf{ 0.740 } &  0.486 &  0.647 &  0.361 &  0.020 &  0.057 &  0.067 &  0.229 &  0.105 &  0.137 &  0.369 &  0.043 &  0.052 &  0.153 &  0.302 &  0.049 &  0.185 &  0.204 &  0.330 &  0.376 &  0.411 &  0.410 \\ 
&  300 &  \textbf{ 0.914 } &  0.701 &  0.826 &  0.487 &  0.011 &  0.048 &  0.090 &  0.367 &  0.144 &  0.196 &  0.539 &  0.047 &  0.050 &  0.222 &  0.466 &  0.048 &  0.214 &  0.246 &  0.514 &  0.561 &  0.588 &  0.639 \\ 
&  400 &  \textbf{ 0.972 } &  0.849 &  0.944 &  0.589 &  0.005 &  0.070 &  0.123 &  0.451 &  0.165 &  0.241 &  0.753 &  0.038 &  0.047 &  0.295 &  0.664 &  0.041 &  0.260 &  0.272 &  0.687 &  0.752 &  0.750 &  0.818 \\ 
\hline 

\multirow{6}{1.5cm}{Variance A} &  30 &  0.275 &  0.229 &  0.206 &  0.133 &  0.038 &  0.162 &  0.097 &  0.148 &  0.099 &  0.202 &  0.230 &  0.043 &  0.069 &  0.177 &  0.065 &  0.049 &  0.299 &  0.057 &  0.337 &  \textbf{ 0.355 } &  0.220 &  0.223 \\ 
&  50 &  0.480 &  0.430 &  0.388 &  0.201 &  0.006 &  0.102 &  0.158 &  0.236 &  0.177 &  0.353 &  0.426 &  0.037 &  0.072 &  0.212 &  0.083 &  0.036 &  0.563 &  0.068 &  0.568 &  \textbf{ 0.589 } &  0.377 &  0.372 \\ 
&  100 &  0.853 &  0.811 &  0.791 &  0.687 &  0.000 &  0.086 &  0.448 &  0.577 &  0.500 &  0.742 &  0.801 &  0.046 &  0.205 &  0.390 &  0.336 &  0.055 &  0.924 &  0.095 &  0.928 &  \textbf{ 0.930 } &  0.707 &  0.756 \\ 
&  200 &  0.991 &  0.995 &  0.986 &  0.992 &  0.000 &  0.069 &  0.894 &  0.922 &  0.913 &  0.976 &  0.986 &  0.051 &  0.666 &  0.608 &  0.883 &  0.052 &  \textbf{ 0.997 } &  0.225 &  0.996 &  0.996 &  0.963 &  0.984 \\ 
&  300 &  \textbf{ 1.000 } &  \textbf{ 1.000 } &  \textbf{ 1.000 } &  \textbf{ 1.000 } &  0.000 &  0.075 &  0.995 &  0.993 &  0.998 &  \textbf{ 1.000 } &  \textbf{ 1.000 } &  0.044 &  0.941 &  0.817 &  0.997 &  0.053 &  \textbf{ 1.000 } &  0.297 &  \textbf{ 1.000 } &  \textbf{ 1.000 } &  \textbf{ 1.000 } &  \textbf{ 1.000 } \\ 
&  400 &  \textbf{ 1.000 } &  \textbf{ 1.000 } &  \textbf{ 1.000 } &  \textbf{ 1.000 } &  0.000 &  0.083 &  \textbf{ 1.000 } &  \textbf{ 1.000 } &  \textbf{ 1.000 } &  \textbf{ 1.000 } &  \textbf{ 1.000 } &  0.039 &  0.994 &  0.930 &  \textbf{ 1.000 } &  0.045 &  \textbf{ 1.000 } &  0.376 &  \textbf{ 1.000 } &  \textbf{ 1.000 } &  \textbf{ 1.000 } &  \textbf{ 1.000 } \\ 
\hline 

\multirow{6}{1.5cm}{Variance B} &  30 &  0.390 &  0.330 &  0.285 &  0.207 &  0.039 &  0.180 &  0.128 &  0.191 &  0.157 &  0.291 &  0.330 &  0.048 &  0.089 &  0.250 &  0.069 &  0.057 &  0.379 &  0.054 &  0.438 &  \textbf{ 0.456 } &  0.326 &  0.320 \\ 
&  50 &  0.660 &  0.591 &  0.555 &  0.320 &  0.003 &  0.116 &  0.266 &  0.343 &  0.292 &  0.507 &  0.584 &  0.050 &  0.114 &  0.301 &  0.097 &  0.055 &  0.715 &  0.087 &  0.725 &  \textbf{ 0.754 } &  0.517 &  0.524 \\ 
&  100 &  0.966 &  0.944 &  0.937 &  0.846 &  0.001 &  0.069 &  0.655 &  0.761 &  0.708 &  0.884 &  0.947 &  0.035 &  0.349 &  0.515 &  0.524 &  0.041 &  0.977 &  0.112 &  0.985 &  \textbf{ 0.988 } &  0.875 &  0.903 \\ 
&  200 &  \textbf{ 1.000 } &  \textbf{ 1.000 } &  \textbf{ 1.000 } &  \textbf{ 1.000 } &  0.000 &  0.083 &  0.986 &  0.989 &  0.989 &  0.995 &  \textbf{ 1.000 } &  0.036 &  0.906 &  0.799 &  0.987 &  0.042 &  \textbf{ 1.000 } &  0.203 &  \textbf{ 1.000 } &  \textbf{ 1.000 } &  0.996 &  0.999 \\ 
&  300 &  \textbf{ 1.000 } &  \textbf{ 1.000 } &  \textbf{ 1.000 } &  \textbf{ 1.000 } &  0.000 &  0.067 &  \textbf{ 1.000 } &  0.999 &  \textbf{ 1.000 } &  \textbf{ 1.000 } &  \textbf{ 1.000 } &  0.031 &  0.998 &  0.952 &  \textbf{ 1.000 } &  0.037 &  \textbf{ 1.000 } &  0.309 &  \textbf{ 1.000 } &  \textbf{ 1.000 } &  \textbf{ 1.000 } &  \textbf{ 1.000 } \\ 
&  400 &  \textbf{ 1.000 } &  \textbf{ 1.000 } &  \textbf{ 1.000 } &  \textbf{ 1.000 } &  0.000 &  0.078 &  \textbf{ 1.000 } &  \textbf{ 1.000 } &  \textbf{ 1.000 } &  \textbf{ 1.000 } &  \textbf{ 1.000 } &  0.053 &  \textbf{ 1.000 } &  0.990 &  \textbf{ 1.000 } &  0.068 &  \textbf{ 1.000 } &  0.416 &  \textbf{ 1.000 } &  \textbf{ 1.000 } &  \textbf{ 1.000 } &  \textbf{ 1.000 } \\ 
\hline 
\end{tabular}  \label{tab:pow3} 
\end{table}

\addtocounter{table}{-1}
\begin{table}\caption{Experiment 2: the 4 additional dependence models and the two images (Appendix \ref{app:testing}) \\ Columns \texttt{ALL-CNN-MLP}, \texttt{ALL-MLP}, \texttt{ALL-CNN} refer to the three deep-tests; other columns refer to the 19 dependence indicators listed in Table \ref{tab:indicators}. \\ Bold values highlight the highest power for a given dependence model and a given sample size.}
\tiny \centering
\begin{tabular}{ m{1.5cm} m{0.75cm}   c c c c c c c c c c c c c c c c c c c c c c }
\hline
  & n & \texttt{ALL-CNN-MLP} & \texttt{ALL-MLP} & \texttt{ALL-CNN} & ACE & AUK & Blom & dcor & Hell & Hoeff & HSIC & Info & Ken & Martdiff & MIC & Rand & Spear & ddrV & ddrTS2 & hhgPs & hhgGs & hhgPm & hhgGm \\
\cline{3-24}
\cline{3-24}

\multirow{6}{1.5cm}{Infinity A} &  30 &  0.669 &  0.498 &  \textbf{ 0.690 } &  0.433 &  0.031 &  0.198 &  0.177 &  0.259 &  0.177 &  0.286 &  0.271 &  0.071 &  0.155 &  0.256 &  0.032 &  0.107 &  0.246 &  0.308 &  0.443 &  0.467 &  0.319 &  0.384 \\ 
&  50 &  0.916 &  0.850 &  \textbf{ 0.939 } &  0.751 &  0.006 &  0.166 &  0.323 &  0.542 &  0.331 &  0.508 &  0.691 &  0.103 &  0.254 &  0.488 &  0.372 &  0.168 &  0.543 &  0.639 &  0.708 &  0.724 &  0.675 &  0.722 \\ 
&  100 &  0.996 &  0.990 &  \textbf{ 0.998 } &  0.982 &  0.000 &  0.192 &  0.636 &  0.887 &  0.654 &  0.778 &  0.989 &  0.222 &  0.476 &  0.750 &  0.971 &  0.317 &  0.741 &  0.839 &  0.909 &  0.919 &  0.966 &  0.956 \\ 
&  200 &  \textbf{ 1.000 } &  \textbf{ 1.000 } &  \textbf{ 1.000 } &  0.998 &  0.000 &  0.249 &  0.835 &  0.986 &  0.839 &  0.936 &  \textbf{ 1.000 } &  0.456 &  0.698 &  0.920 &  \textbf{ 1.000 } &  0.512 &  0.880 &  0.925 &  0.993 &  0.994 &  \textbf{ 1.000 } &  0.998 \\ 
&  300 &  \textbf{ 1.000 } &  \textbf{ 1.000 } &  \textbf{ 1.000 } &  \textbf{ 1.000 } &  0.000 &  0.331 &  0.910 &  0.998 &  0.926 &  0.987 &  \textbf{ 1.000 } &  0.594 &  0.781 &  0.971 &  \textbf{ 1.000 } &  0.620 &  0.939 &  0.978 &  0.999 &  0.999 &  \textbf{ 1.000 } &  \textbf{ 1.000 } \\ 
&  400 &  \textbf{ 1.000 } &  \textbf{ 1.000 } &  \textbf{ 1.000 } &  \textbf{ 1.000 } &  0.000 &  0.426 &  0.943 &  \textbf{ 1.000 } &  0.963 &  0.995 &  \textbf{ 1.000 } &  0.655 &  0.819 &  0.988 &  \textbf{ 1.000 } &  0.672 &  0.974 &  0.993 &  \textbf{ 1.000 } &  \textbf{ 1.000 } &  \textbf{ 1.000 } &  \textbf{ 1.000 } \\ 
\hline 

\multirow{6}{1.5cm}{Infinity B} &  30 &  0.540 &  0.360 &  \textbf{ 0.586 } &  0.300 &  0.038 &  0.197 &  0.114 &  0.202 &  0.122 &  0.216 &  0.202 &  0.048 &  0.106 &  0.182 &  0.022 &  0.072 &  0.214 &  0.272 &  0.316 &  0.343 &  0.267 &  0.304 \\ 
&  50 &  \textbf{ 0.805 } &  0.661 &  \textbf{ 0.805 } &  0.537 &  0.011 &  0.148 &  0.245 &  0.401 &  0.251 &  0.402 &  0.532 &  0.067 &  0.174 &  0.336 &  0.214 &  0.122 &  0.450 &  0.514 &  0.585 &  0.607 &  0.532 &  0.552 \\ 
&  100 &  0.951 &  0.910 &  \textbf{ 0.952 } &  0.856 &  0.001 &  0.185 &  0.475 &  0.715 &  0.510 &  0.669 &  0.907 &  0.160 &  0.325 &  0.560 &  0.774 &  0.226 &  0.615 &  0.700 &  0.820 &  0.826 &  0.831 &  0.814 \\ 
&  200 &  0.991 &  0.981 &  0.988 &  0.971 &  0.001 &  0.264 &  0.725 &  0.913 &  0.756 &  0.847 &  \textbf{ 0.997 } &  0.289 &  0.534 &  0.732 &  0.979 &  0.372 &  0.801 &  0.839 &  0.935 &  0.944 &  0.952 &  0.953 \\ 
&  300 &  \textbf{ 0.998 } &  0.995 &  \textbf{ 0.998 } &  0.989 &  0.002 &  0.334 &  0.831 &  0.958 &  0.873 &  0.927 &  \textbf{ 0.998 } &  0.379 &  0.665 &  0.856 &  0.994 &  0.453 &  0.864 &  0.897 &  0.979 &  0.981 &  0.983 &  0.988 \\ 
&  400 &  0.999 &  0.999 &  0.999 &  0.992 &  0.001 &  0.415 &  0.880 &  0.970 &  0.900 &  0.933 &  \textbf{ 1.000 } &  0.437 &  0.707 &  0.892 &  \textbf{ 1.000 } &  0.495 &  0.901 &  0.913 &  0.992 &  0.993 &  0.987 &  0.993 \\ 
\hline 

\multirow{6}{1.5cm}{Pi A} &  30 &  \textbf{ 0.619 } &  0.530 &  0.567 &  0.351 &  0.267 &  0.407 &  0.476 &  0.513 &  0.535 &  0.448 &  0.513 &  0.474 &  0.463 &  0.404 &  0.087 &  0.443 &  0.387 &  0.419 &  0.523 &  0.519 &  0.304 &  0.319 \\ 
&  50 &  \textbf{ 0.810 } &  0.778 &  0.802 &  0.615 &  0.244 &  0.439 &  0.657 &  0.684 &  0.697 &  0.663 &  0.786 &  0.603 &  0.622 &  0.625 &  0.348 &  0.590 &  0.609 &  0.609 &  0.724 &  0.722 &  0.517 &  0.542 \\ 
&  100 &  0.918 &  0.929 &  0.924 &  0.923 &  0.386 &  0.506 &  0.826 &  0.823 &  0.856 &  0.854 &  \textbf{ 0.935 } &  0.744 &  0.770 &  0.872 &  0.858 &  0.722 &  0.742 &  0.750 &  0.901 &  0.898 &  0.824 &  0.841 \\ 
&  200 &  0.977 &  0.975 &  0.975 &  \textbf{ 0.982 } &  0.539 &  0.581 &  0.936 &  0.922 &  0.941 &  0.950 &  0.979 &  0.833 &  0.883 &  0.958 &  0.963 &  0.805 &  0.881 &  0.876 &  0.968 &  0.967 &  0.954 &  0.969 \\ 
&  300 &  0.994 &  0.992 &  0.988 &  \textbf{ 0.996 } &  0.551 &  0.627 &  0.962 &  0.945 &  0.968 &  0.976 &  0.993 &  0.852 &  0.909 &  0.982 &  0.988 &  0.833 &  0.907 &  0.904 &  0.984 &  0.984 &  0.979 &  0.989 \\ 
&  400 &  0.997 &  0.995 &  0.992 &  \textbf{ 0.998 } &  0.581 &  0.648 &  0.982 &  0.974 &  0.988 &  0.987 &  0.997 &  0.891 &  0.947 &  0.997 &  0.997 &  0.872 &  0.957 &  0.951 &  0.993 &  0.992 &  0.994 &  0.994 \\ 
\hline 

\multirow{6}{1.5cm}{Pi B} &  30 &  \textbf{ 0.734 } &  0.669 &  0.686 &  0.529 &  0.340 &  0.564 &  0.650 &  0.683 &  0.679 &  0.601 &  0.651 &  0.669 &  0.647 &  0.576 &  0.231 &  0.648 &  0.586 &  0.598 &  0.646 &  0.640 &  0.415 &  0.456 \\ 
&  50 &  \textbf{ 0.845 } &  0.816 &  0.833 &  0.726 &  0.347 &  0.599 &  0.771 &  0.792 &  0.804 &  0.750 &  0.832 &  0.754 &  0.759 &  0.719 &  0.516 &  0.745 &  0.725 &  0.732 &  0.797 &  0.795 &  0.631 &  0.654 \\ 
&  100 &  0.938 &  \textbf{ 0.942 } &  0.926 &  0.930 &  0.584 &  0.641 &  0.874 &  0.870 &  0.891 &  0.883 &  0.941 &  0.818 &  0.843 &  0.888 &  0.885 &  0.805 &  0.809 &  0.813 &  0.917 &  0.916 &  0.861 &  0.862 \\ 
&  200 &  0.971 &  0.970 &  0.961 &  0.970 &  0.737 &  0.705 &  0.942 &  0.931 &  0.951 &  0.943 &  \textbf{ 0.972 } &  0.909 &  0.927 &  0.947 &  0.952 &  0.891 &  0.895 &  0.911 &  0.962 &  0.961 &  0.943 &  0.949 \\ 
&  300 &  \textbf{ 0.984 } &  0.981 &  0.982 &  0.982 &  0.784 &  0.716 &  0.967 &  0.955 &  0.972 &  0.970 &  0.983 &  0.919 &  0.945 &  0.973 &  0.980 &  0.906 &  0.938 &  0.947 &  0.977 &  0.976 &  0.967 &  0.977 \\ 
&  400 &  0.984 &  0.986 &  0.980 &  \textbf{ 0.989 } &  0.766 &  0.756 &  0.974 &  0.960 &  0.976 &  0.975 &  0.986 &  0.917 &  0.952 &  0.975 &  0.984 &  0.908 &  0.948 &  0.952 &  0.979 &  0.979 &  0.974 &  0.984 \\ 
\hline 

\multirow{6}{1.5cm}{Average Power} &  30 &  \textbf{ 0.472 } &  0.403 &  0.430 &  0.298 &  0.227 &  0.334 &  0.327 &  0.335 &  0.340 &  0.354 &  0.381 &  0.312 &  0.314 &  0.297 &  0.102 &  0.304 &  0.364 &  0.316 &  0.425 &  0.428 &  0.287 &  0.304 \\ 
&  50 &  \textbf{ 0.687 } &  0.626 &  0.659 &  0.493 &  0.234 &  0.341 &  0.466 &  0.505 &  0.483 &  0.523 &  0.598 &  0.387 &  0.419 &  0.414 &  0.279 &  0.391 &  0.569 &  0.473 &  0.629 &  0.637 &  0.492 &  0.502 \\ 
&  100 &  \textbf{ 0.872 } &  0.830 &  0.848 &  0.793 &  0.312 &  0.388 &  0.650 &  0.719 &  0.673 &  0.724 &  0.805 &  0.462 &  0.565 &  0.615 &  0.681 &  0.473 &  0.727 &  0.579 &  0.815 &  0.818 &  0.755 &  0.769 \\ 
&  200 &  \textbf{ 0.958 } &  0.921 &  0.941 &  0.894 &  0.353 &  0.443 &  0.786 &  0.855 &  0.802 &  0.827 &  0.893 &  0.527 &  0.723 &  0.765 &  0.870 &  0.537 &  0.817 &  0.682 &  0.888 &  0.895 &  0.897 &  0.903 \\ 
&  300 &  \textbf{ 0.986 } &  0.957 &  0.971 &  0.915 &  0.365 &  0.476 &  0.821 &  0.893 &  0.839 &  0.857 &  0.926 &  0.554 &  0.777 &  0.839 &  0.913 &  0.562 &  0.842 &  0.729 &  0.923 &  0.930 &  0.940 &  0.948 \\ 
&  400 &  \textbf{ 0.995 } &  0.981 &  0.990 &  0.933 &  0.367 &  0.506 &  0.837 &  0.913 &  0.849 &  0.863 &  0.958 &  0.569 &  0.792 &  0.870 &  0.944 &  0.575 &  0.857 &  0.754 &  0.952 &  0.962 &  0.968 &  0.976 \\ 
\hline 
\hline 

\multirow{6}{1.5cm}{Average Gap} &  30 &  \textbf{ 0.055 } &  0.124 &  0.097 &  0.229 &  0.300 &  0.193 &  0.200 &  0.191 &  0.187 &  0.173 &  0.146 &  0.215 &  0.213 &  0.230 &  0.424 &  0.223 &  0.163 &  0.211 &  0.102 &  0.099 &  0.240 &  0.222 \\ 
&  50 &  \textbf{ 0.043 } &  0.104 &  0.071 &  0.237 &  0.496 &  0.389 &  0.264 &  0.226 &  0.247 &  0.207 &  0.132 &  0.344 &  0.312 &  0.316 &  0.451 &  0.339 &  0.161 &  0.258 &  0.101 &  0.094 &  0.238 &  0.228 \\ 
&  100 &  \textbf{ 0.016 } &  0.058 &  0.040 &  0.095 &  0.575 &  0.499 &  0.238 &  0.169 &  0.215 &  0.163 &  0.083 &  0.425 &  0.322 &  0.273 &  0.207 &  0.415 &  0.160 &  0.309 &  0.073 &  0.069 &  0.133 &  0.119 \\ 
&  200 &  \textbf{ 0.002 } &  0.039 &  0.019 &  0.066 &  0.606 &  0.517 &  0.173 &  0.105 &  0.157 &  0.133 &  0.066 &  0.432 &  0.237 &  0.195 &  0.090 &  0.423 &  0.142 &  0.278 &  0.072 &  0.065 &  0.062 &  0.057 \\ 
&  300 &  \textbf{ 0.000 } &  0.028 &  0.014 &  0.071 &  0.621 &  0.510 &  0.165 &  0.093 &  0.146 &  0.129 &  0.060 &  0.432 &  0.209 &  0.147 &  0.073 &  0.424 &  0.144 &  0.257 &  0.063 &  0.056 &  0.046 &  0.037 \\ 
&  400 &  \textbf{ 0.001 } &  0.014 &  0.006 &  0.062 &  0.628 &  0.490 &  0.159 &  0.082 &  0.147 &  0.133 &  0.038 &  0.427 &  0.204 &  0.126 &  0.052 &  0.421 &  0.139 &  0.241 &  0.043 &  0.033 &  0.028 &  0.020 \\ 
\hline 
\hline 

\multirow{6}{1.5cm}{Max Gap} &  30 &  \textbf{ 0.199 } &  0.226 &  0.275 &  0.378 &  0.718 &  0.492 &  0.513 &  0.431 &  0.513 &  0.404 &  0.419 &  0.619 &  0.535 &  0.434 &  0.658 &  0.583 &  0.444 &  0.402 &  0.270 &  0.243 &  0.371 &  0.345 \\ 
&  50 &  \textbf{ 0.182 } &  0.204 &  0.203 &  0.434 &  0.954 &  0.773 &  0.616 &  0.411 &  0.608 &  0.431 &  0.273 &  0.836 &  0.685 &  0.501 &  0.657 &  0.771 &  0.396 &  0.667 &  0.231 &  0.215 &  0.419 &  0.378 \\ 
&  100 &  \textbf{ 0.077 } &  0.195 &  0.139 &  0.243 &  1.000 &  0.919 &  0.482 &  0.353 &  0.442 &  0.398 &  0.351 &  0.953 &  0.725 &  0.540 &  0.594 &  0.947 &  0.384 &  0.876 &  0.248 &  0.235 &  0.280 &  0.231 \\ 
&  200 &  \textbf{ 0.006 } &  0.254 &  0.093 &  0.379 &  1.000 &  0.928 &  0.755 &  0.511 &  0.681 &  0.651 &  0.407 &  0.964 &  0.778 &  0.587 &  0.438 &  0.958 &  0.615 &  0.797 &  0.410 &  0.364 &  0.329 &  0.330 \\ 
&  300 &  \textbf{ 0.002 } &  0.213 &  0.088 &  0.427 &  1.000 &  0.933 &  0.841 &  0.547 &  0.770 &  0.718 &  0.375 &  0.969 &  0.897 &  0.692 &  0.448 &  0.963 &  0.700 &  0.703 &  0.400 &  0.353 &  0.326 &  0.275 \\ 
&  400 &  \textbf{ 0.005 } &  0.123 &  0.029 &  0.383 &  1.000 &  0.944 &  0.849 &  0.521 &  0.807 &  0.763 &  0.228 &  0.961 &  0.949 &  0.677 &  0.308 &  0.955 &  0.744 &  0.737 &  0.285 &  0.220 &  0.222 &  0.154 \\ 
\hline 
\end{tabular}  \label{tab:pow4} 
\end{table}

\end{landscape}

\includepdf[pages=-]{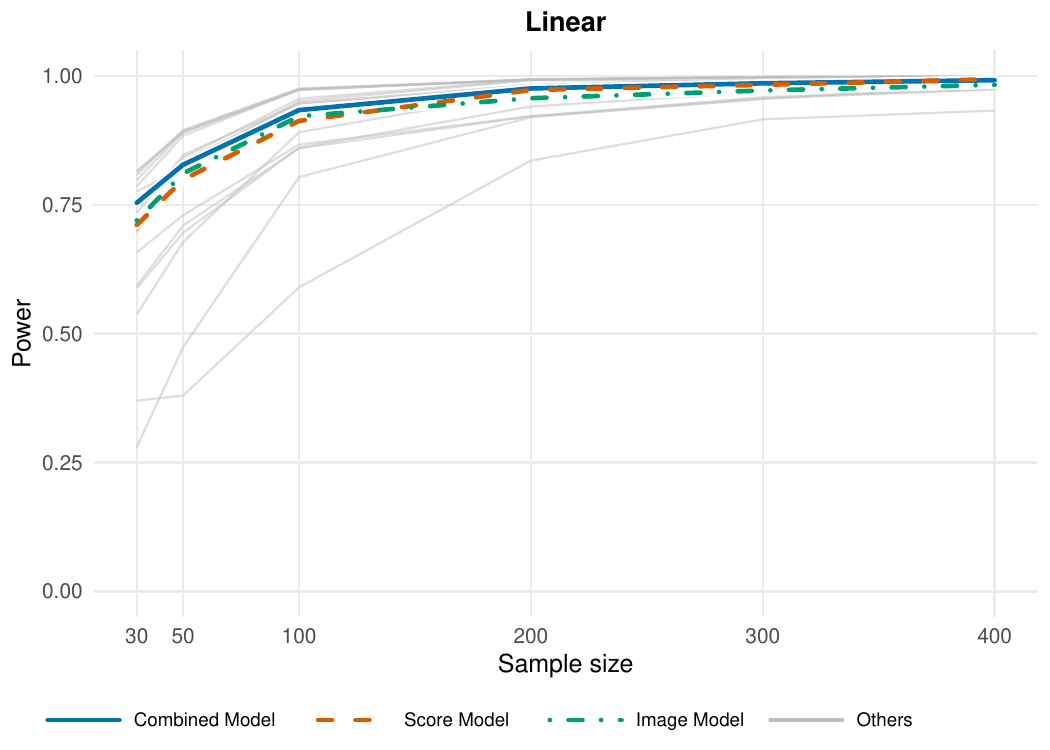}

\includepdf[pages=-]{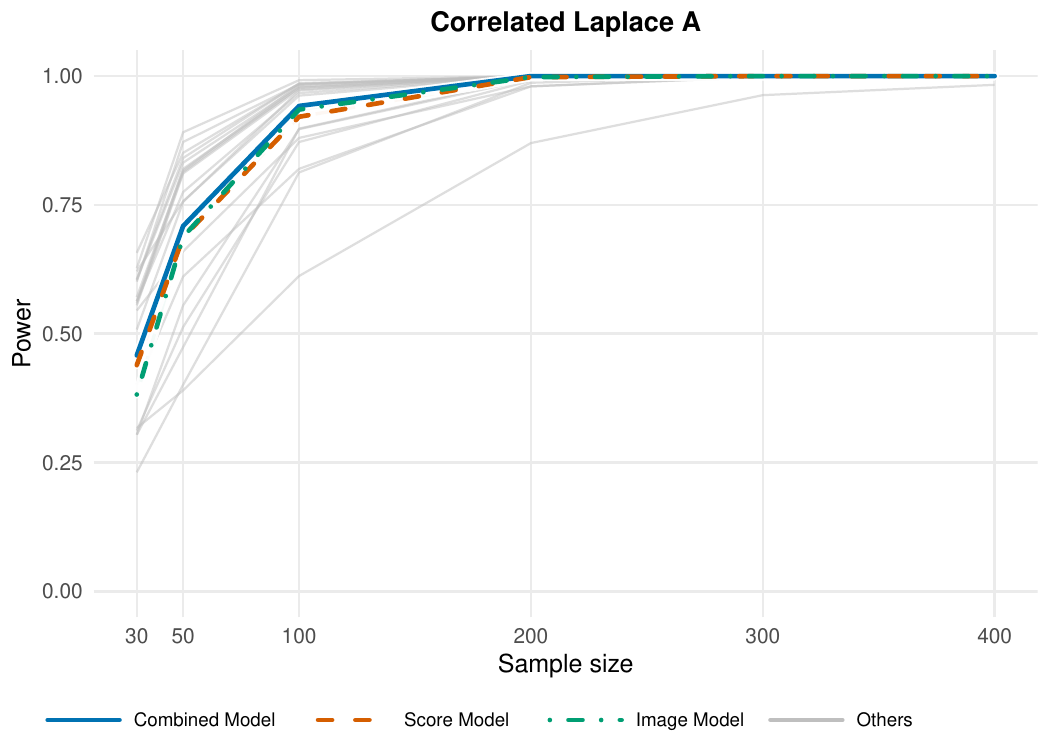}

\end{document}